\documentclass{article}
\newif\ifemail
\newif\ifchecklist
\newif\ifappendix
\newif\ifbanner
\usepackage{soul}
\usepackage[small]{caption}
\usepackage{graphicx}
\usepackage{amsmath}
\usepackage{bbm}
\usepackage{amsthm}
\usepackage{booktabs}
\usepackage{algorithm}
\usepackage{amssymb}
\usepackage{multirow}
\usepackage{wrapfig}
\usepackage{makecell}
\usepackage{colortbl}
\usepackage{bbding} 
\usepackage{color,xcolor}
\usepackage{subfigure}
\usepackage{pifont} 
\newcommand{\cmark}{\ding{51}}%
\newcommand{\xmark}{\ding{55}}%
\newtheorem{theorem}{Theorem}[section]
\newtheorem{sketch_definition}{Sketch Definition}[section]

\newtheorem{lemma}[theorem]{Lemma}
\newtheorem{corollary}[theorem]{Corollary}
\newtheorem{definition}[theorem]{Definition}
\newtheorem{assumption}[theorem]{Assumption}
\usepackage{setspace}
\usepackage{algorithmicx}

\definecolor{Blue}{RGB}{0, 0, 255}
\definecolor{Aquamarine}{RGB}{127, 255, 212}
\definecolor{Sepia}{RGB}{112, 66, 20}
\definecolor{BrickRed}{RGB}{203, 65, 84}
\colorlet{my-red}{BrickRed!90!Sepia}
\colorlet{my-blue}{Aquamarine!30!Blue}
\usepackage[backref=page]{hyperref}
\hypersetup{
  colorlinks,
  urlcolor  = my-red,
  linkcolor = my-blue,
  citecolor = my-blue,
}

\newcommand{\imineq}[2]{\vcenter{\hbox{\includegraphics[height=#2ex]{#1}}}}
\usepackage{tikz}
\usepackage{mathtools}
\usepackage[normalem]{ulem}  %
\usepackage{tabu}
\usepackage{pgfplots,pgfplotstable}
\usepgfplotslibrary{fillbetween}
\usepackage{calc}
\usepackage{pbox}
\usepackage[noend]{algpseudocode}
\usepackage{afterpage}

\newcommand{\config}[1]{$\mathbb{C}\mathbb{O}\mathbb{N}\mathbb{F}\mathbb{I}\mathbb{G}$\ #1}
\newcommand{\graycell}{\hspace{-1pt}\cellcolor{gray!25}}
\pgfplotsset{compat=1.14}	 %
\pgfplotsset{compat/show suggested version=false}

\usepackage[final]{neurips_2024}\emailfalse\checklistfalse\appendixtrue\bannertrue
\usepackage[utf8]{inputenc} 
\usepackage[T1]{fontenc}    
\usepackage{url}            
\usepackage{amsfonts}       
\usepackage{nicefrac}       
\usepackage{microtype}      

\definecolor{olive}{rgb}{0.5, 0.5, 0.0}
\definecolor{maroon}{rgb}{0.69, 0.19, 0.38}
\definecolor{celestialblue}{rgb}{0.29, 0.59, 0.82}
\definecolor{darkgreen}{rgb}{0.0, 0.6, 0.0}
\definecolor{grey}{rgb}{0.5,0.5,0.5}
\definecolor{darkblue}{rgb}{0.19, 0.19, 0.62}
\definecolor{silver}{rgb}{0.7,0.7,0.7}
\definecolor{darkcyan}{rgb}{0.0, 0.55, 0.55}

\def\clap#1{\hbox to 0pt{\hss #1\hss}}%

\newcommand\undefcolumntype[1]{\expandafter\let\csname NC@find@#1\endcsname\relax}

\newcommand\ve[1]{\mathbf{#1}}

\definecolor{C0}{rgb}{0.121569, 0.466667, 0.705882}
\definecolor{C1}{rgb}{1.000000, 0.498039, 0.054902}
\definecolor{C2}{rgb}{0.172549, 0.627451, 0.172549}
\definecolor{C3}{rgb}{0.839216, 0.152941, 0.156863}
\definecolor{C4}{rgb}{0.580392, 0.403922, 0.741176}
\definecolor{C5}{rgb}{0.549020, 0.337255, 0.294118}
\definecolor{C6}{rgb}{0.890196, 0.466667, 0.760784}
\definecolor{C7}{rgb}{0.498039, 0.498039, 0.498039}
\definecolor{C8}{rgb}{0.737255, 0.741176, 0.133333}
\definecolor{C9}{rgb}{0.090196, 0.745098, 0.811765}

\title{Elucidating the Design Space of Dataset Condensation}
\author{Shitong Shao$^\dag$$^\diamondsuit$, Zikai Zhou$^\dag$$^\diamondsuit$, Huanran Chen$^\dag$$^{\ddag}$, Zhiqiang Shen$^{\dag}$$^*$ \\
$^\dag$ Mohamed bin Zayed University of AI, $^{\ddag}$ Tsinghua University\\
$^\diamondsuit$ The Hong Kong University of Science and Technology (Guangzhou)\\
\texttt{\{1090784053sst,choukai003\}@gmail.com},\quad\texttt{huanran\_chen@outlook.com}\\
\texttt{zhiqiang.shen@mbzuai.ac.ae},\quad$*$: Corresponding author\\
}
\begin{document}
\maketitle

\begin{abstract}

Dataset condensation, a concept within \textit{data-centric learning}, aims to efficiently transfer critical attributes from an original dataset to a synthetic version, meanwhile maintaining both diversity and realism of syntheses. This approach can significantly improve model training efficiency and is also adaptable for multiple application areas. Previous methods in dataset condensation have faced several challenges: some incur high computational costs which limit scalability to larger datasets (\textit{e.g.,} MTT, DREAM, and TESLA), while others are restricted to less optimal design spaces, which could hinder potential improvements, especially in smaller datasets (\textit{e.g.,} SRe$^2$L, G-VBSM, and RDED). To address these limitations, we propose a comprehensive designing-centric framework that includes specific, effective strategies like implementing soft category-aware matching, adjusting the learning rate schedule and applying small batch-size. These strategies are grounded in both empirical evidence and theoretical backing. Our resulting approach, \textbf{E}lucidate \textbf{D}ataset \textbf{C}ondensation (\textbf{EDC}), establishes a benchmark for both small and large-scale dataset condensation. In our testing, EDC achieves state-of-the-art accuracy, reaching 48.6\% on ImageNet-1k with a ResNet-18 model at an IPC of 10, which corresponds to a compression ratio of 0.78\%. This performance surpasses those of SRe$^2$L, G-VBSM, and RDED by margins of 27.3\%, 17.2\%, and 6.6\%, respectively.

\end{abstract}

\section{Introduction}
Dataset condensation, also known as dataset distillation, has emerged in response to the ever-increasing training demands of advanced deep learning models~\citep{ResNet,ResNetv2,GPT3}. This task addresses the challenge of requiring massive amount of data to train high-precision models while also being bounded by resource constraints~\citep{VIT,FM_KT}. In the conventional setup of this problem, the original dataset acts as a ``teacher'', distilling and preserving essential information into a smaller, surrogate ``student'' dataset. The ultimate goal of this technique is to achieve comparable performance of models trained on the original and condensed datasets from scratch. This task has become popular in various downstream applications, including continual learning~\citep{dd_continual_learning_1,dd_continual_learning_2,dd_continual_learning_3}, neural architecture search~\citep{dd_nas_1,dd_dist_matching,dd_gradient_matching}, and training-free network slimming~\citep{liu2017learning}.

However, the common solution in traditional dataset distillation methods of bi-level optimization requires prohibitively expensive computation, which limits the practical usage, as in prior works~\citep{dd_mtt,dd_datadam,dd_dream}. This has become more severe particularly when being applied to large-scale datasets like ImageNet-1k~\citep{ILSVRC15}. In response, the uni-level optimization paradigm has gained significant attention as an alternative solution, with recent contributions from the research community~\citep{dd_sre2l,yin2024dataset,shao2023generalized} highlighting its applicability. These methods primarily leverage the rich and extensive information from static, pre-trained observer models, to facilitate a more streamlined optimization process for synthesizing a condensed dataset without the need to adjust other parameters (\textit{e.g.,} those within the observer models). While uni-level optimization has demonstrated remarkable performance on large datasets, it has yet to achieve the competitive accuracy levels seen with classical methods on small-scale datasets like CIFAR-10/100~\citep{CIFAR}. Moreover, the recently proposed training-free method RDED~\citep{RDED} outperforms training-based methods in efficiency and maintains effectiveness, yet it overlooks the potential information incompleteness due to the lack of optimization on syntheses. Also, some simple but promising skills (\textit{e.g.,} smoothing learning rate schedule) that could enhance performance have not been well-explored in the existing literature. We observe that a performance improvement of 16.2\% in RDED comes from these techniques in this paper rather than the proposed data synthesis approach.

\begin{figure*}[t]
\centering
\includegraphics[width=1.\textwidth]{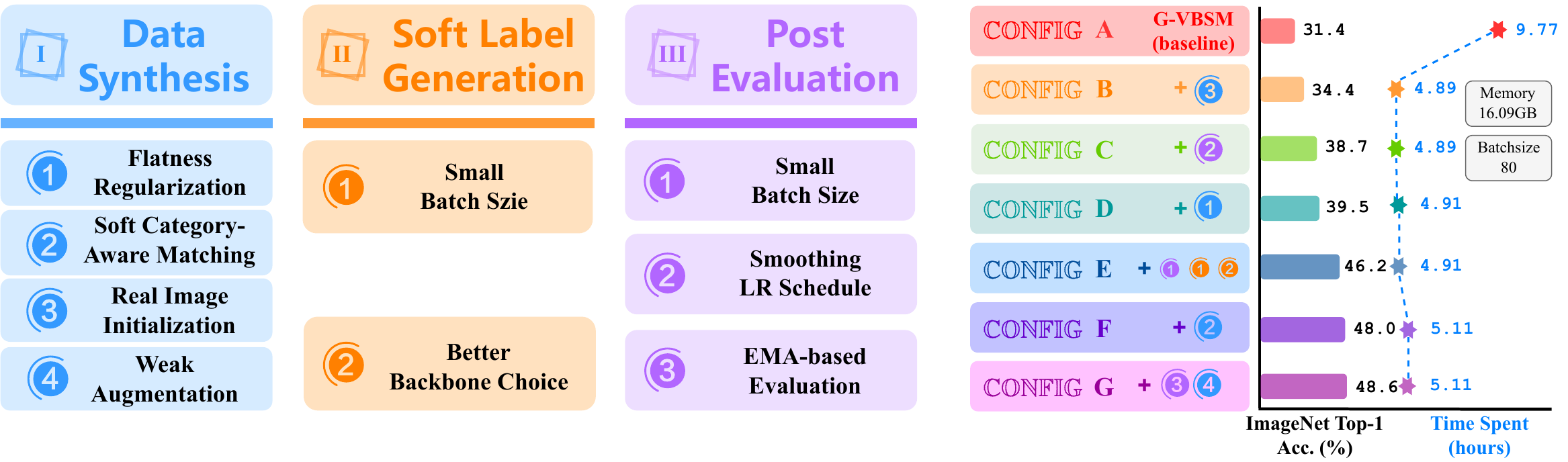}
\caption{\textbf{Illustration of Elucidating Dataset Condensation (EDC).} \textbf{Left:} The overall of our better design choices in dataset condensation on ImageNet-1k. \textbf{Right:} The evaluation performance and data synthesis required time of different configurations on ResNet-18 with IPC 10. Our integral EDC refers to \config{G}.}
\vspace{-1ex}
\label{figure:illustration}
\end{figure*}

These drawbacks show the constraints of previous methods in several respects, highlighting the need for a thorough investigation and assessment of potential limitations in prior frameworks. In contrast to earlier strategies that targeted one or a few specific improvements, our approach systematically examines all possible facets and integrates them into our comprehensive framework. To establish a strong framework, we carefully analyze all potential deficiencies in different stages of the data synthesis, soft label generation, and post-evaluation stages during dataset condensation, resulting in an extensive exploration of the design space on both large-scale and small-scale datasets. As a result, we introduce \textbf{E}lucidate \textbf{D}ataset \textbf{C}ondensation (\textbf{EDC}), which includes a range of concrete and effective enhancement skills for dataset condensation (refer to Fig.~\ref{figure:illustration}). For instance, \textit{soft category-aware matching ($\imineq{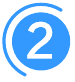}{2.4}$)} ensures consistent category representation between the original and condensed data batches for more precise matching. Overall, EDC not only achieves state-of-the-art performance on CIFAR-10, CIFAR-100, Tiny-ImageNet, ImageNet-10, and ImageNet-1k, using only half of the computational cost compared to the \textit{baseline} G-VBSM, but it also provides in-depth both empirical and theoretical insights and explanations that affirm the soundness of our design decisions. Our code is available at: \url{https://github.com/shaoshitong/EDC}.
\section{Dataset Condensation}

\noindent{\bf Preliminary.} Dataset condensation involves generating a synthetic dataset $\mathcal{D}^\mathcal{S} := \{\ve{x}^\mathcal{S}_i, \ve{y}^\mathcal{S}_i\}_{i=1}^{\lvert \mathcal{D}^\mathcal{S} \rvert}$ consisting of images $\mathcal{X}^\mathcal{S}$ and labels $\mathcal{Y}^\mathcal{S}$, designed to be as informative as the original dataset $\mathcal{D}^\mathcal{T} := \{\ve{x}^\mathcal{T}_i, \ve{y}^\mathcal{T}_i\}_{i=1}^{\lvert \mathcal{D}^\mathcal{T} \rvert}$, which includes images $\mathcal{X}^\mathcal{T}$ and labels $\mathcal{Y}^\mathcal{T}$. The synthetic dataset $\mathcal{D}^\mathcal{S}$ is substantially smaller in size than $\mathcal{D}^\mathcal{T}$ ($\lvert \mathcal{D}^\mathcal{S} \rvert \ll \lvert \mathcal{D}^\mathcal{T} \rvert$). The goal of this process is to maintain the critical attributes of $\mathcal{D}^\mathcal{T}$ to ensure robust or comparable performance during evaluations on real test protocol $\mathcal{P}_\mathcal{D}$.
\begin{equation}
\begin{aligned}
   \operatorname{arg\,min} \mathbb{E}_{(\ve{x},\ve{y})\sim \mathcal{P}_\mathcal{D}}[\ell_\textbf{\textrm{eval}}(\ve{x},\ve{y},\phi^*)],\ \ \textrm{where}\ \phi^* = \operatorname{arg\,min}_{\phi}\mathbb{E}_{(\ve{x}_i^\mathcal{S},\ve{y}_i^\mathcal{S})\sim \mathcal{D}^\mathcal{S}}[\ell (\phi(\ve{x}_i^\mathcal{S}),\ve{y}_i^\mathcal{S})].
\end{aligned}
\label{bg:eq1}
\end{equation}
Here, $\ell_\textbf{\textrm{eval}}(\cdot,\cdot,\phi^*)$ represents the evaluation loss function, such as cross-entropy loss, which is parameterized by the neural network $\phi^*$ that has been optimized from the distilled dataset $\mathcal{D}^\mathcal{S}$. The data synthesis process primarily determines the quality of the distilled datasets, which transfers desirable knowledge from $\mathcal{D}^\mathcal{T}$ to $\mathcal{D}^\mathcal{S}$ through various matching mechanisms, such as trajectory matching~\citep{dd_mtt}, gradient matching~\citep{dd_gradient_matching}, distribution matching~\citep{dd_dist_matching} and generalized matching~\citep{shao2023generalized}.

\noindent{\bf Small-scale vs. Large-scale Dataset Condensation/Distillation.} Traditional dataset condensation algorithms, as referenced in studies such as~\citep{dd_begin,dd_mtt,dd_tesla,dd_CAFE,dd_kip}, encounter computational challenges and are generally confined to small-scale datasets like CIFAR-10/100~\citep{CIFAR}, or larger datasets with limited class diversity, such as ImageNette~\citep{dd_mtt} and ImageNet-10~\citep{dd_efficient_parameterization}. The primary inefficiency of these methods stems from their reliance on a bi-level optimization framework, which involves alternating updates between the synthetic dataset and the observer model utilized for distillation. This approach not only heavily depends on the model's intrinsic ability but also limits the versatility of the distilled datasets in generalizing across different architectures. In contrast, the uni-level optimization strategy, noted for its efficiency and enhanced performance on the regular 224$\times$224 scale of ImageNet-1k in recent research~\citep{dd_sre2l,shao2023generalized,yin2024dataset}, shows reduced effectiveness in smaller-scale datasets due to the massive optimization-based iterations required in the data synthesis process without a direct connection to actual data. Recent new methods in training-free distillation paradigms, such as in~\citep{RDED,dataset_quantization}, offer advancements in efficiency. However, these methods compromise data privacy by sharing original data and do not leverage statistical information from observer models to enhance the capability of synthetic data, thereby restraining their potential in a real environment.

\noindent{\bf Generalized Data Synthesis Paradigm.} We consistently describe algorithms \citep{dd_sre2l,yin2024dataset,shao2023generalized,RDED} that efficiently conduct data synthesis on ImageNet-1k as  ``generalized data synthesis'' as these methods are applicable for both small and large-scale datasets. This direction usually avoids the inefficient bi-level optimization and includes both image and label synthesis phases. Note that several recent works~\citep{add_related_work1,add_related_work2,add_related_work3}, particularly DANCE~\citep{add_related_work1}, can also effectively be applied to ImageNet-1k, but these methods lack enhancements in soft label generation and post-evaluation. Specifically, generalized data synthesis involves first generating highly condensed images followed by acquiring soft labels through predictions from a pre-trained model. The evaluation process resembles knowledge distillation~\citep{vanillakd}, aiming to transfer knowledge from a teacher to a student model~\citep{kdsurvey,vanillakd}. The primary distinction between the training-dependent \citep{dd_sre2l,yin2024dataset,shao2023generalized} and training-free paradigm \citep{RDED} centers on their approach to data synthesis. In detail, the training-dependent paradigm employs \textit{Statistical Matching (SM)} to extract pertinent information from the entire dataset.
\begin{equation}
\footnotesize
\begin{aligned}
\mathcal{L}_\textbf{\textrm{syn}}&=||p(\mu|\mathcal{X}^\mathcal{S})-p(\mu|\mathcal{X}^\mathcal{T})||_2+||p(\sigma^2|\mathcal{X}^\mathcal{S})-p(\sigma^2|\mathcal{X}^\mathcal{T})||_2,\ s.t.\ \mathcal{L}_\textbf{\textrm{syn}}\sim \mathbb{S}_\textrm{match}, \\
\mathcal{X}^{\mathcal{S}*} &= \operatorname*{arg\,min}_{\mathcal{X}^{\mathcal{S}}} \mathbb{E}_{\mathcal{L}_\textbf{\textrm{syn}}\sim \mathbb{S}_\textrm{match}}[\mathcal{L}_\textbf{\textrm{syn}}(\mathcal{X}^{\mathcal{S}},\mathcal{X}^{\mathcal{T}})] ,\\
\end{aligned}
\label{eq:definition_sm}
\end{equation} 
where $\mathbb{S}_\textrm{match}$ represents the extensive collection of statistical matching operators, which operate across a variety of network architectures and layers as described by \citep{shao2023generalized}. Here, $\mu$ and $\sigma^2$ are defined as the mean and variance, respectively. For more detailed theoretical insights, please refer to Definition~\ref{the:sketch_sm}. The training-free approach, as discussed in \citep{RDED, dataset_quantization}, employs a direct reconstruction method for the original dataset, aiming to generate simplified representations of images.
\vspace{-0.05in}
\begin{equation}
\footnotesize
\begin{aligned}
\mathcal{X}^\mathcal{S} = \bigcup_{i=1}^\mathbf{C}\mathcal{X}^\mathcal{S}_i,\ \mathcal{X}^\mathcal{S}_i=\{\ve{x}_j^i =\textrm{concat}(\{\tilde{\ve{x}}_k\}_{k=1}^N\subset \mathcal{X}^\mathcal{T}_i)\}_{j=1}^{\mathsf{IPC}}, \\
\end{aligned}
\label{eq:definition_rede}
\end{equation}
where $\mathbf{C}$ denotes the number of classes, $\textrm{concat}(\cdot)$ represents the concatenation operator, $\mathcal{X}^\mathcal{S}_i$ signifies the set of condensed images belonging to the $i$-th class, and $\mathcal{X}^\mathcal{T}_i$ corresponds to the set of original images of the $i$-th class. It is important to note that the default settings for $N$ are 1 and 4, as specified in the works~\citep{dataset_quantization} and~\citep{RDED}, respectively. Using one or more observer models, denoted as $\{\phi_{i}\}_{i=1}^N$, we then derive the soft labels $\mathcal{Y}^\mathcal{S}$ from the condensed image set $\mathcal{X}^\mathcal{S}$.
\vspace{-0.05in}
\begin{equation}
\footnotesize
\begin{aligned}
\mathcal{Y}^\mathcal{S} = \bigcup_{\ve{x}_i^\mathcal{S}\subset\mathcal{X}^\mathcal{S}}\frac{1}{N}\sum_{i=1}^N\phi_i(\ve{x}_i^\mathcal{S}). \\
\end{aligned}
\label{eq:definition_soft_label}
\end{equation}
This plug-and-play component, as outlined in SRe$^2$L~\citep{dd_sre2l} and IDC~\citep{dd_efficient_parameterization}, plays a crucial role for enhancing the generalization ability of the distilled dataset $\mathcal{D}^\mathcal{S}$.
\section{Improved Design Choices}

\begin{figure*}[t]
\centering    
\subfigure[] {
 \label{fig:a}     
\includegraphics[width=0.48\columnwidth,trim={0cm 1cm 0cm 1cm},clip]{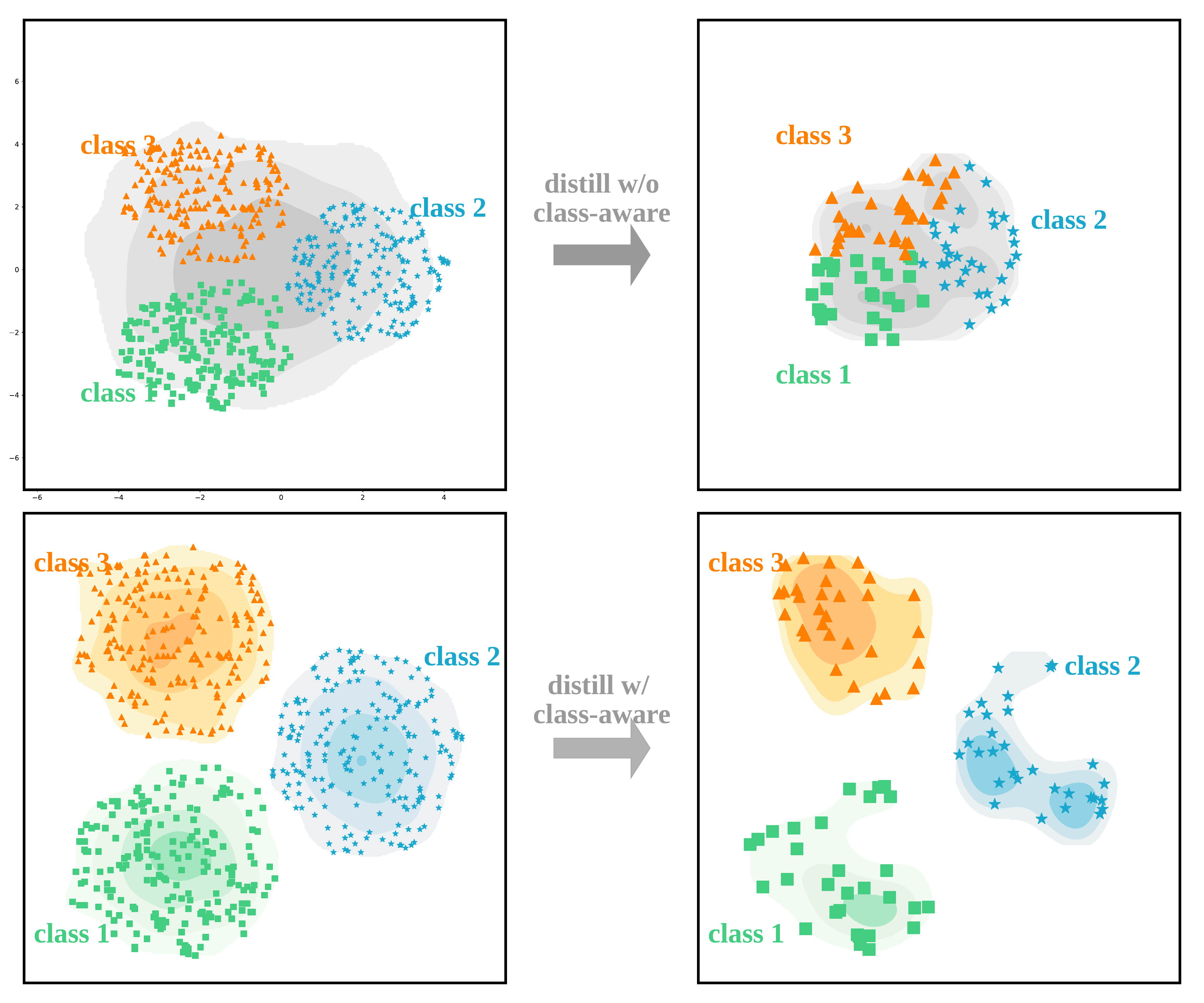}  
}
\hspace{-3pt}
\subfigure[] {
\label{fig:b}
\includegraphics[width=0.25\columnwidth,trim={0cm 2.74cm 0cm 0cm},clip]{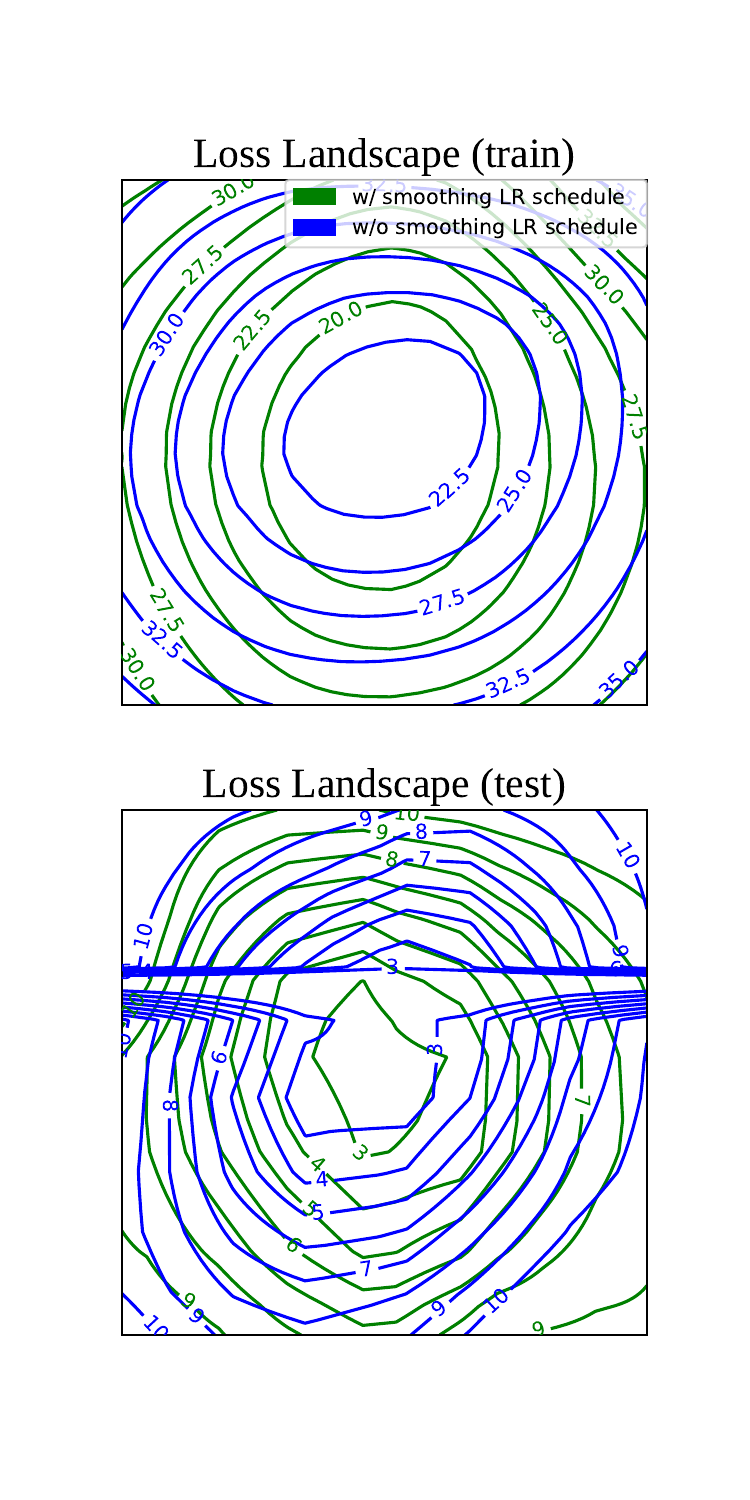}
} 
\hspace{-3pt}
\subfigure[] {
\label{fig:b}     
\includegraphics[width=0.21\columnwidth,trim={0cm 0.8cm 0cm 0cm},clip]{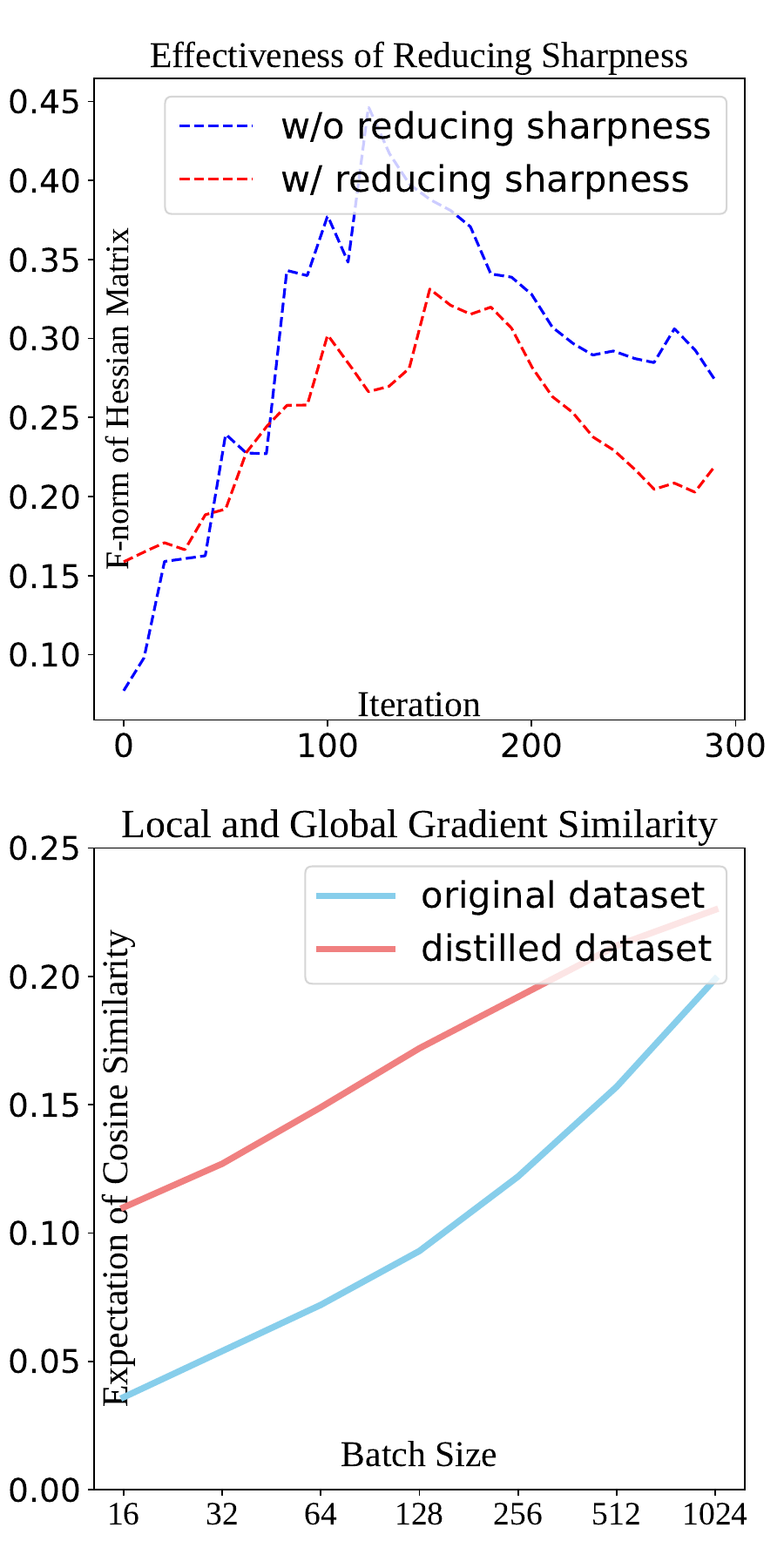}
} 
\vspace{-0.1in}
\caption{\textbf{(a):} Illustration of soft category-aware matching $\left(\imineq{figures/serial_number/data_synthesis_2.pdf}{2.4}\right)$ using a Gaussian distribution in $\mathbb{R}^2$. \textbf{(b):} The effect of employing smoothing LR schedule $\left(\imineq{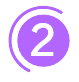}{2.4}\right)$ on loss landscape sharpness reduction. \textbf{(c) top:} The role of flatness regularization $\left(\imineq{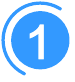}{2.4}\right)$ in reducing the Frobenius norm of the Hessian matrix driven by data synthesis iteration. \textbf{(c) bottom:} Cosine similarity comparison between local gradients (obtained from original and distilled datasets via random batch selection) and the global gradient (obtained from gradient accumulation).}     
\label{fig:empirical_analysis}
\vspace{-7pt}
\end{figure*}
Design choices in data synthesis, soft label generation, and post-evaluation significantly influence the generalization capabilities of condensed datasets. Effective strategies for small-scale datasets are well-explored, yet these approaches are less examined for large-scale datasets. We first delineate the limitations of existing algorithms' design choices on ImageNet-1k. We then propose solutions, providing experimental results as shown in Fig.~\ref{figure:illustration}. For most design choices, we offer both theoretical analysis and empirical insights to facilitate a thorough understanding, as detailed in Sec.~\ref{sec:solution}.

\subsection{Limitations of Prior Methods}
\label{sec:prior_limitation}

\textbf{Lacking Realism \textit{$\left(\text{solved by }\imineq{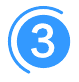}{2.4}\right)$}.} Training-dependent condensation algorithms for datasets, particularly those employed for large-scale datasets, typically initiate the optimization process using Gaussian noise inputs~\citep{dd_sre2l, yin2024dataset, shao2023generalized}. This initial choice complicates the optimization process and often results in the generation of synthetic images that do not exhibit high levels of realism. The limitations in visualization associated with previous approaches are detailed in Appendix~\ref{apd:visualization}.

\textbf{Coarse-grained Matching Mechanism \textit{$\left(\text{solved by }\imineq{figures/serial_number/data_synthesis_2.pdf}{2.4}\right)$}.} The \textit{Statistical Matching (SM)}-based pipeline~\citep{dd_sre2l,yin2024dataset,shao2023generalized} computes the global mean and variance by aggregating samples across all categories and uses these statistical parameters for matching purposes. However, this strategy exhibits two critical drawbacks: it does not account for the domain discrepancies among different categories, and it fails to preserve the integrity of category-specific information across the original and condensed samples within each batch. These limitations result in a coarse-grained matching approach that diminishes the accuracy of the matching process.

\textbf{Overly Sharp of Loss Landscape \textit{$\left(\text{solved by }\imineq{figures/serial_number/data_synthesis_1.pdf}{2.4}\ \text{and}\ \imineq{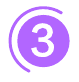}{2.4}\right)$}.} The optimization objective $\mathcal{L}(\theta)$ can be expanded through a second-order Taylor expansion as $\mathcal{L}(\theta^*)+(\theta-\theta^*)^\mathrm{T}\nabla_\theta\mathcal{L}(\theta^*)+(\theta-\theta^*)^\mathrm{T}\mathbf{H}(\theta-\theta^*)$, with an upper bound of $\mathcal{L}(\theta^*)+||\mathbf{H}||_\mathbf{F}\mathbb{E}[||\theta-\theta^*||_2^2]$ upon model convergence~\citep{cwa}. However, earlier training-dependent condensation algorithms neglect to minimize the Frobenius norm of the Hessian matrix $\mathbf{H}$ to obtain a flat loss landscape for enhancing its generalization capability through sharpness-aware minimization theory~\citep{iclr2020_sam,eccvw_sam}. Please see Appendix~\ref{apd:decouple} for more formal information.

\textbf{Irrational Hyperparameter Settings \textit{$\left(\text{solved by }\imineq{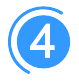}{2.4}\text{, }\imineq{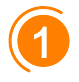}{2.4}\text{, }\imineq{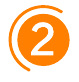}{2.4}\text{, }\imineq{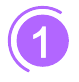}{2.4}\ \text{and}\ \imineq{figures/serial_number/post_evaluation_2.pdf}{2.4}\right)$}.} RDED~\citep{RDED} adopts a smoothing LR schedule $\left(\imineq{figures/serial_number/post_evaluation_2.pdf}{2.4}\right)$ and~\citep{liu2023dataset,yin2024dataset,RDED} use a reduced batch size $\left(\imineq{figures/serial_number/post_evaluation_1.pdf}{2.4}\imineq{figures/serial_number/soft_label_1.pdf}{2.4}\right)$ for post-evaluation on the full 224$\times$224 ImageNet-1k. These changes, although critical, lack detailed explanations and impact assessments in the existing literature. Our empirical analysis highlights a remarkable impact on performance: absent these modifications, RDED achieves only 25.8\% accuracy on ResNet18 with IPC 10. With these modifications, however, accuracy jumps to 42.0\%. In contrast, SRe$^2$L and G-VBSM do not incorporate such strategies in their experimental frameworks. This work aims to fill the gap by providing the first comprehensive empirical analysis and ablation study on the effects of these and similar improvements in the field.

\vspace{-0.07in}
\subsection{Our Solutions}
\label{sec:solution}
\vspace{-0.07in}
To address these limitations described above, we explore the design space and elaborately present a range of optimal solutions at both empirical and theoretical levels, as illustrated in Fig.~\ref{figure:illustration}.

\begin{wrapfigure}{r}{5.3cm}
\vspace{-10pt}
\includegraphics[height=0.23\textwidth,trim={0cm 0cm 0cm 0cm},clip]{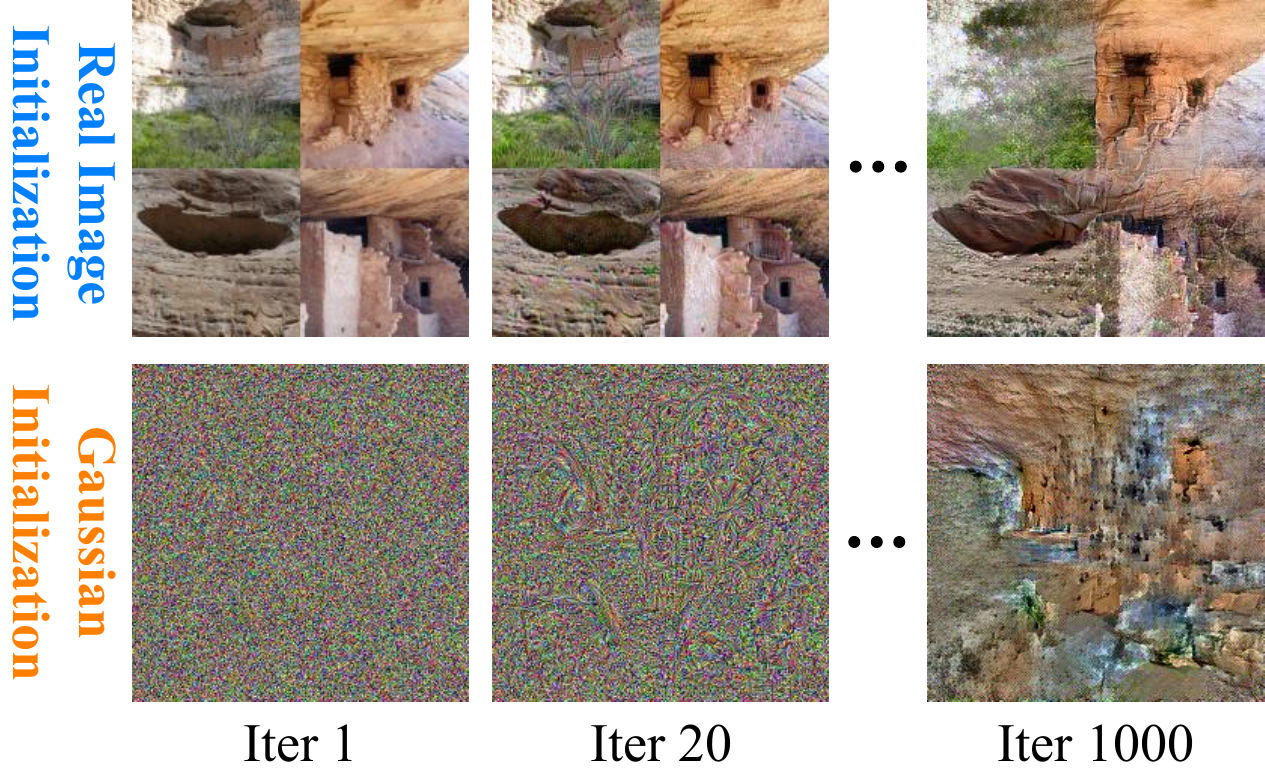}
\vspace{-4pt}
\caption{Comparison between real image initialization and random initialization.}
\label{fig:real_image_initialization}
\vspace{-15pt}
\end{wrapfigure}
\textbf{Real Image Initialization $\left(\imineq{figures/serial_number/data_synthesis_3.pdf}{2.4}\right)$.} Intuitively, using real images instead of Gaussian noise for data initialization during the data synthesis phase is a practical and effective strategy. As shown in Fig.~\ref{fig:real_image_initialization}, this method significantly improves the realism of the condensed dataset and simplifies the optimization process, thus enhancing the synthesized dataset's ability to generalize in post-evaluation tests. Additionally, we incorporate considerations of information density and efficiency by employing a training-free condensed dataset (e.g., RDED) for initialization at the start of the synthesis process. According to Theorem~\ref{the:real_vs_random}, based on optimal transport theory, the cost of transporting from a Gaussian distribution to the original data distribution is higher than using the training-free condensed distribution as the initial reference. This advantage also allows us to reduce the number of iterations needed to achieve results to half of those required by our baseline G-VBSM model, significantly boosting synthesis efficiency. 

\begin{theorem}
\label{the:real_vs_random}
(proof in Appendix~\ref{apd:random_vs_real}) Considering samples $\mathcal{X}^\mathcal{S}_\textbf{\textrm{real}}$, $\mathcal{X}^\mathcal{S}_\textbf{\textrm{free}}$, and $\mathcal{X}^\mathcal{S}_\textbf{\textrm{random}}$ from the original data, training-free condensed (\textit{e.g.,} RDED), and Gaussian distributions, respectively, let us assume a cost function defined in optimal transport theory that satisfies $\mathbb{E}[c(a-b)]\propto 1/I(\textrm{Law}(a),\textrm{Law}(b))$. Under this assumption, it follows that $\mathbb{E}[c(\mathcal{X}^\mathcal{S}_\textbf{\textrm{real}}-\mathcal{X}^\mathcal{S}_\textbf{\textrm{free}})] \leq \mathbb{E}[c(\mathcal{X}^\mathcal{S}_\textbf{\textrm{real}}-\mathcal{X}^\mathcal{S}_\textbf{\textrm{random}})]$.
\end{theorem}

\textbf{Soft Category-Aware Matching $\left(\imineq{figures/serial_number/data_synthesis_2.pdf}{2.4}\right)$.} Previous dataset condensation methods~\citep{dd_sre2l,yin2024dataset,shao2023generalized} based on the \textit{Statistical Matching} (SM) framework have shown satisfactory results predominantly when the data follows a unimodal distribution (\textit{e.g.,} a single Gaussian). This limitation is illustrated with a simple example in Fig.~\ref{fig:empirical_analysis} (a). Typically, datasets consist of multiple classes with significant variations among their class distributions. Traditional SM-based methods compress data by collectively processing all samples, thus neglecting the differences between classes. As shown in the top part of Fig.~\ref{fig:empirical_analysis} (a), this method enhances information density but also creates a big mismatch between the condensed source distribution $\mathcal{X}^\mathcal{S}$ and the target distribution $\mathcal{X}^\mathcal{T}$. To tackle this problem, we propose the use of a Gaussian Mixture Model (GMM) to effectively approximate any complex distribution. This solution is theoretically justifiable by the Tauberian Theorem under certain conditions (detailed proof is provided in Appendix~\ref{apd:category_matching}). In light of this, we define two specific approaches to \textit{Statistical Matching}:
\begin{sketch_definition}
\label{the:sketch_sm}
(formal definition in Appendix~\ref{apd:category_matching}) Given $N$ random samples $\{x_i\}_{i=1}^{N}$ with an unknown distribution $p_{\textrm{mix}}(x)$, we define two forms to statistical matching. \textbf{Form (1):} involves synthesizing $M$ distilled samples $\{y_i\}_{i=1}^{M}$, where $M\ll N$, ensuring that the variances and means of both $\{x_i\}_{i=1}^{N}$ and $\{y_i\}_{i=1}^{M}$ are consistent. \textbf{Form (2):} treats $p_{\textrm{mix}}(x)$ as a GMM with $\mathbf{C}$ components. For random samples $\{x_i^j\}_{i=1}^{N_j}$ ($\sum_j N_j = N$) within each component $c_j$, we synthesize $M_j$ ($\sum_j M_j = M$) distilled samples $\{y^j_i\}_{i=1}^{M_j}$, where $M_j\ll N_j$, to maintain the consistency of variances and means between $\{x^j_i\}_{i=1}^{N_j}$ and $\{y^j_i\}_{i=1}^{M_j}$.
\end{sketch_definition}
In general, SRe$^2$L, CDA, and G-VBSM are all categorized under \textbf{Form (1)}, as shown in Fig.~\ref{fig:empirical_analysis} (a) at the top, which leads to coarse-grained matching. According to Fig.~\ref{fig:empirical_analysis} (a) at the bottom, transitioning to \textbf{Form (2)} is identified as a practical and appropriate alternative. However, our empirical result indicates that exclusive reliance on \textbf{Form (2)} yields a synthesized dataset that lacks sufficient information density. Consequently, we propose a hybrid method that effectively integrates \textbf{Form (1)} and \textbf{Form (2)} using a weighted average, which we term soft category-aware matching.
\begin{equation}
\fontsize{8pt}{11pt}\selectfont
\begin{aligned}
&\mathcal{L}_\textbf{\textrm{syn}}^\prime =\alpha||p(\mu|\mathcal{X}^\mathcal{S})-p(\mu|\mathcal{X}^\mathcal{T})||_2+||p(\sigma^2|\mathcal{X}^\mathcal{S})-p(\sigma^2|\mathcal{X}^\mathcal{T})||_2\textcolor{C3}{\quad \#\textrm{Form (1)}}\\
&+(1-\alpha)\sum_i^\mathbf{C}p(c_i)\left[||p(\mu|\mathcal{X}^\mathcal{S},c_i)-p(\mu|\mathcal{X}^\mathcal{T},c_i)||_2+||p(\sigma^2|\mathcal{X}^\mathcal{S},c_i)-p(\sigma^2|\mathcal{X}^\mathcal{T},c_i)||_2\right],\textcolor{C3}{\quad \#\textrm{Form (2)}} \\
\end{aligned}
\label{eq:category_aware_matching}
\end{equation}
where $\mathbf{C}$ represents the total number of components, $c_i$ indicates the $i$-th component within a GMM, and $\alpha$ is a coefficient for adjusting the balance. The modified loss function $\mathcal{L}_\textbf{\textrm{syn}}^\prime$ is designed to effectively regulate the information density of $\mathcal{X}^{\mathcal{S}}$ and to align the distribution of $\mathcal{X}^{\mathcal{S}}$ with that of $\mathcal{X}^{\mathcal{T}}$. Operationally, each category in the original dataset is mapped to a distinct component in the GMM framework. Particularly, when $\alpha=1$, the sophisticated category-aware matching described by $\mathcal{L}_\textbf{\textrm{syn}}^\prime$ in Eq.~\ref{eq:category_aware_matching} simplifies to the basic statistical matching defined by $\mathcal{L}_\textbf{\textrm{syn}}$ in Eq.~\ref{eq:definition_sm}.

\begin{theorem}
\label{the:total_category_aware_matching}
(proofs in Theorems~\ref{tho:mean_and_variance_consistent},~\ref{the:entropy},~\ref{the:kl} and Corollary~\ref{cor:same}) Given the original data distribution $p_\textrm{mix}(x)$, and define condensed samples as $x$ and $y$ in \textit{\textbf{Form (1)}} and \textit{\textbf{Form (2)}} with their distributions characterized by $P$ and $Q$. Subsequently, it follows that \textcolor{C3}{(i)} $\mathbb{E}[x]\equiv \mathbb{E}[y]$, \textcolor{C3}{(ii)} $\mathbb{D}[x]\equiv \mathbb{D}[y]$, \textcolor{C3}{(iii)} $\mathcal{H}(P)- \frac{1}{2}\left[\log(\mathbb{E}[\mathbb{D}[y^j]]+\mathbb{D}[\mathbb{E}[y^j]])-\mathbb{E}[\log(\mathbb{D}[y^j])]\right] \leq \mathcal{H}(Q) \leq \mathcal{H}(P)+\frac{1}{4}\mathbb{E}_{(i,j)\sim \prod[\mathbf{C},\mathbf{C}]}\left[\frac{(\mathbb{E}[y^i]-\mathbb{E}[y^j])^2(\mathbb{D}[y^i]+\mathbb{D}[y^j])}{\mathbb{D}[y^i]\mathbb{D}[y^j]}\right]$ and \textcolor{C3}{(iv)} $D_\textrm{KL}[p_\textrm{mix}||P]\leq \mathbb{E}_{i\sim \mathcal{U}[1,\ldots,\mathbf{C}]}\mathbb{E}_{j\sim \mathcal{U}[1,\ldots,\mathbf{C}]}\frac{\mathbb{E}[y^j]^2}{\mathbb{D}[y^i]}$ and $D_\textrm{KL}[p_\textrm{mix}||Q] = 0$.
\end{theorem}

We further analyze the properties of distributions $P$ and $Q$ as in {\textbf{Form (1)}} and {\textbf{Form (2)}}. According to parts \textcolor{C3}{(i)} and \textcolor{C3}{(ii)} of Theorem~\ref{the:total_category_aware_matching}, $Q$ retains the same variance and mean as $P$. Regarding diversity, part \textcolor{C3}{(iii)} of Theorem~\ref{the:total_category_aware_matching} states that the entropy $\mathcal{H}(\cdot)$ of $P$ and $Q$ is equivalent, $\mathcal{H}(P) \equiv \mathcal{H}(Q)$, provided the mean and variance of all components in the GMM are uniform, suggesting a single Gaussian profile. Absent this condition, there is no guarantee that $\mathcal{H}(P)$ and $\mathcal{H}(Q)$ will consistently increase or decrease. These findings underscore the advantages of using GMM, especially when the initial data conforms to an unimodal distribution, thus aligning the mean, variance, and entropy of distributions $P$ and $Q$ in the reduced dataset. Moreover, even in diverse scenarios, the mean, variance, and entropy of $Q$ tend to remain stable. Furthermore, when the original dataset exhibits a more complex bimodal distribution and the parameters of the Gaussian components are precisely estimated, utilizing GMM can effectively reduce the Kullback-Leibler divergence between the mixed original distribution $p_\textrm{mix}$ and $Q$ to near zero. In contrast, the divergence $D_\textrm{KL}[p_\textrm{mix}||P]$ always maintains a non-zero upper bound, as noted in part \textcolor{C3}{(iv)} of Theorem~\ref{the:total_category_aware_matching}. Therefore, by modulating the weight $\alpha$ in Eq.~\ref{eq:category_aware_matching}, we can derive an optimally balanced solution that minimizes loss in data characteristics while maximizing fidelity between the synthesized and original distributions.

\textbf{Flatness Regularization $\left(\imineq{figures/serial_number/data_synthesis_1.pdf}{2.4}\right)$ and EMA-based Evaluation $\left(\imineq{figures/serial_number/post_evaluation_3.pdf}{2.4}\right)$.} Choices $\imineq{figures/serial_number/data_synthesis_1.pdf}{2.4}$ and $\imineq{figures/serial_number/post_evaluation_3.pdf}{2.4}$ are utilized to ensure flat loss landscapes during the stages of data synthesis and post-evaluation, respectively. 

During the data synthesis phase, the use of sharpness-aware minimization (SAM) algorithms is beneficial for reducing the sharpness of the loss landscape, as presented in prior research~\citep{iclr2020_sam,nips2022_sam,sam_llm}. Nonetheless, traditional SAM approaches, as detailed in Eq.~\ref{eq:proof_sam_9} in the Appendix, generally double the computational load due to their two-stage parameter update process. This increase in computational demand is often impractical during data synthesis. Inspired by MESA~\citep{nips2022_sam}, which achieves sharpness-aware training without additional computational overhead through self-distillation, we introduce a lightweight flatness regularization approach for implementing SAM during data synthesis. This method utilizes a teacher dataset, $\mathcal{X}_\textbf{\textrm{EMA}}^\mathcal{S}$, maintained via exponential moving average (EMA). The newly formulated optimization goal aims to obtain a flat loss landscape in the following manner:
\begin{equation}
\footnotesize
\begin{aligned}
\mathcal{L}_\textbf{\textrm{FR}} &= \mathbb{E}_{\mathcal{L}_{\textrm{syn}}\sim \mathbb{S}_\textrm{match}}[\mathcal{L}_\textbf{\textrm{syn}}(\mathcal{X}^{\mathcal{S}},\mathcal{X}^{\mathcal{S}}_\textbf{\textrm{EMA}})],\text{ }\mathcal{X}^{\mathcal{S}}_\textbf{\textrm{EMA}} = \beta \mathcal{X}^{\mathcal{S}}_\textbf{\textrm{EMA}} + (1-\beta)\mathcal{X}^{\mathcal{S}},  \\
\end{aligned}
\label{eq:flatness_regularization}
\end{equation}
where $\beta$ is the weighting coefficient, which is empirically set to 0.99 in our experiments. The detailed derivation of Eq.~\ref{eq:flatness_regularization} is in Appendix~\ref{apd:sharpness}. And the critical theoretical result is articulated as follows:
\begin{theorem}
\label{the:ema_updated_flatness}
(proof in Appendix~\ref{apd:sharpness}) The optimization objective $\mathcal{L}_\textbf{\textrm{FR}}$ can ensure sharpness-aware minimization within a $\rho$-ball for each point along a straight path between $\mathcal{X}^\mathcal{S}$ and $\mathcal{X}^\mathcal{S}_\textbf{\textrm{EMA}}$.
\end{theorem}
This indicates that the primary optimization goal of $\mathcal{L}_\textbf{\textrm{FR}}$ deviates somewhat from that of traditional SAM-based algorithms, which are designed to achieve a flat loss landscape around $\mathcal{X}^\mathcal{S}$. The constraint on flatness needs to ensure that the first-order term of the Taylor expansion equals zero, indicating normal model convergence. However, our exploratory experiments found that despite the good performance of EDC, the loss of statistical matching at the end of data synthesis still fluctuated significantly and did not reach zero. As a result, we choose to apply flatness regularization exclusively to the logits of the observer model, since the cross-entropy loss for these can more straightforwardly reach zero.
\begin{equation}
\footnotesize
\begin{aligned}
\mathcal{L}^{\prime}_\textbf{\textrm{FR}} &=D_\textrm{KL}(\textrm{softmax}(\phi(\mathcal{X}^\mathcal{S})/\tau)||\textrm{softmax}(\phi(\mathcal{X}^\mathcal{S}_\textbf{\textrm{EMA}})/\tau)),\text{ }\mathcal{X}^{\mathcal{S}}_\textbf{\textrm{EMA}} = \beta \mathcal{X}^{\mathcal{S}}_\textbf{\textrm{EMA}} + (1-\beta)\mathcal{X}^{\mathcal{S}},  \\
\end{aligned}
\label{eq:flatness_regularization}
\end{equation}
where $\textrm{softmax}(\cdot)$, $\tau$ and $\phi$ represent the softmax operator, the temperature coefficient and the pre-trained observer model, respectively. As illustrated in Fig.~\ref{fig:empirical_analysis} (c) top, it is evident that $\mathcal{L}^{\prime}_\textbf{\textrm{FR}}$ significantly lowers the Frobenius norm of the Hessian matrix relative to standard training, thus confirming its efficacy in pushing a flatter loss landscape.

In post-evaluation, we observe that a method analogous to $\mathcal{L}^{\prime}_\textbf{\textrm{FR}}$ employing SAM does not lead to appreciable performance improvements. This result is likely due to the limited sample size of the condensed dataset, which hinders the model’s ability to fully converge post-training, thereby undermining the advantages of flatness regularization. Conversely, the integration of an EMA-updated model as the validated model noticeably stabilizes performance variations during evaluations. We term this strategy EMA-based evaluation and apply it across all benchmark experiments.

\begin{table*}[t]
    \centering
    \resizebox{1.0\textwidth}{!}{
    \begin{tabular}{cc|cccc|cc|cc|c}
        \toprule
               \multirow{2}{*}{Dataset}       &   \multirow{2}{*}{IPC}  & \multicolumn{4}{c|}{ResNet-18} & \multicolumn{2}{c|}{ResNet-50} & \multicolumn{2}{c|}{ResNet-101} & \multicolumn{1}{c}{MobileNet-V2} \\ 
        \cmidrule(lr){3-6} \cmidrule(lr){7-8} \cmidrule(lr){9-10} \cmidrule(lr){11-11} 
         &  & SRe$^2$L & G-VBSM & RDED & EDC (Ours) & G-VBSM & EDC (Ours) & RDED & EDC (Ours) & EDC (Ours)\\     \cmidrule(lr){0-2}   \cmidrule(lr){3-6} \cmidrule(lr){7-8} \cmidrule(lr){9-10} \cmidrule(lr){11-11} 
                      & 1   & - & - & 22.9 $\pm$ 0.4 & \graycell 32.6 $\pm$ 0.1 & - & \graycell 30.6 $\pm$ 0.4 & - & \graycell 26.1 $\pm$ 0.2 & \graycell 20.2 $\pm$ 0.4 \\
        CIFAR-10       & 10  & 27.2 $\pm$ 0.4 & 53.5 $\pm$ 0.6 & 37.1 $\pm$ 0.3 & \graycell 79.1 $\pm$ 0.3 & - & \graycell 76.0 $\pm$ 0.3 & - & \graycell 67.1 $\pm$ 0.5 & \graycell 42.0 $\pm$ 0.4 \\
                      & 50  & 47.5 $\pm$ 0.5 & 59.2 $\pm$ 0.4 & 62.1 $\pm$ 0.1 & \graycell 87.0 $\pm$ 0.1 & - & \graycell 86.9 $\pm$ 0.0 & -& \graycell 85.8 $\pm$ 0.1 & \graycell 70.8 $\pm$ 0.2 \\ 
        \midrule
                      & 1   & 2.0 $\pm$ 0.2 & 25.9 $\pm$ 0.5 & 11.0 $\pm$ 0.3 & \graycell 39.7 $\pm$ 0.1 & - & \graycell 36.1 $\pm$ 0.5 & - & \graycell 32.3 $\pm$ 0.3 & \graycell 10.6 $\pm$ 0.3 \\
        CIFAR-100     & 10  & 31.6 $\pm$ 0.5 & 59.5 $\pm$ 0.4 & 42.6 $\pm$ 0.2 & \graycell 63.7 $\pm$ 0.3 & - & \graycell 62.1 $\pm$ 0.1 & - & \graycell 61.7 $\pm$ 0.1 & \graycell 44.3 $\pm$ 0.4 \\
                      & 50  & 49.5 $\pm$ 0.3 & 65.0 $\pm$ 0.5 & 62.6 $\pm$ 0.1 & \graycell 68.6 $\pm$ 0.2 & - & \graycell 69.4 $\pm$ 0.3 & - & \graycell 68.5 $\pm$ 0.1 & \graycell 59.5 $\pm$ 0.1 \\ 
        \midrule
                      & 1   & - & - & 9.7 $\pm$ 0.4 & \graycell 39.2 $\pm$ 0.4 & - & \graycell 35.9 $\pm$ 0.2 & 3.8 $\pm$ 0.1 & \graycell 40.6 $\pm$ 0.3 & \graycell 18.8 $\pm$ 0.1 \\
        Tiny-ImageNet & 10  & - & - & 41.9 $\pm$ 0.2 & \graycell 51.2 $\pm$ 0.5 & - & \graycell 50.2 $\pm$ 0.3 & 22.9 $\pm$ 3.3 & \graycell 51.6 $\pm$ 0.2 & \graycell 40.6 $\pm$ 0.6 \\
                      & 50  & 41.1 $\pm$ 0.4 & 47.6 $\pm$ 0.3 & \graycell 58.2 $\pm$ 0.1 & 57.2 $\pm$ 0.2 & 48.7 $\pm$ 0.2 & \graycell 58.8 $\pm$ 0.4 & 41.2 $\pm$ 0.4 & \graycell 58.6 $\pm$ 0.1 & \graycell 50.7 $\pm$ 0.1 \\
        \midrule
                      & 1   & - & - & 24.9 $\pm$ 0.5 & \graycell 45.2 $\pm$ 0.2 & - & \graycell 38.2 $\pm$ 0.1 & 21.7 $\pm$ 1.3 & \graycell 36.4 $\pm$ 0.1 & \graycell 36.4 $\pm$ 0.3 \\
        ImageNet-10  & 10  & - & - & 53.3 $\pm$ 0.1 & \graycell 63.4 $\pm$ 0.2 & - & \graycell 62.4 $\pm$ 0.1 & 45.5 $\pm$ 1.7 & \graycell 59.8 $\pm$ 0.1 & \graycell 54.2 $\pm$ 0.1 \\
                      & 50  & - & - & 75.5 $\pm$ 0.5 & \graycell 82.2 $\pm$ 0.1 & - & \graycell 80.8 $\pm$ 0.2 & 71.4 $\pm$ 0.2 & \graycell 80.8 $\pm$ 0.0 & \graycell 80.2 $\pm$ 0.2 \\
        \midrule
                      & 1   & - & - & 6.6 $\pm$ 0.2 & \graycell 12.8 $\pm$ 0.1 & - & \graycell 13.3 $\pm$ 0.3 & 5.9 $\pm$ 0.4 & \graycell 12.2 $\pm$ 0.2 & \graycell 8.4 $\pm$ 0.3 \\
        ImageNet-1k   & 10  & 21.3 $\pm$ 0.6 & 31.4 $\pm$ 0.5 & 42.0 $\pm$ 0.1 & \graycell 48.6 $\pm$ 0.3 & 35.4 $\pm$ 0.8 & \graycell 54.1 $\pm$ 0.2 & 48.3 $\pm$ 1.0 & \graycell 51.7 $\pm$ 0.3 & \graycell 45.0 $\pm$ 0.2 \\
                      & 50  & 46.8 $\pm$ 0.2 & 51.8 $\pm$ 0.4 & 56.5 $\pm$ 0.1 & \graycell 58.0 $\pm$ 0.2 & 58.7 $\pm$ 0.3 & \graycell 64.3 $\pm$ 0.2 & 61.2 $\pm$ 0.4 & \graycell 64.9 $\pm$ 0.2 & \graycell 57.8 $\pm$ 0.1 \\
        \bottomrule
    \end{tabular}
    }
    \vspace{-5pt}
    \caption{\small \textbf{Comparison with the SOTA baseline dataset condensation methods.} SRe$^2$L and RDED utilize ResNet-18 for data synthesis, whereas G-VBSM and EDC leverage various backbones for this purpose.}
    \label{tab:main}
    \vspace{-9pt}
\end{table*}

\begin{table}[t]
\vspace{0pt}
\centering
\resizebox{1.0\textwidth}{!}{%
\begin{tabular}{c|ccccccccccc}
\toprule
IPC &  Method & ResNet-18 & ResNet-50 & ResNet-101 & MobileNet-V2 & EfficientNet-B0 & DeiT-Tiny & Swin-Tiny & ConvNext-Tiny & ShuffleNet-V2 \\
\midrule
\multirow{3}{*}{10} & RDED & 42.0 & 46.0 & 48.3 & 34.4 & 42.8 & 14.0 & 29.2 & 48.3 & 19.4 \\
& EDC (Ours) & 48.6 & 54.1 & 51.7 & 45.0 & 51.1 & 18.4 & 38.3 & 54.4 & 29.8 \\
& \graycell$+\Delta$ & \graycell6.6 & \graycell8.1 & \graycell3.4 & \graycell10.6 & \graycell8.3 & \graycell4.4 & \graycell9.1 & \graycell6.1 & \graycell10.4 \\\hline
\multirow{3}{*}{20} & RDED & 45.6 & 57.6 & 58.0 & 41.3 & 48.1 & 22.1 & 44.6 & 54.0 & 20.7 \\
& EDC (Ours) & 52.0 & 58.2 & 60.0 & 48.6 & 55.6 & 24.0 & 49.6 & 61.4 & 33.0 \\
& \graycell$+\Delta$ & \graycell6.4 & \graycell0.6 & \graycell2.0 & \graycell7.3 & \graycell7.5 & \graycell1.9 & \graycell5.0 & \graycell7.4 & \graycell12.3 \\\hline
\multirow{3}{*}{30} & RDED & 49.9 & 59.4 & 58.1 & 44.9 & 54.1 & 30.5 & 47.7 & 62.1 & 23.5 \\
& EDC (Ours) & 55.0 & 61.5 & 60.3 & 53.8 & 58.4 & 46.5 & 59.1 & 63.9 & 41.1 \\
& \graycell$+\Delta$ & \graycell5.1 & \graycell2.1 & \graycell2.2 & \graycell8.9 & \graycell4.3 & \graycell16.0 & \graycell11.4 & \graycell1.8 & \graycell17.6 \\\hline
\multirow{3}{*}{40} & RDED & 53.9 & 61.8 & 60.1 & 50.3 & 56.3 & 43.7 & 58.1 & 63.7 & 27.7 \\
& EDC (Ours) & 56.4 & 62.2 & 62.3 & 54.7 & 59.7 & 51.9 & 61.1 & 65.2 & 44.7 \\
& \graycell$+\Delta$ & \graycell2.5 & \graycell0.4 & \graycell2.2 & \graycell4.4 & \graycell3.4 & \graycell8.2 & \graycell3.0 & \graycell1.5 & \graycell17.0 \\\hline
\multirow{3}{*}{50} & RDED & 56.5 & 63.7 & 61.2 & 53.9 & 57.6 & 44.5 & 56.9 & 65.4 & 30.9 \\
& EDC (Ours) & 58.0 & 64.3 & 64.9 & 57.8 & 60.9 & 55.0 & 63.3 & 66.6 & 45.7 \\
& \graycell$+\Delta$ & \graycell1.5 & \graycell0.6 & \graycell3.7 & \graycell3.9 & \graycell3.3 & \graycell10.5 & \graycell6.4 & \graycell1.2 & \graycell14.8 \\
\bottomrule
\end{tabular}
}
\vspace{2pt}
\caption{\textbf{Cross-architecture generalization comparison with different IPCs on ImageNet-1k.} RDED refers to the latest SOTA method on ImageNet-1k and $+\Delta$ stands for the improvement for each architecture.}
\label{tab:cross_arch_generalization}
\vspace{-18pt}
\end{table}

\textbf{Smoothing Learning Rate (LR) Schedule $\left(\imineq{figures/serial_number/post_evaluation_2.pdf}{2.4}\right)$ and Smaller Batch Size $\left(\imineq{figures/serial_number/soft_label_1.pdf}{2.4}\imineq{figures/serial_number/post_evaluation_1.pdf}{2.4}\right)$.} Here, we introduce two effective strategies for post-evaluation training. Firstly, it is crucial to clarify and distinguish between standard or conventional deep model training and post-evaluation in the context of dataset condensation. Specifically, (1) in dataset condensation, the limited number of samples in $\mathcal{X}^\mathcal{S}$ results in fewer training iterations per epoch, typically leading to underfitting; and (2) the gradient of a random batch from $\mathcal{X}^\mathcal{S}$ aligns more closely with the global gradient than that from a random batch in $\mathcal{X}^\mathcal{T}$. To support the latter observation, we utilize a ResNet-18 model with randomly initialized parameters to calculate the gradient of a random batch and assess the cosine similarity with the global gradient of $\mathcal{X}^\mathcal{T}$. After conducting over 100 iterations of this procedure, the average cosine similarity is consistently higher between $\mathcal{X}^\mathcal{S}$ and the global gradient than with $\mathcal{X}^\mathcal{T}$, indicating a greater similarity and reduced sensitivity to batch size fluctuations. Our findings further illustrate that the gradient from a random batch in $\mathcal{X}^\mathcal{S}$ effectively approximates the global gradient, as shown in Fig.~\ref{fig:empirical_analysis} (c) bottom. Given this, the inaccurate gradient direction problem introduced by the small batch size becomes less problematic. Instead, using a small batch size effectively increases the number of iterations, thereby helping prevent model under-convergence.

To optimize the training with condensed samples, we implement a smoothed LR schedule that moderates the learning rate reduction throughout the training duration. This approach helps avoid early convergence to suboptimal minima, thereby enhancing the model's generalization capabilities. The mathematical formulation of this schedule is given by $\mu(i) = \frac{1+\textrm{cos}(i\pi/\zeta N)}{2}$, where $i$ represents the current epoch, $N$ is the total number of epochs, $\mu(i)$ is the learning rate for the $i$-th epoch, and $\zeta$ is the deceleration factor. Notably, a $\zeta$ value of 1 corresponds to a typical cosine learning rate schedule, whereas setting $\zeta$ to 2 improves performance metrics from 34.4\% to 38.7\% and effectively moderates loss landscape sharpness during post-evaluation.

\textbf{Weak Augmentation $\left(\imineq{figures/serial_number/data_synthesis_4.pdf}{2.4}\right)$ and Better Backbone Choice $\left(\imineq{figures/serial_number/soft_label_2.pdf}{2.4}\right)$.} The principal role of these two design decisions is to address the flawed settings in the \textit{baseline} G-VBSM. The key finding reveals that the minimum area threshold for cropping during data synthesis was overly restrictive, thereby diminishing the quality of the condensed dataset. To rectify this, we implement mild augmentations to increase this minimum cropping threshold, thereby improving the dataset condensation's ability to generalize. Additionally, we substitute the computationally demanding EfficientNet-B0 with more streamlined AlexNet for generating soft labels on ImageNet-1k, a change we refer to as an improved backbone selection. This modification maintains the performance without degradation. More details on the ablation studies for mild augmentation and improved backbone selection are in Appendix~\ref{apd:add_ab_experiment}.
\vspace{-0.07in}
\section{Experiments}
To validate the effectiveness of our proposed EDC, we conduct comparative experiments across various datasets, including ImageNet-1k~\citep{ILSVRC15}, ImageNet-10~\citep{dd_efficient_parameterization}, Tiny-ImageNet~\citep{tiny_imagenet}, CIFAR-100~\citep{CIFAR}, and CIFAR-10~\citep{CIFAR}. Additionally, we explore cross-architecture generalization and ablation studies on ImageNet-1k. All experiments are conducted using 4$\times$ RTX 4090 GPUs. Due to space constraints, detailed descriptions of the hyperparameter settings, additional ablation studies, and visualizations of synthesized images are provided in the Appendix~\ref{tab:hyperparameter_settings}, ~\ref{apd:add_ab_experiment}, and ~\ref{apd:syn_image_vis}, respectively.

\textbf{Network Architectures.} Following prior dataset condensation work~\citep{dd_sre2l,yin2024dataset,shao2023generalized,RDED}, our comparison uses ResNet-\{18, 50, 101\}~\citep{ResNet} as our verified models. We also extend our evaluation to include MobileNet-V2~\citep{mobilenetv2} in Table~\ref{tab:main} and explore cross-architecture generalization further with recently advanced backbones such as DeiT-Tiny~\citep{deit} and Swin-Tiny~\citep{SWIN-T} (detailed in Table~\ref{tab:cross_arch_generalization}).

\begin{table}[t!]
\centering
\resizebox{1.0\textwidth}{!}{
\footnotesize
\begin{tabular}{lc|ccccl|ccc}
\cmidrule(lr){1-5} \cmidrule(lr){7-10}
Design Choices & $\zeta$ & ResNet-18 & ResNet-50 & ResNet-101 & & Design Choices & ResNet-18 & ResNet-50 & ResNet-101 \\
\cmidrule(lr){1-5} \cmidrule(lr){7-10}
\config{C} & 1.0 & 34.4 & 36.8 & 42.0 & & RDED & 25.8 & 32.7 & 34.8 \\
\config{C} & 1.5 & 38.7 & 42.0 & 46.3 & & RDED+$\left(\imineq{figures/serial_number/soft_label_1.pdf}{2.4}\imineq{figures/serial_number/post_evaluation_1.pdf}{2.4}\imineq{figures/serial_number/post_evaluation_2.pdf}{2.4}\right)$  & 42.3 & 48.4 & 47.0 \\
\config{C} & 2.0 & 38.8 & 45.8 & \graycell 47.9 & & G-VBSM+$\left(\imineq{figures/serial_number/data_synthesis_3.pdf}{2.4}\right)$ & 34.4 & 36.8 & 42.0 \\
\config{C} & 2.5 & \graycell 39.0 & 44.6 & 46.0 & & G-VBSM+$\left(\imineq{figures/serial_number/data_synthesis_3.pdf}{2.4}\imineq{figures/serial_number/post_evaluation_2.pdf}{2.4}\right)$ & 38.8 & 45.8 & 47.9 \\
\config{C} & 3.0 & 38.8 & \graycell 45.6 & 46.2 & &G-VBSM+$\left(\imineq{figures/serial_number/data_synthesis_3.pdf}{2.4}\imineq{figures/serial_number/post_evaluation_2.pdf}{2.4}\imineq{figures/serial_number/soft_label_1.pdf}{2.4}\imineq{figures/serial_number/post_evaluation_1.pdf}{2.4}\right)$ & \hspace{-2pt}\cellcolor{gray!25} 45.0 & \hspace{-2pt}\cellcolor{gray!25} 51.6 & \hspace{-2pt}\cellcolor{gray!25} 48.1 \\\cline{1-5}\cline{7-10}
\end{tabular}}\vspace{2pt}
\caption{\textbf{Ablation studies on ImageNet-1k with IPC 10.} \textbf{Left:} Explore the influence of the slowdown coefficient $\zeta$ with \config{C}. \textbf{Right:} Evaluate the effectiveness of real image initialization $\left(\imineq{figures/serial_number/data_synthesis_3.pdf}{2.4}\right)$, smoothing LR schedule $\left(\imineq{figures/serial_number/post_evaluation_2.pdf}{2.4}\right)$ and smaller batch size $\left(\imineq{figures/serial_number/soft_label_1.pdf}{2.4}\imineq{figures/serial_number/post_evaluation_1.pdf}{2.4}\right)$ with $\zeta=2$.}
\label{tab:imagenet_scheduler}
\vspace{-8pt}
\end{table}
\begin{table}[t!]
\centering
\resizebox{0.95\textwidth}{!}{
\footnotesize
\renewcommand\arraystretch{0.7}
\begin{tabular}{lccccc|ccc}
\toprule
Design Choices & Loss Type & Loss Weight & $\zeta$ & $\beta$ & $\tau$ & ResNet-18 & ResNet-50 & DenseNet-121 \\\midrule
\config{C} & - & - & 1.5 & - & - & 38.7 & 42.0 & 40.6 \\
\config{D} & $\mathcal{L}_\textbf{\textrm{FR}}$ & 0.025 & 1.5 & 0.999 & 4 & 38.8 & 43.2 & 40.3 \\
\config{D} & $\mathcal{L}_\textbf{\textrm{FR}}$ & 0.25 & 1.5 & 0.999 & 4 & 37.9 & 43.5 & 40.3 \\
\config{D} & $\mathcal{L}_\textbf{\textrm{FR}}$ & 2.5 & 1.5 & 0.999 & 4 & 31.7 & 37.0 & 32.9 \\
\config{D} & $\mathcal{L}_\textbf{\textrm{FR}}$ & 0.25 & 1.5 & 0.99 & 4 & 39.0 & 43.3 & 40.2 \\
\config{D} & $\mathcal{L}^\prime_\textbf{\textrm{FR}}$ & 0.25 & 1.5 & 0.99 & 4 & \graycell 39.5 & \graycell 44.1 & \graycell 41.9 \\
\config{D} & $\mathcal{L}^\prime_\textbf{\textrm{FR}}$ & 0.25 & 1.5 & 0.99 & 1 & 38.9 & 43.5 & 40.7 \\
\config{D} & vanilla SAM & 0.25 & 1.5 & - & - & 38.8 & 44.0 & 41.2 \\\bottomrule
\end{tabular}}\vspace{2pt}
\caption{\textbf{Ablation studies on ImageNet-1k with IPC 10.} Investigate the potential effects of several factors, including loss type, loss weight, $\beta$, and $\tau$, amid flatness regularization $\left(\imineq{figures/serial_number/data_synthesis_1.pdf}{2.4}\right)$.}
\label{tab:flatness}
\vspace{-8pt}
\end{table}
\begin{table}[t!]
\centering
\resizebox{0.95\textwidth}{!}{
\footnotesize
\renewcommand\arraystretch{0.8}
\begin{tabular}{lcccccccc}
\toprule
Design Choices & $\alpha$ & $\zeta$ & \makecell{Weak Augmentation\\Scale=(0.5,1.0)} & \makecell{EMA-based Evaluation\\EMA Rate=0.99} & ResNet-18 & ResNet-50 & ResNet-101 \\\midrule
\config{F} & 0.00 & 2.0 & \xmark & \xmark & 46.2 & 53.2 & 49.5 \\
\config{F} & 0.00 & 2.0 & \cmark & \xmark & 46.7 & 53.7 & 49.4 \\
\config{F} & 0.00 & 2.0 & \cmark & \cmark & 46.9 & 53.8 & 48.5 \\
\config{F} & 0.25 & 2.0 & \xmark & \xmark & 46.7 & 53.4 & 50.6 \\
\config{F} & 0.25 & 2.0 & \cmark & \xmark & 46.8 & 53.6 & 50.8 \\
\config{F} & 0.25 & 2.0 & \cmark & \cmark & 47.1 & 53.7 & 48.2 \\
\config{F} & 0.50 & 2.0 & \xmark & \xmark & 48.1 & 53.9 & 50.4 \\
\config{F} & 0.50 & 2.0 & \cmark & \xmark & 48.4 & 53.9 & \graycell 52.7 \\
\config{F} & 0.50 & 2.0 & \cmark & \cmark & \graycell 48.6 & \graycell 54.1 & 51.7 \\
\config{F} & 0.75 & 2.0 & \xmark & \xmark & 46.1 & 52.7 & 51.0 \\
\config{F} & 0.75 & 2.0 & \cmark & \xmark & 46.9 & 52.8 & 51.6 \\
\config{F} & 0.75 & 2.0 & \cmark & \cmark & 47.0 & 53.2 & 49.3 \\
\bottomrule
\end{tabular}}\vspace{2pt}
\caption{\textbf{Ablation studies on ImageNet-1k with IPC 10.} Evaluate the effectiveness of several design choices, including soft category-aware matching $\left(\imineq{figures/serial_number/data_synthesis_2.pdf}{2.4}\right)$, weak augmentation $\left(\imineq{figures/serial_number/data_synthesis_4.pdf}{2.4}\right)$ and EMA-based evaluation $\left(\imineq{figures/serial_number/post_evaluation_3.pdf}{2.4}\right)$.}
\label{tab:cate_aware_matching}
\vspace{-18pt}
\end{table}

\textbf{Baselines.} We compare our work with several recent state-of-the-art methods, including SRe$^2$L~\citep{dd_sre2l}, G-VBSM~\citep{shao2023generalized}, and RDED~\citep{RDED} to assess broader practical impacts. It is important to note that we have omitted several traditional methods~\citep{dd_mtt,dd_dream,dd_tesla} from our analysis. This exclusion is due to their inadequate performance on the large-scale ImageNet-1k and their lesser effectiveness when applied to practical networks such as ResNet, MobileNet-V2, and Swin-Tiny~\citep{SWIN-T}. For instance, the MTT method~\citep{dd_mtt} encounters an out-of-memory issue on ImageNet-1k, and ResNet-18 achieves only a 46.4\% accuracy on CIFAR-10 with IPC 10, which is significantly lower than the 79.1\% accuracy reported for our EDC in Table~\ref{tab:main}.

\subsection{Main Results}

\textbf{Experimental Comparison.} Our integral EDC, represented as \config{G} in Fig.~\ref{figure:illustration}, provides a versatile solution that outperforms other approaches across various dataset sizes. The results in Table~\ref{tab:main} affirm its ability to consistently deliver substantial performance gains across different IPCs, datasets, and model architectures. Particularly notable is the performance leap in the highly compressed IPC 1 scenario using ResNet-18, where EDC markedly outperforms the latest state-of-the-art method, RDED. Performance rises from 22.9\%, 11.0\%, 7.0\%, 24.9\%, and 6.6\% to 32.6\%, 39.7\%, 39.2\%, 45.2\%, and 12.8\% for CIFAR-10, CIFAR-100, Tiny-ImageNet, ImageNet-10, and ImageNet-1k, respectively. These improvements clearly highlight EDC's superior information encapsulation and enhanced generalization capability, attributed to the efficiently synthesized condensed dataset.

\textbf{Cross-Architecture Generalization.} To verify the generalization ability of our condensed datasets, it is essential to assess their performance across various architectures such as ResNet-\{18, 50, 101\}~\citep{ResNet}, MobileNet-V2~\citep{mobilenetv2}, EfficientNet-B0~\citep{efficientnet}, DeiT-Tiny~\citep{deit}, Swin-Tiny~\citep{SWIN-T}, ConvNext-Tiny~\citep{convnext} and ShuffleNet-V2~\citep{shufflenet}. The results of these evaluations are presented in Table~\ref{tab:cross_arch_generalization}. During cross-validation that includes all IPCs and the mentioned architectures, our EDC consistently achieves higher accuracy than RDED, demonstrating its strong generalization capabilities. Specifically, EDC surpasses RDED by significant margins of 8.2\% and 14.42\% on DeiT-Tiny and ShuffleNet-V2, respectively.

\begin{wrapfigure}{r}{7.5cm}
\vspace{-10pt}
\includegraphics[height=0.23\textwidth,trim={0cm 0cm 0cm 0cm},clip]{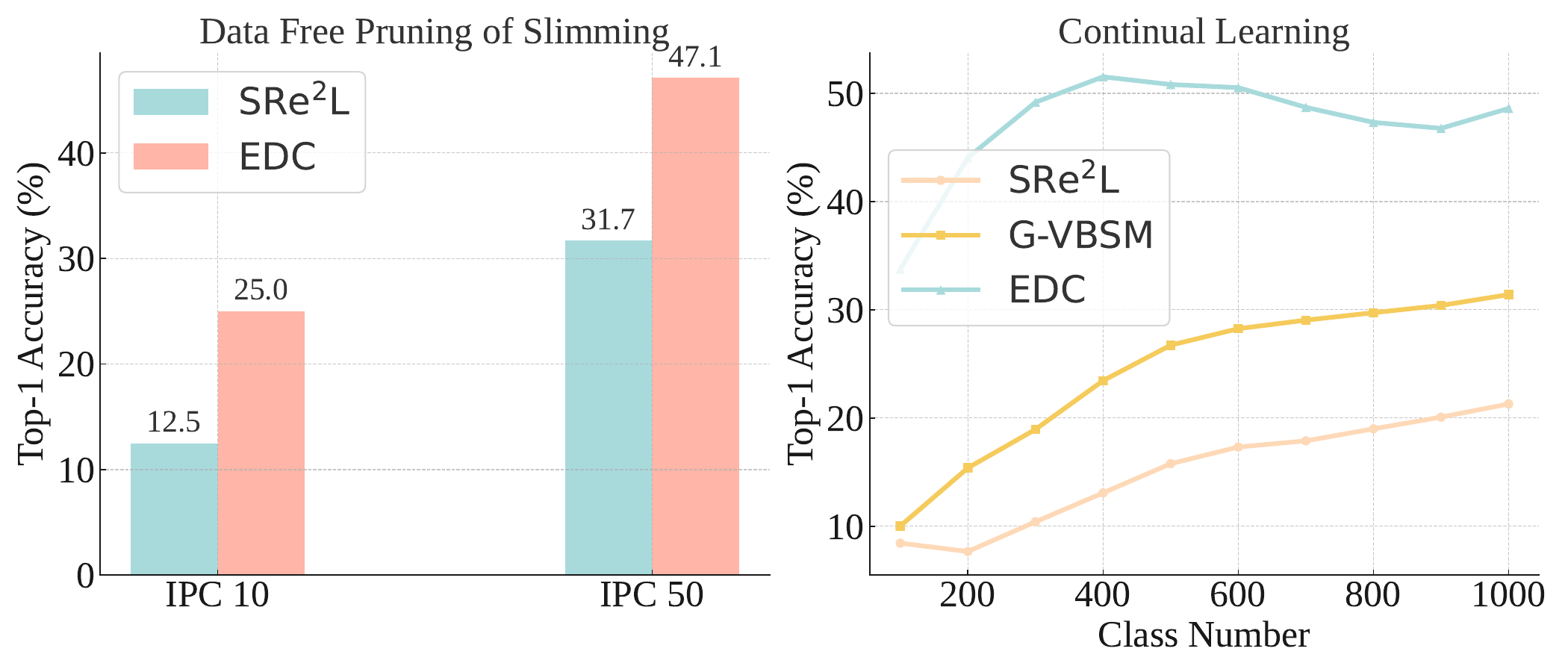}
\vspace{-15pt}
\caption{\textbf{Application on ImageNet-1k.} We evaluate the effectiveness of data-free network slimming and continual learning using VGG11-BN and ResNet-18, respectively.}
\label{fig:application}
\vspace{-12pt}
\end{wrapfigure}\textbf{Application.} Our condensed dataset not only serves as a versatile training resource but also enhances the adaptability of models across various downstream tasks. We demonstrate its effectiveness by employing it in scenarios such as data-free network slimming~\citep{liu2017learning} (\textit{w.r.t.,} parameter pruning~\citep{srinivas2015data}) and class-incremental continual learning~\citep{Prabhu_Torr_Dokania_2020} outlined in DM~\citep{dd_dist_matching}. Fig.~\ref{fig:application} shows the wide applicability of our condensed dataset in both data-free network slimming and class-incremental continual learning. It substantially outperforms SRe$^2$L and G-VBSM, achieving significantly better results.

\subsection{Ablation Studies}

\textbf{Real Image Initialization $\left(\imineq{figures/serial_number/data_synthesis_3.pdf}{2.4}\right)$, Smoothing LR Schedule $\left(\imineq{figures/serial_number/post_evaluation_2.pdf}{2.4}\right)$ and Smaller Batch Size $\left(\imineq{figures/serial_number/soft_label_1.pdf}{2.4}\imineq{figures/serial_number/post_evaluation_1.pdf}{2.4}\right)$.} As shown in Table~\ref{tab:imagenet_scheduler} (left), these design choices, with zero additional computational cost, sufficiently enhance the performance of both G-VBSM and RDED. Furthermore, we investigate the influence of $\zeta$ within smoothing LR schedule in Table~\ref{tab:imagenet_scheduler} (right), concluding that a smoothing learning rate decay is worthwhile for the condensed dataset's generalization ability and the optimal $\zeta$ is model-dependent.

\textbf{Flatness Regularization $\left(\imineq{figures/serial_number/data_synthesis_1.pdf}{2.4}\right)$.} The results in Table~\ref{tab:flatness} demonstrate the effectiveness of flatness regularization, while requiring a well-designed setup. Specifically, attempting to minimize sharpness across all statistics (\textit{i.e.,} $\mathcal{L}_\textbf{FR}$) proves ineffective, instead, it is more effective to apply this regularization exclusively to the logit (\textit{i.e.,} $\mathcal{L}^\prime_\textbf{FR}$). Setting the loss weights $\beta$ and $\tau$ at 0.25, 0.99, and 4, respectively, yields the best accuracy of 39.5\%, 44.1\%, and 45.9\% for ResNet-18, ResNet-50, and DenseNet-121. Moreover, our design of $\mathcal{L}^\prime_\textbf{FR}$ surpasses the performance of the vanilla SAM, while requiring only half the computational resources.

\textbf{Soft Category-Aware Matching $\left(\imineq{figures/serial_number/data_synthesis_2.pdf}{2.4}\right)$, Weak Augmentation $\left(\imineq{figures/serial_number/data_synthesis_4.pdf}{2.4}\right)$ and EMA-based Evaluation $\left(\imineq{figures/serial_number/post_evaluation_3.pdf}{2.4}\right)$.} Table~\ref{tab:cate_aware_matching} illustrates the effectiveness of weak augmentation and EMA-based evaluation, with EMA evaluation also playing a crucial role in minimizing performance fluctuations during assessment. The evaluation of soft category-aware matching primarily involves exploring the effect of parameter $\alpha$ across the range $[0,1]$. The results in Table~\ref{tab:cate_aware_matching} suggest that setting $\alpha$ to 0.5 yields the best results based on our empirical analysis. This finding not only confirms the utility of soft category-aware matching but also emphasizes the importance of ensuring that the condensed dataset maintains a high level of information density and bears a distributional resemblance to the original dataset.
\vspace{-5pt}
\section{Conclusion}
In this paper, we have conducted an extensive exploration and analysis of the design possibilities for scalable dataset condensation techniques. This comprehensive investigation helped us pinpoint a variety of effective and flexible design options, ultimately leading to the construction of a novel framework, which we call EDC. We have extensively examined EDC across five different datasets, which vary in size and number of classes, effectively proving EDC's robustness and scalability. Our results suggest that previous dataset distillation methods have not yet reached their full potential, largely due to suboptimal design decisions. We aim for our findings to motivate further research into developing algorithms capable of efficiently managing datasets of diverse sizes, thus advancing the field of dataset condensation task.

{\small
\bibliographystyle{IEEEtranN}
\bibliography{egbib}

\begin{thebibliography}{52}
\providecommand{\natexlab}[1]{#1}
\providecommand{\url}[1]{#1}
\csname url@samestyle\endcsname
\providecommand{\newblock}{\relax}
\providecommand{\bibinfo}[2]{#2}
\providecommand{\BIBentrySTDinterwordspacing}{\spaceskip=0pt\relax}
\providecommand{\BIBentryALTinterwordstretchfactor}{4}
\providecommand{\BIBentryALTinterwordspacing}{\spaceskip=\fontdimen2\font plus
\BIBentryALTinterwordstretchfactor\fontdimen3\font minus \fontdimen4\font\relax}
\providecommand{\BIBforeignlanguage}[2]{{%
\expandafter\ifx\csname l@#1\endcsname\relax
\typeout{** WARNING: IEEEtranN.bst: No hyphenation pattern has been}%
\typeout{** loaded for the language `#1'. Using the pattern for}%
\typeout{** the default language instead.}%
\else
\language=\csname l@#1\endcsname
\fi
#2}}
\providecommand{\BIBdecl}{\relax}
\BIBdecl

\bibitem[He et~al.(2016{\natexlab{a}})He, Zhang, and Ren]{ResNet}
K.~He, X.~Zhang, and S.~Ren, ``Deep residual learning for image recognition,'' in \emph{Computer Vision and Pattern Recognition}.\hskip 1em plus 0.5em minus 0.4em\relax Las Vegas, NV, USA: IEEE, Jun. 2016, pp. 770--778.

\bibitem[He et~al.(2016{\natexlab{b}})He, Zhang, Ren, and Sun]{ResNetv2}
K.~He, X.~Zhang, S.~Ren, and J.~Sun, ``Identity mappings in deep residual networks,'' in \emph{European Conference on Computer Vision}.\hskip 1em plus 0.5em minus 0.4em\relax Amsterdam, North Holland, The Netherlands: Springer, Oct. 2016, pp. 630--645.

\bibitem[Brown et~al.(2020)Brown, Mann, Ryder, Subbiah, Kaplan, Dhariwal, Neelakantan, Shyam, Sastry, Askell, et~al.]{GPT3}
T.~Brown, B.~Mann, N.~Ryder, M.~Subbiah, J.~D. Kaplan, P.~Dhariwal, A.~Neelakantan, P.~Shyam, G.~Sastry, A.~Askell \emph{et~al.}, ``Language models are few-shot learners,'' \emph{Advances in neural information processing systems}, vol.~33, pp. 1877--1901, 2020.

\bibitem[Dosovitskiy et~al.(2020)Dosovitskiy, Beyer, Kolesnikov, Weissenborn, Zhai, Unterthiner, Dehghani, Minderer, Heigold, Gelly, et~al.]{VIT}
A.~Dosovitskiy, L.~Beyer, A.~Kolesnikov, D.~Weissenborn, X.~Zhai, T.~Unterthiner, M.~Dehghani, M.~Minderer, G.~Heigold, S.~Gelly \emph{et~al.}, ``An image is worth 16x16 words: Transformers for image recognition at scale,'' in \emph{International Conference on Learning Representations}.\hskip 1em plus 0.5em minus 0.4em\relax Event Virtual: OpenReview.net, May 2020.

\bibitem[Shao et~al.(2024)Shao, Shen, Gong, Chen, and Dai]{FM_KT}
S.~Shao, Z.~Shen, L.~Gong, H.~Chen, and X.~Dai, ``Precise knowledge transfer via flow matching,'' \emph{arXiv preprint arXiv:2402.02012}, 2024.

\bibitem[Masarczyk and Tautkute(2020)]{dd_continual_learning_1}
W.~Masarczyk and I.~Tautkute, ``Reducing catastrophic forgetting with learning on synthetic data,'' in \emph{Computer Vision and Pattern Recognition Workshops}.\hskip 1em plus 0.5em minus 0.4em\relax Virtual Event: {IEEE}, Jun. 2020, pp. 252--253.

\bibitem[Sangermano et~al.(2022)Sangermano, Carta, Cossu, and Bacciu]{dd_continual_learning_2}
M.~Sangermano, A.~Carta, A.~Cossu, and D.~Bacciu, ``Sample condensation in online continual learning,'' in \emph{International Joint Conference on Neural Networks}.\hskip 1em plus 0.5em minus 0.4em\relax Padua, Italy: {IEEE}, Jul. 2022, pp. 1--8.

\bibitem[Zhao and Bilen(2021)]{dd_continual_learning_3}
B.~Zhao and H.~Bilen, ``Dataset condensation with differentiable siamese augmentation,'' in \emph{International Conference on Machine Learning}, M.~Meila and T.~Zhang, Eds., vol. 139.\hskip 1em plus 0.5em minus 0.4em\relax Virtual Event: {PMLR}, 2021, pp. 12\,674--12\,685.

\bibitem[Such et~al.(2020)Such, Rawal, Lehman, Stanley, and Clune]{dd_nas_1}
F.~P. Such, A.~Rawal, J.~Lehman, K.~O. Stanley, and J.~Clune, ``Generative teaching networks: Accelerating neural architecture search by learning to generate synthetic training data,'' in \emph{International Conference on Machine Learning}, vol. 119.\hskip 1em plus 0.5em minus 0.4em\relax Virtual Event: {PMLR}, Jul. 2020, pp. 9206--9216.

\bibitem[Zhao and Bilen(2023)]{dd_dist_matching}
B.~Zhao and H.~Bilen, ``Dataset condensation with distribution matching,'' in \emph{Winter Conference on Applications of Computer Vision}.\hskip 1em plus 0.5em minus 0.4em\relax Waikoloa, Hawaii: {IEEE}, Jan. 2023, pp. 6514--6523.

\bibitem[Zhao et~al.(2021)Zhao, Mopuri, and Bilen]{dd_gradient_matching}
B.~Zhao, K.~R. Mopuri, and H.~Bilen, ``Dataset condensation with gradient matching,'' in \emph{International Conference on Learning Representations}.\hskip 1em plus 0.5em minus 0.4em\relax Virtual Event: OpenReview.net, May 2021.

\bibitem[Liu et~al.(2017)Liu, Li, Shen, Huang, Yan, and Zhang]{liu2017learning}
Z.~Liu, J.~Li, Z.~Shen, G.~Huang, S.~Yan, and C.~Zhang, ``Learning efficient convolutional networks through network slimming,'' in \emph{International Conference on Computer Vision}.\hskip 1em plus 0.5em minus 0.4em\relax IEEE, 2017, pp. 2736--2744.

\bibitem[Cazenavette et~al.(2022)Cazenavette, Wang, Torralba, Efros, and Zhu]{dd_mtt}
G.~Cazenavette, T.~Wang, A.~Torralba, A.~A. Efros, and J.~Zhu, ``Dataset distillation by matching training trajectories,'' in \emph{Computer Vision and Pattern Recognition}.\hskip 1em plus 0.5em minus 0.4em\relax New Orleans, LA, USA: {IEEE}, Jun. 2022.

\bibitem[Sajedi et~al.(2023)Sajedi, Khaki, Amjadian, Liu, Lawryshyn, and Plataniotis]{dd_datadam}
A.~Sajedi, S.~Khaki, E.~Amjadian, L.~Z. Liu, Y.~A. Lawryshyn, and K.~N. Plataniotis, ``Datadam: Efficient dataset distillation with attention matching,'' in \emph{International Conference on Computer Vision}.\hskip 1em plus 0.5em minus 0.4em\relax Paris, France: {IEEE}, Oct. 2023, pp. 17\,097--17\,107.

\bibitem[Liu et~al.(2023{\natexlab{a}})Liu, Gu, Wang, Zhu, Jiang, and You]{dd_dream}
Y.~Liu, J.~Gu, K.~Wang, Z.~Zhu, W.~Jiang, and Y.~You, ``{DREAM:} efficient dataset distillation by representative matching,'' \emph{arXiv preprint arXiv:2302.14416}, 2023.

\bibitem[Russakovsky et~al.(2015)Russakovsky, Deng, Su, Krause, Satheesh, Ma, Huang, Karpathy, Khosla, Bernstein, et~al.]{ILSVRC15}
O.~Russakovsky, J.~Deng, H.~Su, J.~Krause, S.~Satheesh, S.~Ma, Z.~Huang, A.~Karpathy, A.~Khosla, M.~Bernstein \emph{et~al.}, ``Imagenet large scale visual recognition challenge,'' \emph{International Journal of Computer Vision}, vol. 115, no.~3, pp. 211--252, 2015.

\bibitem[Yin et~al.(2023)Yin, Xing, and Shen]{dd_sre2l}
Z.~Yin, E.~P. Xing, and Z.~Shen, ``Squeeze, recover and relabel: Dataset condensation at imagenet scale from {A} new perspective,'' in \emph{Neural Information Processing Systems}.\hskip 1em plus 0.5em minus 0.4em\relax NeurIPS, 2023.

\bibitem[Yin and Shen(2024)]{yin2024dataset}
\BIBentryALTinterwordspacing
Z.~Yin and Z.~Shen, ``Dataset distillation in large data era,'' 2024. [Online]. Available: \url{https://openreview.net/forum?id=kpEz4Bxs6e}
\BIBentrySTDinterwordspacing

\bibitem[Shao et~al.(2023)Shao, Yin, Zhang, and Shen]{shao2023generalized}
S.~Shao, Z.~Yin, X.~Zhang, and Z.~Shen, ``Generalized large-scale data condensation via various backbone and statistical matching,'' \emph{arXiv preprint arXiv:2311.17950}, 2023.

\bibitem[Krizhevsky et~al.(2009)Krizhevsky, Hinton, et~al.]{CIFAR}
A.~Krizhevsky, G.~Hinton \emph{et~al.}, ``Learning multiple layers of features from tiny images,'' 2009.

\bibitem[Sun et~al.(2024)Sun, Shi, Yu, and Lin]{RDED}
P.~Sun, B.~Shi, D.~Yu, and T.~Lin, ``On the diversity and realism of distilled dataset: An efficient dataset distillation paradigm,'' in \emph{Computer Vision and Pattern Recognition}.\hskip 1em plus 0.5em minus 0.4em\relax IEEE, 2024.

\bibitem[Wang et~al.(2018)Wang, Zhu, Torralba, and Efros]{dd_begin}
T.~Wang, J.-Y. Zhu, A.~Torralba, and A.~A. Efros, ``Dataset distillation,'' \emph{arXiv preprint arXiv:1811.10959}, 2018.

\bibitem[Cui et~al.(2023)Cui, Wang, Si, and Hsieh]{dd_tesla}
J.~Cui, R.~Wang, S.~Si, and C.~Hsieh, ``Scaling up dataset distillation to imagenet-1k with constant memory,'' in \emph{International Conference on Machine Learning}, vol. 202.\hskip 1em plus 0.5em minus 0.4em\relax Honolulu, Hawaii, {USA}: {PMLR}, 2023, pp. 6565--6590.

\bibitem[Wang et~al.(2022)Wang, Zhao, Peng, Zhu, Yang, Wang, Huang, Bilen, Wang, and You]{dd_CAFE}
K.~Wang, B.~Zhao, X.~Peng, Z.~Zhu, S.~Yang, S.~Wang, G.~Huang, H.~Bilen, X.~Wang, and Y.~You, ``Cafe: Learning to condense dataset by aligning features,'' in \emph{Computer Vision and Pattern Recognition}.\hskip 1em plus 0.5em minus 0.4em\relax New Orleans, LA, USA: {IEEE}, Jun. 2022, pp. 12\,196--12\,205.

\bibitem[Nguyen et~al.(2020)Nguyen, Chen, and Lee]{dd_kip}
T.~Nguyen, Z.~Chen, and J.~Lee, ``Dataset meta-learning from kernel ridge-regression,'' \emph{arXiv preprint arXiv:2011.00050}, 2020.

\bibitem[Kim et~al.(2022)Kim, Kim, Oh, Yun, Song, Jeong, Ha, and Song]{dd_efficient_parameterization}
J.~Kim, J.~Kim, S.~J. Oh, S.~Yun, H.~Song, J.~Jeong, J.~Ha, and H.~O. Song, ``Dataset condensation via efficient synthetic-data parameterization,'' in \emph{International Conference on Machine Learning}, vol. 162.\hskip 1em plus 0.5em minus 0.4em\relax Baltimore, Maryland, {USA}: {PMLR}, Jul. 2022, pp. 11\,102--11\,118.

\bibitem[Zhou et~al.(2023)Zhou, Wang, Gu, Peng, Lian, Zhang, You, and Feng]{dataset_quantization}
D.~Zhou, K.~Wang, J.~Gu, X.~Peng, D.~Lian, Y.~Zhang, Y.~You, and J.~Feng, ``Dataset quantization,'' in \emph{Proceedings of the IEEE/CVF International Conference on Computer Vision}, 2023, pp. 17\,205--17\,216.

\bibitem[Zhang et~al.(2024{\natexlab{a}})Zhang, Li, Lin, Wang, Qian, and Ge]{add_related_work1}
H.~Zhang, S.~Li, F.~Lin, W.~Wang, Z.~Qian, and S.~Ge, ``Dance: Dual-view distribution alignment for dataset condensation,'' \emph{arXiv preprint arXiv:2406.01063}, 2024.

\bibitem[Zhang et~al.(2024{\natexlab{b}})Zhang, Li, Wang, Zeng, and Ge]{add_related_work2}
H.~Zhang, S.~Li, P.~Wang, D.~Zeng, and S.~Ge, ``M3d: Dataset condensation by minimizing maximum mean discrepancy,'' in \emph{Proceedings of the AAAI Conference on Artificial Intelligence}, vol.~38, no.~8, 2024, pp. 9314--9322.

\bibitem[Deng et~al.(2024)Deng, Li, Ding, Wang, Zhang, Huang, Huo, and Gao]{add_related_work3}
W.~Deng, W.~Li, T.~Ding, L.~Wang, H.~Zhang, K.~Huang, J.~Huo, and Y.~Gao, ``Exploiting inter-sample and inter-feature relations in dataset distillation,'' in \emph{Proceedings of the IEEE/CVF Conference on Computer Vision and Pattern Recognition}, 2024, pp. 17\,057--17\,066.

\bibitem[Hinton et~al.(2015)Hinton, Vinyals, and Dean]{vanillakd}
\BIBentryALTinterwordspacing
G.~Hinton, O.~Vinyals, and J.~Dean, ``Distilling the knowledge in a neural network,'' 2015. [Online]. Available: \url{https://arxiv.org/abs/1503.02531}
\BIBentrySTDinterwordspacing

\bibitem[Gou et~al.(2021)Gou, Yu, Maybank, and Tao]{kdsurvey}
J.~Gou, B.~Yu, S.~J. Maybank, and D.~Tao, ``Knowledge distillation: A survey,'' \emph{International Journal of Computer Vision}, vol. 129, no.~6, pp. 1789--1819, 2021.

\bibitem[Chen et~al.(2024)Chen, Zhang, Dong, and Zhu]{cwa}
H.~Chen, Y.~Zhang, Y.~Dong, and J.~Zhu, ``Rethinking model ensemble in transfer-based adversarial attacks,'' in \emph{International Conference on Learning Representations}.\hskip 1em plus 0.5em minus 0.4em\relax Vienna, Austria: OpenReview.net, May 2024.

\bibitem[Foret et~al.(2020)Foret, Kleiner, Mobahi, and Neyshabur]{iclr2020_sam}
P.~Foret, A.~Kleiner, H.~Mobahi, and B.~Neyshabur, ``Sharpness-aware minimization for efficiently improving generalization,'' in \emph{International Conference on Learning Representations}, 2020.

\bibitem[Chen et~al.(2022)Chen, Shao, Wang, Shang, Chen, Ji, and Wu]{eccvw_sam}
H.~Chen, S.~Shao, Z.~Wang, Z.~Shang, J.~Chen, X.~Ji, and X.~Wu, ``Bootstrap generalization ability from loss landscape perspective,'' in \emph{European Conference on Computer Vision}.\hskip 1em plus 0.5em minus 0.4em\relax Springer, 2022, pp. 500--517.

\bibitem[Liu et~al.(2023{\natexlab{b}})Liu, Xing, Li, Dalal, He, and Wang]{liu2023dataset}
H.~Liu, T.~Xing, L.~Li, V.~Dalal, J.~He, and H.~Wang, ``Dataset distillation via the wasserstein metric,'' \emph{arXiv preprint arXiv:2311.18531}, 2023.

\bibitem[Du et~al.(2022)Du, Zhou, Feng, Tan, and Zhou]{nips2022_sam}
J.~Du, D.~Zhou, J.~Feng, V.~Tan, and J.~T. Zhou, ``Sharpness-aware training for free,'' in \emph{Advances in Neural Information Processing Systems}, vol.~35.\hskip 1em plus 0.5em minus 0.4em\relax New Orleans, Louisiana, USA: NeurIPS, Dec. 2022, pp. 23\,439--23\,451.

\bibitem[Bahri et~al.(2021)Bahri, Mobahi, and Tay]{sam_llm}
D.~Bahri, H.~Mobahi, and Y.~Tay, ``Sharpness-aware minimization improves language model generalization,'' \emph{arXiv preprint arXiv:2110.08529}, 2021.

\bibitem[Tavanaei(2020)]{tiny_imagenet}
\BIBentryALTinterwordspacing
A.~Tavanaei, ``Embedded encoder-decoder in convolutional networks towards explainable {AI},'' vol. abs/2007.06712, 2020. [Online]. Available: \url{https://arxiv.org/abs/2007.06712}
\BIBentrySTDinterwordspacing

\bibitem[Sandler et~al.(2018)Sandler, Howard, Zhu, Zhmoginov, and Chen]{mobilenetv2}
M.~Sandler, A.~G. Howard, M.~Zhu, A.~Zhmoginov, and L.~Chen, ``Mobilenetv2: Inverted residuals and linear bottlenecks,'' in \emph{Computer Vision and Pattern Recognition}.\hskip 1em plus 0.5em minus 0.4em\relax Salt Lake City, UT, USA: IEEE, Jun. 2018, pp. 4510--4520.

\bibitem[Touvron et~al.(2021)Touvron, Cord, Douze, Massa, Sablayrolles, and J{\'{e}}gou]{deit}
H.~Touvron, M.~Cord, M.~Douze, F.~Massa, A.~Sablayrolles, and H.~J{\'{e}}gou, ``Training data-efficient image transformers {\&} distillation through attention,'' in \emph{International Conference on Machine Learning}, M.~Meila and T.~Zhang, Eds., vol. 139.\hskip 1em plus 0.5em minus 0.4em\relax Virtual Event: {PMLR}, Jul. 2021, pp. 10\,347--10\,357.

\bibitem[Liu et~al.(2021)Liu, Lin, Cao, Hu, Wei, Zhang, Lin, and Guo]{SWIN-T}
Z.~Liu, Y.~Lin, Y.~Cao, H.~Hu, Y.~Wei, Z.~Zhang, S.~Lin, and B.~Guo, ``Swin transformer: Hierarchical vision transformer using shifted windows,'' in \emph{International Conference on Computer Vision}, 2021, pp. 10\,012--10\,022.

\bibitem[Tan and Le(2019)]{efficientnet}
M.~Tan and Q.~Le, ``Efficientnet: Rethinking model scaling for convolutional neural networks,'' in \emph{International conference on machine learning}.\hskip 1em plus 0.5em minus 0.4em\relax PMLR, 2019, pp. 6105--6114.

\bibitem[Liu et~al.(2022)Liu, Mao, Wu, Feichtenhofer, Darrell, and Xie]{convnext}
Z.~Liu, H.~Mao, C.-Y. Wu, C.~Feichtenhofer, T.~Darrell, and S.~Xie, ``A convnet for the 2020s,'' in \emph{Proceedings of the IEEE/CVF Conference on Computer Vision and Pattern Recognition}, 2022, pp. 11\,976--11\,986.

\bibitem[Zhang et~al.(2018)Zhang, Zhou, Lin, and Sun]{shufflenet}
X.~Zhang, X.~Zhou, M.~Lin, and J.~Sun, ``Shufflenet: An extremely efficient convolutional neural network for mobile devices,'' in \emph{Computer Vision and Pattern Recognition}, 2018, pp. 6848--6856.

\bibitem[Srinivas and Babu(2015)]{srinivas2015data}
S.~Srinivas and R.~V. Babu, ``Data-free parameter pruning for deep neural networks,'' \emph{arXiv preprint arXiv:1507.06149}, 2015.

\bibitem[Prabhu et~al.(2020)Prabhu, Torr, and Dokania]{Prabhu_Torr_Dokania_2020}
A.~Prabhu, P.~H.~S. Torr, and P.~K. Dokania, ``\BIBforeignlanguage{en-US}{Gdumb: A simple approach that questions our progress in continual learning},'' in \emph{\BIBforeignlanguage{en-US}{European Conference on Computer Vision}}.\hskip 1em plus 0.5em minus 0.4em\relax Springer, Jan 2020, p. 524–540.

\bibitem[Paszke et~al.(2019)Paszke, Gross, Massa, Lerer, Bradbury, Chanan, Killeen, Lin, Gimelshein, Antiga, et~al.]{pytorch}
A.~Paszke, S.~Gross, F.~Massa, A.~Lerer, J.~Bradbury, G.~Chanan, T.~Killeen, Z.~Lin, N.~Gimelshein, L.~Antiga \emph{et~al.}, ``Pytorch: An imperative style, high-performance deep learning library,'' in \emph{Neural Information Processing Systems}, Vancouver, BC, Canada, Dec. 2019.

\bibitem[Ostle et~al.(1963)]{ostle1963statistics}
B.~Ostle \emph{et~al.}, ``Statistics in research.'' \emph{Statistics in research.}, no. 2nd Ed, 1963.

\bibitem[Zhou et~al.(2024)Zhou, Yin, Shao, and Shen]{zhou2024self}
M.~Zhou, Z.~Yin, S.~Shao, and Z.~Shen, ``Self-supervised dataset distillation: A good compression is all you need,'' \emph{arXiv preprint arXiv:2404.07976}, 2024.

\bibitem[Wu et~al.(2024)Wu, Du, Liu, Lin, Cheng, and Xu]{wu2024dd}
Y.~Wu, J.~Du, P.~Liu, Y.~Lin, W.~Cheng, and W.~Xu, ``Dd-robustbench: An adversarial robustness benchmark for dataset distillation,'' \emph{arXiv preprint arXiv:2403.13322}, 2024.

\bibitem[Kim(2020)]{kim2020torchattacks}
H.~Kim, ``Torchattacks: A pytorch repository for adversarial attacks,'' \emph{arXiv preprint arXiv:2010.01950}, 2020.

\end{thebibliography}
}

\clearpage
\appendix

\section*{\Large{Appendix}}

\section{Implementation Details}
\label{tab:imple_details}
Here, we complement both the hyperparameter settings and the backbone choices utilized for the comparison and ablation experiments in the main paper.

\subsection{Hyperparameter Settings}
\label{tab:hyperparameter_settings}
\begin{table}[ht]
\centering
\begin{minipage}[b]{0.48\textwidth}
\centering
\subfigure[Data Synthesis]{
\resizebox{1.0\textwidth}{!}{%
\begin{tabular}{lcc}
\toprule
\textbf{Config} & \textbf{Value} & \textbf{Explanation} \\
\midrule
Iteration & 2000 & NA \\
Optimizer & Adam & $\beta_1,\beta_2=(0.5,0.9)$\\
Learning Rate & 0.05 & NA \\
Batch Size & 80 & NA \\
Initialization & RDED & \makecell{Initialized using images\\ synthesized from RDED} \\
$\alpha$, $\beta$, $\tau$ & 0.5, 0.99, 4 & \makecell{Control\\category-aware matching\\and flatness regularization}\\
\bottomrule
\end{tabular}}
}
\end{minipage}
\hfill 
\begin{minipage}[b]{0.48\textwidth}
\centering
\subfigure[Soft Label Generation and Post-Evaluation]{
\resizebox{1.0\textwidth}{!}{%
\begin{tabular}{lcc}
\toprule
\textbf{Config} & \textbf{Value} & \textbf{Explanation} \\
\midrule
Epochs & 300 & NA \\
Optimizer & AdamW & NA \\
Learning Rate & 0.001 & \makecell{Only use 1e-4\\for Swin-Tiny}  \\
Batch Size & 100 & NA \\
EMA Rate & 0.99 & \makecell{Control EMA-based\\ Evaluation} \\
Scheduler & Smoothing LR Schedule & $\zeta=2$ \\\vspace{3pt}
Augmentation & \makecell{RandomResizedCrop\\RandomHorizontalFlip} & NA \\
\bottomrule
\end{tabular}}
}
\end{minipage}
\caption{Hyperparameter setting on ImageNet-1k.}
\label{tab:setting_imagenet_1k}
\end{table}
\begin{table}[ht]
\centering
\begin{minipage}[b]{0.48\textwidth}
\centering
\subfigure[Data Synthesis]{
\resizebox{1.0\textwidth}{!}{%
\begin{tabular}{lcc}
\toprule
\textbf{Config} & \textbf{Value} & \textbf{Explanation} \\
\midrule
Iteration & 2000 & NA \\
Optimizer & Adam & $\beta_1,\beta_2=(0.5,0.9)$\\
Learning Rate & 0.05 & NA \\
Batch Size & 100 & NA \\
Initialization & RDED & \makecell{Initialized using images\\ synthesized from RDED} \\
$\alpha$, $\beta$, $\tau$ & 0.5, 0.99, 4 & \makecell{Control\\category-aware matching\\and flatness regularization}\\
\bottomrule
\end{tabular}}
}
\end{minipage}
\hfill 
\begin{minipage}[b]{0.48\textwidth}
\centering
\subfigure[Soft Label Generation and Post-Evaluation]{
\resizebox{1.0\textwidth}{!}{%
\begin{tabular}{lcc}
\toprule
\textbf{Config} & \textbf{Value} & \textbf{Explanation} \\
\midrule
Epochs & 1000 & NA \\
Optimizer & AdamW & NA \\
Learning Rate & 0.001 & NA \\
Batch Size & 50 & NA \\
EMA Rate & 0.99 & \makecell{Control EMA-based\\ Evaluation} \\
Scheduler & Smoothing LR Schedule & $\zeta=2$ \\\vspace{3pt}
Augmentation & \makecell{RandAugment\\RandomResizedCrop\\RandomHorizontalFlip} & NA \\
\bottomrule
\end{tabular}}
}
\end{minipage}
\caption{Hyperparameter setting on ImageNet-10.}
\label{tab:setting_imagenet_10}
\end{table}
\begin{table}[ht]
\centering
\begin{minipage}[b]{0.48\textwidth}
\centering
\subfigure[Data Synthesis]{
\resizebox{1.0\textwidth}{!}{%
\begin{tabular}{lcc}
\toprule
\textbf{Config} & \textbf{Value} & \textbf{Explanation} \\
\midrule
Iteration & 2000 & NA \\
Optimizer & Adam & $\beta_1,\beta_2=(0.5,0.9)$\\
Learning Rate & 0.05 & NA \\
Batch Size & 100 & NA \\
Initialization & Original Image & \makecell{Initialized using images\\ from training dataset} \\
$\alpha$, $\beta$, $\tau$ & 0.5, 0.99, 4 & \makecell{Control\\category-aware matching\\and flatness regularization}\\
\bottomrule
\end{tabular}}
}
\end{minipage}
\hfill 
\begin{minipage}[b]{0.48\textwidth}
\centering
\subfigure[Soft Label Generation and Post-Evaluation]{
\resizebox{1.0\textwidth}{!}{%
\begin{tabular}{lcc}
\toprule
\textbf{Config} & \textbf{Value} & \textbf{Explanation} \\
\midrule
Epochs & 300 & \makecell{Only use 1000 for IPC 1} \\
Optimizer & AdamW & NA \\
Learning Rate & 0.001 & NA \\
Batch Size & 100 & NA \\
EMA Rate & 0.99 & \makecell{Control EMA-based\\ Evaluation} \\
Scheduler & Smoothing LR Schedule & $\zeta=2$ \\
Augmentation & \makecell{RandAugment\\RandomResizedCrop\\RandomHorizontalFlip} & NA \\
\bottomrule
\end{tabular}}
}
\end{minipage}
\caption{Hyperparameter setting on Tiny-ImageNet.}
\label{tab:setting_tiny_imagenet}
\end{table}
\begin{table}[ht]
\centering
\begin{minipage}[b]{0.48\textwidth}
\centering
\subfigure[Data Synthesis]{
\resizebox{1.0\textwidth}{!}{%
\begin{tabular}{lcc}
\toprule
\textbf{Config} & \textbf{Value} & \textbf{Explanation} \\
\midrule
Iteration & 2000 & NA \\
Optimizer & Adam & $\beta_1,\beta_2=(0.5,0.9)$\\
Learning Rate & 0.05 & NA \\
Batch Size & 100 & NA \\
Initialization & Original Image & \makecell{Initialized using images\\ from training dataset} \\\vspace{3pt}
$\alpha$, $\beta$, $\tau$ & 0.5, 0.99, 4 & \makecell{Control\\category-aware matching\\and flatness regularization}\\
\bottomrule
\end{tabular}}
}
\end{minipage}
\hfill 
\begin{minipage}[b]{0.48\textwidth}
\centering
\subfigure[Soft Label Generation and Post-Evaluation]{
\resizebox{1.0\textwidth}{!}{%
\begin{tabular}{lcc}
\toprule
\textbf{Config} & \textbf{Value} & \textbf{Explanation} \\
\midrule
Epochs & 1000 & NA \\
Optimizer & AdamW & NA \\
Learning Rate & 0.001 & NA \\
Batch Size & 50 & NA \\
EMA Rate & 0.99 & \makecell{Control EMA-based\\ Evaluation} \\
Scheduler & Smoothing LR Schedule & $\zeta=2$ \\
Augmentation & \makecell{RandAugment\\RandomResizedCrop\\RandomHorizontalFlip} & NA \\
\bottomrule
\end{tabular}}
}
\end{minipage}
\caption{Hyperparameter setting on CIFAR-100.}
\label{tab:setting_cifar_100}
\end{table}
\begin{table}[h!]
\centering
\begin{minipage}[b]{0.48\textwidth}
\centering
\subfigure[Data Synthesis]{
\resizebox{1.0\textwidth}{!}{%
\begin{tabular}{lcc}
\toprule
\textbf{Config} & \textbf{Value} & \textbf{Explanation} \\
\midrule
Iteration & 75 & NA \\
Optimizer & Adam & $\beta_1,\beta_2=(0.5,0.9)$\\
Learning Rate & 0.05 & NA \\
Batch Size & All & \makecell{The number of \\synthesized images} \\
Initialization & Original Image & \makecell{Initialized using images\\ from training dataset} \\
$\alpha$, $\beta$, $\tau$ & 0.5, 0.99, 4 & \makecell{Control\\category-aware matching\\and flatness regularization}\\
\bottomrule
\end{tabular}}
}
\end{minipage}
\hfill 
\begin{minipage}[b]{0.48\textwidth}
\centering
\subfigure[Soft Label Generation and Post-Evaluation]{
\resizebox{1.0\textwidth}{!}{%
\begin{tabular}{lcc}
\toprule
\textbf{Config} & \textbf{Value} & \textbf{Explanation} \\
\midrule
Epochs & 1000 & NA \\
Optimizer & AdamW & NA \\
Learning Rate & 0.001 & NA \\
Batch Size & 25 & NA \\
EMA Rate & 0.99 & \makecell{Control EMA-based\\ Evaluation} \\\vspace{3pt}
Scheduler & MultiStepLR & \makecell{$\gamma=0.5$\\milestones=[800,900,950]} \\
Augmentation & \makecell{RandAugment\\RandomResizedCrop\\RandomHorizontalFlip} & NA \\
\bottomrule
\end{tabular}}
}
\end{minipage}
\caption{Hyperparameter setting on CIFAR-10.}
\label{tab:setting_cifar_10}
\end{table}

We detail the hyperparameter settings of EDC for various datasets, including ImageNet-1k, ImageNet-10, Tiny-ImageNet, CIFAR-100, and CIFAR-10, in Tables~\ref{tab:setting_imagenet_1k}, \ref{tab:setting_imagenet_10}, \ref{tab:setting_tiny_imagenet}, \ref{tab:setting_cifar_100}, and \ref{tab:setting_cifar_10}, respectively. For epochs, a critical factor affecting computational cost, we utilize strategies from SRe$^2$L, G-VBSM, and RDED for ImageNet-1k and follow RDED for the other datasets. In the data synthesis phase, we reduce the iteration count of hyperparameters by half compared to those used in SRe$^2$L and G-VBSM.

\subsection{Network Architectures on Different Datasets}
\label{apd:employed_backbone}
This section outlines the specific configurations of the backbones employed in the data synthesis and soft label generation phases, details of which are omitted from the main paper.

\paragraph{ImageNet-1k.} We utilize pre-trained models \{ResNet-18, MobileNet-V2, ShuffleNet-V2, EfficientNet-V2, AlexNet\} from torchvision~\citep{pytorch} as observer models in data synthesis. To reduce computational load, we exclude EfficientNet-V2 from the soft label generation process, a decision in line with our strategy of selecting more efficient backbones, a concept referred to as better backbone choice in the main paper. An extensive ablation analysis is available in Appendix~\ref{apd:add_ab_experiment}.

\paragraph{ImageNet-10.} Prior to data synthesis, we train \{ResNet-18, MobileNet-V2, ShuffleNet-V2, EfficientNet-V2\} from scratch for 20 epochs and save their respective checkpoints. Subsequently, these pre-trained models are consistently employed for both data synthesis and soft label generation.

\paragraph{Tiny-ImageNet.} We adopt the same backbone configurations as G-VBSM, specifically utilizing \{ResNet-18, MobileNet-V2, ShuffleNet-V2, EfficientNet-V2\} for both data synthesis and soft label generation. Each of these models has been trained on the original dataset with 50 epochs.

\paragraph{CIFAR-10\&CIFAR-100.} For small-scale datasets, we enhance the \textit{baseline} G-VBSM model by incorporating three additional lightweight backbones. Consequently, the backbones utilized for data synthesis and soft label generation comprise \{ResNet-18, ConvNet-W128, MobileNet-V2, WRN-16-2, ShuffleNet-V2, ConvNet-D1, ConvNet-D2, ConvNet-W32\}. To demonstrate the effectiveness of our approach, we conduct comparative experiments and present results in Table~\ref{tab:comparison_of_backbone_c10}, which illustrates that G-VBSM achieves improved performance with this enhanced backbone configuration.

\begin{table}[!h]
\centering
\footnotesize
\renewcommand\arraystretch{1.04}
\vspace{-4pt}
\resizebox{0.8\textwidth}{!}{%
\begin{tabular}{c|l|ccc}
\hline
\multirow{4}{*}{\makecell{CIFAR-10\\(IPC 10)}} & Verified Model & ResNet-18 & AlexNet & VGG11-BN \\\cline{2-5}
&100 backbones (MTT) & 46.4 & 34.2 & 50.3 \\
&5 backbones (original setting of G-VBSM) & 53.5 & 31.7 & 55.2 \\
&8 backbones (new setting of G-VBSM) & \cellcolor{gray!25} 58.9 & \cellcolor{gray!25} 36.2 & \cellcolor{gray!25} 58.0 \\
\hline
\end{tabular}}
\vspace{2pt}
\caption{\textbf{Ablation studies on CIFAR-10 with IPC 10.} With the remaining settings are the same as those of G-VBSM, our new backbone setting achieves better performance.}
\label{tab:comparison_of_backbone_c10}
\vspace{-4pt}
\end{table}

\section{Theoretical Derivations} \label{theory}
Here, we give a detailed statement of the definitions, assumptions, theorems, and corollaries relevant to this paper.

\subsection{Random Initialization vs. Real Image Initialization}
\label{apd:random_vs_real}
In the data synthesis phase, random initialization involves using Gaussian noise, while real image initialization uses condensed images derived from training-free algorithms, such as RDED. Specifically, we denote the datasets initialized via random and real image methods as $\mathcal{X}^{\mathcal{S}}_\textbf{\textrm{random}}$ and $\mathcal{X}^{\mathcal{S}}_\textbf{\textrm{real}}$, respectively. For coupling ($\mathcal{X}^{\mathcal{S}}_\textbf{\textrm{random}}$, $\mathcal{X}^{\mathcal{S}}_\textbf{\textrm{real}}$), where $\mathcal{X}^{\mathcal{S}}_\textbf{\textrm{random}}\sim \pi_\textbf{\textrm{random}},\ \mathcal{X}^{\mathcal{S}}_\textbf{\textrm{real}}\sim \pi_\textbf{\textrm{real}}$ and satisfies $p(\pi_\textbf{\textrm{random}},\pi_\textbf{\textrm{real}}) = p(\pi_\textbf{\textrm{random}})p(\pi_\textbf{\textrm{real}})$, we have the mutual information (MI) between $\pi_\textbf{\textrm{random}}$ and $\pi_\textbf{\textrm{real}}$ is $0$, \textit{a.k.a.}, $I(\pi_\textbf{\textrm{random}},\pi_\textbf{\textrm{real}})=0$. By contrast, training-free algorithms~\citep{RDED,dataset_quantization} synthesize the compressed data $\mathcal{X}^\mathcal{S}_\textbf{\textrm{free}}:=\phi(\mathcal{X}^\mathcal{S}_\textbf{\textrm{real}})$ via $\mathcal{X}^\mathcal{S}_\textbf{\textrm{real}}$, satisfying $p(\mathcal{X}^\mathcal{S}_\textbf{\textrm{free}}|\mathcal{X}^\mathcal{S}_\textbf{\textrm{real}})>0$. When the cost function $\mathbb{E}[c(a-b)]\propto 1/I(\textrm{Law}(a),\textrm{Law}(b))$, we have $\mathbb{E}[c(\mathcal{X}^\mathcal{S}_\textbf{\textrm{real}}-\mathcal{X}^\mathcal{S}_\textbf{\textrm{free}})] \leq \mathbb{E}[c(\mathcal{X}^\mathcal{S}_\textbf{\textrm{real}}-\mathcal{X}^\mathcal{S}_\textbf{\textrm{random}})]$.

\begin{proof}
\begin{equation}
\small
\begin{aligned}
\mathbb{E}[c(\mathcal{X}^\mathcal{S}_\textbf{\textrm{real}}-\mathcal{X}^\mathcal{S}_\textbf{\textrm{free}})] &  = k/ I(\textrm{Law}(\mathcal{X}^\mathcal{S}_\textbf{\textrm{real}}),\textrm{Law}(\mathcal{X}^\mathcal{S}_\textbf{\textrm{free}}))  \\
& = k/\textrm{D}_\textrm{KL}(p(\pi_\textbf{\textrm{real}},\pi_\textbf{\textrm{free}})||p(\pi_\textbf{\textrm{real}})p(\pi_\textbf{\textrm{free}})) \\
& = k/[H(\pi_\textbf{\textrm{real}})-H(\pi_\textbf{\textrm{real}}|\pi_\textbf{\textrm{free}})]\\
& \leq k/[H(\pi_\textbf{\textrm{real}})]\\
& = k/[H(\pi_\textbf{\textrm{real}})-H(\pi_\textbf{\textrm{real}}|\pi_\textbf{\textrm{random}})]\\
& =  k/I(\textrm{Law}(\mathcal{X}^\mathcal{S}_\textbf{\textrm{real}}),\textrm{Law}(\mathcal{X}^\mathcal{S}_\textbf{\textrm{random}})) \\
& =\mathbb{E}[c(\mathcal{X}^\mathcal{S}_\textbf{\textrm{real}}-\mathcal{X}^\mathcal{S}_\textbf{\textrm{random}})], \\
\end{aligned}
\label{eq:mi_cost_function}
\end{equation}
where $k\in \mathbb{R}^+$ denotes a constant. And $D_\textrm{KL}(\cdot||\cdot)$ and $H(\cdot)$ stand for Kullback-Leibler divergence and entropy, respectively.
\end{proof}
From the theoretical perspective described, it becomes evident that initializing with real images enhances MI more significantly than random initialization between the distilled and the original datasets at the start of the data synthesis phase. This improvement substantially alleviates the challenges inherent in data synthesis. Furthermore, our exploratory experiments demonstrate that the generalized matching loss~\citep{shao2023generalized} for real image initialization remains consistently lower compared to that of random initialization throughout the data synthesis phase.

\subsection{Theoretical Derivations of Soft Category-Aware Matching}
\label{apd:category_matching}
\begin{definition}
\label{def:statistical_matching}
(Statistical Matching) Assume that we have $N$ $D$-dimensional random samples $\{x_i\in \mathcal{R}^D\}_{i=1}^{N}$ with an unknown distribution $p_\textrm{mix}(x)$, we define two forms of statistical matching for dataset distillation:

\textbf{Form (1):} Estimate the mean $\mathbb{E}[x]$ and variance $\mathbb{D}[x]$ of samples $\{x_i\in \mathcal{R}^D\}_{i=1}^{N}$. Then, synthesize $M$ ($M\ll N$) distilled samples $\{y_i\in \mathcal{R}^D\}_{i=1}^{M}$ such that the absolute differences between the variances ($|\mathbb{D}[x]-\mathbb{D}[y]|$) and means ($|\mathbb{E}[x]-\mathbb{E}[y]|$) of the original and distilled samples are $\leq\epsilon$.
\\

\textbf{Form (2):} Consider $p_{\textrm{mix}}(x)$ to be a linear combination of multiple subdistributions, expressed as $p_{\textrm{mix}}(x) = \int_{\mathbf{C}} p(x|c_i)p(c_i)dc_i$, where $c_i$ denotes a component of the original distribution. Given Assumption~\ref{ass:gmm}, we can treat $p_{\textrm{mix}}(x)$ as a GMM, with each component $p(x|c_i)$ following a Gaussian distribution. For each component, estimate the mean $\mathbb{E}[x^j]$ and variance $\mathbb{D}[x^j]$ using $N_j$ samples $\{x^j_i\}_{i=1}^{N_j}$, ensuring that $\sum_{j=1}^\mathbf{C}N_j=N$. Subsequently, synthesize $M$ ($M\ll N$) distilled samples across all components $\bigcup_{j=1}^{\mathbf{C}}\{y^j_i\}_{i=1}^{M_j}$, where $\sum_{j=1}^\mathbf{C}M_j=M$. This process aims to ensure that for each component, the absolute differences between the variances ($|\mathbb{D}[x^j]-\mathbb{D}[y^j]|$) and means ($|\mathbb{E}[x^j]-\mathbb{E}[y^j]|$) of the original and distilled samples $\leq\epsilon$.
\end{definition}

Based on Definition~\ref{def:statistical_matching}, here we provide several relevant theoretical conclusion.
\begin{lemma}
\label{lemma:variance_error}
Consider a sample set $\mathbb{S}$, where each sample $\mathcal{X}$ within $\mathbb{S}$ belongs to $\mathcal{R}^{D}$. Assume any two variables $x_i$ and $x_j$ in $\mathbb{S}$ satisfies $p(x_i,x_j)=p(x_i)p(x_j)$. This set $\mathbb{S}$ comprises \textbf{C} disjoint subsets $\{\mathbb{S}_\textrm{1}, \mathbb{S}_\textrm{2}, \ldots, \mathbb{S}_\textrm{C}\}$, ensuring that for any $1 \leq i < j \leq C$, the intersection $\mathbb{S}_\textrm{i} \cap \mathbb{S}_\textrm{j} = \emptyset$ and the union $\bigcup_{k=1}^{C} \mathbb{S}_\textrm{k} = \mathbb{S}$. Consequently, the expected value over the variance within the subsets, denoted as $\mathbb{E}_{\mathbb{S}_\textrm{sub} \sim \{\mathbb{S}_\textrm{1}, \ldots, \mathbb{S}_\textrm{C}\}}\mathbb{D}_{\mathcal{X} \sim \mathbb{S}_\textrm{sub}}[\mathcal{X}]$, is smaller than or equal to the variance within the entire set, $\mathbb{D}_{\mathcal{X} \sim \mathbb{S}}[\mathcal{X}]$.
\end{lemma}
\begin{proof}
\begin{equation}
\footnotesize
\begin{aligned}
& \mathbb{E}_{\mathbb{S}_\textrm{sub} \sim \{\mathbb{S}_\textrm{1}, \ldots, \mathbb{S}_\textrm{C}\}}\mathbb{D}_{\mathcal{X} \sim \mathbb{S}_\textrm{sub}}[\mathcal{X}] \\
&=  \mathbb{E}_{\mathbb{S}_\textrm{sub} \sim \{\mathbb{S}_\textrm{1}, \ldots, \mathbb{S}_\textrm{C}\}}(\mathbb{E}_{\mathcal{X} \sim \mathbb{S}_\textrm{sub}}[\mathcal{X}\circ\mathcal{X}] - \mathbb{E}_{\mathcal{X} \sim \mathbb{S}_\textrm{sub}}[\mathcal{X}]\circ\mathbb{E}_{\mathcal{X} \sim \mathbb{S}_\textrm{sub}}[\mathcal{X}])   \\
&= \mathbb{E}_{\mathcal{X} \sim \mathbb{S}}[\mathcal{X}\circ\mathcal{X}] - \mathbb{E}_{\mathcal{X} \sim \mathbb{S}}[\mathcal{X}]\circ\mathbb{E}_{\mathcal{X} \sim \mathbb{S}}[\mathcal{X}] + \mathbb{E}_{\mathcal{X} \sim \mathbb{S}}[\mathcal{X}]\circ\mathbb{E}_{\mathcal{X} \sim \mathbb{S}}[\mathcal{X}] \\
&\quad- \mathbb{E}_{\mathbb{S}_\textrm{sub} \sim \{\mathbb{S}_\textrm{1}, \ldots, \mathbb{S}_\textrm{C}\}}\mathbb{E}_{\mathcal{X} \sim \mathbb{S}_\textrm{sub}}[\mathcal{X}]\circ\mathbb{E}_{\mathcal{X} \sim \mathbb{S}_\textrm{sub}}[\mathcal{X}]\\
& = \mathbb{D}_{\mathcal{X} \sim \mathbb{S}}[\mathcal{X}] - \mathbb{E}_{\mathbb{S}_\textrm{sub} \sim \{\mathbb{S}_\textrm{1}, \ldots, \mathbb{S}_\textrm{C}\}}\mathbb{E}_{\mathcal{X} \sim \mathbb{S}_\textrm{sub}}[\mathcal{X}]\circ\mathbb{E}_{\mathcal{X} \sim \mathbb{S}_\textrm{sub}}[\mathcal{X}]\\
&\quad + \mathbb{E}_{\mathcal{X} \sim \mathbb{S}}[\mathcal{X}]\circ \mathbb{E}_{\mathcal{X} \sim \mathbb{S}}[\mathcal{X}]  \\
& = \mathbb{D}_{\mathcal{X} \sim \mathbb{S}}[\mathcal{X}] - \mathbb{E}_{\mathbb{S}_\textrm{sub} \sim \{\mathbb{S}_\textrm{1}, \ldots, \mathbb{S}_\textrm{C}\}}\mathbb{E}_{\mathcal{X} \sim \mathbb{S}_\textrm{sub}}[\mathcal{X}]\circ\mathbb{E}_{\mathcal{X} \sim \mathbb{S}_\textrm{sub}}[\mathcal{X}] \\
&\quad + \mathbb{E}_{\mathbb{S}_\textrm{sub} \sim \{\mathbb{S}_\textrm{1}, \ldots, \mathbb{S}_\textrm{C}\}}\mathbb{E}_{\mathcal{X} \sim \mathbb{S}_\textrm{sub}}[\mathcal{X}]\circ\mathbb{E}_{\mathbb{S}_\textrm{sub} \sim \{\mathbb{S}_\textrm{1}, \ldots, \mathbb{S}_\textrm{C}\}}\mathbb{E}_{\mathcal{X} \sim \mathbb{S}_\textrm{sub}}[\mathcal{X}] \\
& = \mathbb{D}_{\mathcal{X} \sim \mathbb{S}}[\mathcal{X}] -  \mathbb{D}_{\mathbb{S}_\textrm{sub} \sim \{\mathbb{S}_\textrm{1}, \ldots, \mathbb{S}_\textrm{C}\}}\mathbb{E}_{\mathcal{X} \sim \mathbb{S}_\textrm{sub}}[\mathcal{X}] \\
&\leq  \mathbb{D}_{\mathcal{X} \sim \mathbb{S}}[\mathcal{X}]. \\
\end{aligned}
\label{eq:apd_b_lemma1}
\end{equation}
\end{proof}

\begin{lemma}
\label{lemma:gmm_mean_and_variance} 
Consider a Gaussian Mixture Model (GMM) $p_{\text{mix}}(x)$ comprising $\mathbf{C}$ components (\textit{i.e.,} sub-Gaussian distributions). These components are characterized by their means, variances, and weights, denoted as $\{\mu_i\}_{i=1}^{\mathbf{C}}$, $\{\sigma^2_i\}_{i=1}^{\mathbf{C}}$, and $\{\omega_i\}_{i=1}^{\mathbf{C}}$, respectively. The mean $\mathbb{E}[x]$ and variance $\mathbb{D}[x]$ of the distribution are given by $\sum_{i=1}^\mathbf{C}\omega_i\mu_i$ and $\sum_{i=1}^{\mathbf{C}}\omega_i(\mu_i^2+\sigma_i^2) - (\sum_{i=1}^\mathbf{C}\omega_i\mu_i)^2$, respectively~\citep{ostle1963statistics}.
\end{lemma}
\begin{proof}
\begin{equation}
\footnotesize
\begin{aligned}
\mathbb{E}[x] &= \int_{\Theta}x\sum_{i=1}^\mathbf{C}\omega_{i}\frac{1}{\sqrt{2\pi}\sigma_i}e^{-\frac{(x-\mu_i)^2}{2\sigma^2_i}}\\
&= \sum_{i=1}^\mathbf{C}\omega_i \left[\int_{\Theta}x\frac{1}{\sqrt{2\pi}\sigma_i}e^{-\frac{(x-\mu_i)^2}{2\sigma^2_i}}\right]\\
&= \sum_{i=1}^\mathbf{C}\omega_i\mu_i, \\
\mathbb{D}[x] &=  \mathbb{E}[x^2] - \mathbb{E}[x]^2\\
&= \int_{\Theta}x^2\sum_{i=1}^\mathbf{C}\omega_{i}\frac{1}{\sqrt{2\pi}\sigma_i}e^{-\frac{(x-\mu_i)^2}{2\sigma^2_i}} -\mathbb{E}[x]^2\\
&= \sum_{i=1}^\mathbf{C}\omega_{i}\left[\int_{\Theta}x^2\frac{1}{\sqrt{2\pi}\sigma_i}e^{-\frac{(x-\mu_i)^2}{2\sigma^2_i}}\right] -\mathbb{E}[x]^2\\
&= \sum_{i=1}^\mathbf{C}\omega_{i}[\mu_i^2+\sigma^2_i] - (\sum_{i=1}^\mathbf{C}\omega_i\mu_i)^2.\\
\end{aligned}
\label{eq:apd_b_lemma2}
\end{equation}
\end{proof}

\begin{assumption}
\label{ass:gmm}
For any distribution \(Q\), there exists a constant  $\mathbf{C}$ enabling the approximation of \(Q\) by a Gaussian Mixture Model \(P\) with  $\mathbf{C}$ components. More generally, this is expressed as the existence of a  $\mathbf{C}$ such that the distance between \(P\) and \(Q\), denoted by the distance metric function \(\ell(P,Q)\), is bounded above by an infinitesimal \(\epsilon\).
\end{assumption}
\vspace{-5pt}
\textit{Sketch Proof.} The Fourier transform of a Gaussian function does not possess true zeros, indicating that such a function, $f(x)$, along with its shifted variant, $f(x+a)$, densely populates the function space through the Tauberian Theorem. In the context of $L^2$, the space of all square-integrable functions, where Gaussian functions form a subspace denoted as $G$, any linear functional defined on $G$—such as convolution operators—can be extended to all of $L^2$ through the application of the Hahn-Banach Theorem. This extension underscores the completeness of Gaussian Mixture Models (GMM) within $L^2$ spaces.

\paragraph{Remarks.} The proof presents two primary limitations: firstly, it relies solely on shift, which allows the argument to remain valid even when the variances of all components within GMM are identical (a relatively loose condition). Secondly, it imposes an additional constraint by requiring that the coefficients $\omega_i>0$ and $\sum_i \omega_i=1$ in GMM. Accordingly, this study proposes, rather than empirically demonstrates, that GMM can approximate any specified distribution.

\begin{theorem}
\label{tho:mean_and_variance_consistent}
   Given Assumption~\ref{ass:gmm} and Definition~\ref{def:statistical_matching}, the variances and means of $x$ and $y$, estimated through maximum likelihood, remain consistent across scenarios \textit{\textbf{Form (1)}} and \textit{\textbf{Form (2)}}.
\end{theorem}
\begin{proof}
The maximum likelihood estimation mean $\mathbb{E}[x]$ and variance $\mathbb{D}[x]$ of samples $\{x_i\}_{i=1}^N$ within a Gaussian distribution are calculated as $\frac{\sum_{i=1}^N x_i}{N}$ and $\frac{\sum_{i=1}^N(x_i - \mathbb{E}[x])^2}{N}$, respectively. These estimations enable us to characterize the distribution's behavior across different scenarios as follows:

\textit{\textbf{Form (1)}:} $P(x)\sim \mathcal{N}\left(\frac{\sum_{i=1}^Nx_i}{N},\frac{\sum_{i=1}^N\left(x_i-\frac{\sum_{i=1}^Nx_i}{N}\right)^2}{N}\right)$.

\textit{\textbf{Form (2)}:} $Q(y)\sim \sum_i \frac{N_i}{\sum_{j=1}^\mathbf{C}N_j}\mathcal{N}\left(\frac{\sum_{k=1}^{N_i}x^i_k}{N_i},\frac{\sum_{k=1}^{N_i}\left(x^i_k-\frac{\sum_{k=1}^{N_i}x^i_k}{N_i}\right)^2}{N_i}\right)$.

Intuitively, the distilled samples $\{y_i\}_{i=1}^M$ will obey distributions $P(x)$ and $Q(y)$ in scenarios \textit{\textbf{Form (1)}} and \textit{\textbf{Form (2)}}, respectively. Then, the difference of the means between \textit{\textbf{Form (1)}} and \textit{\textbf{Form (2)}} can be derived as
\begin{equation}
\footnotesize
\begin{aligned}
\int_{\Theta} [xP(x)dx - xQ(x)dx] &= \frac{\sum_{i=1}^Nx_i}{N} - \sum_i \frac{N_i}{\sum_{j=1}^\mathbf{C}N_j} \frac{\sum_{k=1}^{N_i}x^i_k}{N_i} \\
& = 0. \\
\end{aligned}
\label{eq:apd_b_theorem4_1}
\end{equation}
To further enhance the explanation on proving the consistency of the variance, the setup introduces two sample sets, $\{x_i\}_{i=1}^{N}$ and $\bigcup_{j=1}^{\mathbf{C}}\{y^j_i\}_{i=1}^{N_j}$, each drawn from their respective distributions, $P(x)$ and $Q(y)$. After that, we can acquire:
\begin{equation}
\footnotesize
\begin{aligned}
\mathbb{D}[x] - \mathbb{D}[y] &= \mathbb{D}[x] - \sum_{i=1}^\mathbf{C}\frac{N_i}{\sum_j N_j} (\mathbb{E}[y^j]^2+\mathbb{D}[y^j]) + \left(\sum_{i=1}^\mathbf{C}\frac{N_i}{\sum_j N_j}\mathbb{E}[y^j]\right)^2 \quad\quad \textcolor{C3}{\#\ Lemma~\ref{lemma:gmm_mean_and_variance}} \\
& = \mathbb{D}[x] - \mathbb{E}[\mathbb{E}[y^j]^2] - \mathbb{E}[\mathbb{D}[y^j]] + \mathbb{E}[\mathbb{E}[y^j]]^2 \\
& = (\mathbb{D}[x] - \mathbb{E}[\mathbb{D}[y^j]]) 
 - \mathbb{E}[\mathbb{E}[y^j]^2] + \mathbb{E}[\mathbb{E}[y^j]]^2 \\
& = \mathbb{D}[\mathbb{E}[y^j]] - \mathbb{E}[\mathbb{E}[y^j]^2] + \mathbb{E}[\mathbb{E}[y^j]]^2  \quad\quad \textcolor{C3}{\#\ Lemma~\ref{lemma:variance_error}} \\
& = 0. \\
\end{aligned}
\label{eq:apd_b_theorem4_2}
\end{equation}
\end{proof}

\begin{corollary}
\label{cor:same}
The mean and variance obtained from maximum likelihood for any split form $\{c_1,c_2,\ldots,c_\mathbf{C}\}$ in \textit{\textbf{Form (2)}} remain consistent.
\end{corollary}
\vspace{-5pt}
\textit{Sketch Proof.} According to Theorem~\ref{tho:mean_and_variance_consistent} the mean and variance obtained from maximum likelihood for each split form in \textit{\textbf{Form (2)}} remain consistent within \textit{\textbf{Form (1)}}, so that any split form $\{c_1, c_2, \ldots, c_\mathbf{C}\}$ in \textit{\textbf{Form (2)}} remain consistent.

\begin{theorem}
\label{the:entropy}
Based on Definition~\ref{def:statistical_matching}, the entropy—pertaining to diversity—of the distributions characterized as $\mathcal{H}(P)$ from \textit{\textbf{Form (1)}} and $\mathcal{H}(Q)$ from \textit{\textbf{Form (2)}}, which are estimated through maximum likelihood, exhibits the subsequent relationship: $\mathcal{H}(P)- \frac{1}{2}\left[\log(\mathbb{E}[\mathbb{D}[y^j]]+\mathbb{D}[\mathbb{E}[y^j]])-\mathbb{E}[\log(\mathbb{D}[y^j])]\right] \leq \mathcal{H}(Q) \leq \mathcal{H}(P)+\frac{1}{4}\mathbb{E}_{(i,j)\sim \prod[\mathbf{C},\mathbf{C}]}\left[\frac{(\mathbb{E}[y^i]-\mathbb{E}[y^j])^2(\mathbb{D}[y^i]+\mathbb{D}[y^j])}{\mathbb{D}[y^i]\mathbb{D}[y^j]}\right]$. The two-sided equality (\textit{i.e.,} $\mathcal{H}(P)\equiv \mathcal{H}(Q)$) holds if and only if both the variance and the mean of each component are consistent.
\end{theorem}
\begin{proof}

\begin{equation}
\footnotesize
\begin{aligned}
&\textcolor{C3}{\#\textrm{Lower bound:}} \\
&\mathbb{E}[-\log(P(x))] - \mathbb{E}[-\log(Q(y))] \\
&= \int_{\Theta}-\log(P(x))P(x)dx + \int_{\Theta}\log(P(y))P(y)dy\\
&= \frac{1}{2}\log(2\pi\mathbb{D}[x])+\frac{1}{2} + \int_{\Theta} \log(\int_j p(y^j)\frac{1}{\sqrt{2\pi\mathbb{D}[y^j]}}e^{\frac{(y-\mathbb{E}[y^j])^2}{-2\mathbb{D}[y^j]}}dj)(\int_j p(y^j)\frac{1}{\sqrt{2\pi\mathbb{D}[y^j]}}e^{\frac{(y-\mathbb{E}[y^j])^2}{-2\mathbb{D}[y^j]}}dj) dy\\
&=\frac{1}{2}\log(2\pi\mathbb{D}[x])+\frac{1}{2} + \int_{\Theta} \log(\mathbb{E}[\frac{1}{\sqrt{2\pi\mathbb{D}[y^j]}}e^{\frac{(y-\mathbb{E}[y^j])^2}{-2\mathbb{D}[y^j]}}])\mathbb{E}[\frac{1}{\sqrt{2\pi\mathbb{D}[y^j]}}e^{\frac{(y-\mathbb{E}[y^j])^2}{-2\mathbb{D}[y^j]}}] dy\\
&\geq \frac{1}{2}\log(2\pi\mathbb{D}[x])+\frac{1}{2} + \int_{\Theta} \mathbb{E}[\log(\frac{1}{\sqrt{2\pi\mathbb{D}[y^j]}}e^{\frac{(y-\mathbb{E}[y^j])^2}{-2\mathbb{D}[y^j]}})]\mathbb{E}[\frac{1}{\sqrt{2\pi\mathbb{D}[y^j]}}e^{\frac{(y-\mathbb{E}[y^j])^2}{-2\mathbb{D}[y^j]}}] dy\\
&=\frac{1}{2}\log(2\pi\mathbb{D}[x])+\frac{1}{2} + \mathbb{E}_{(i,j)\sim \prod[\mathbf{C},\mathbf{C}]}\left[\int_{\Theta} \log(\frac{1}{\sqrt{2\pi\mathbb{D}[y^i]}}e^{\frac{(y-\mathbb{E}[y^i])^2}{-2\mathbb{D}[y^i]}})(\frac{1}{\sqrt{2\pi\mathbb{D}[y^j]}}e^{\frac{(y-\mathbb{E}[y^j])^2}{-2\mathbb{D}[y^j]}})dy\right] \\
&=\frac{1}{2}\log(2\pi\mathbb{D}[x])+\frac{1}{2} -\mathbb{E}_{(i,j)\sim \prod[\mathbf{C},\mathbf{C}]}\left[\frac{1}{2}\log(2\pi\mathbb{D}[y^j])+\frac{\mathbb{D}[y^i]+(\mathbb{E}[y^i]-\mathbb{E}[y^j])^2}{2\mathbb{D}[y^j]}\right] \\
&\geq \frac{1}{2}\log(2\pi\mathbb{D}[x]) - \frac{1}{2}\log(\mathbb{E}[2\pi\mathbb{D}[y^j]]) + \frac{1}{2} -\mathbb{E}_{(i,j)\sim \prod[\mathbf{C},\mathbf{C}]}\left[\frac{\mathbb{D}[y^i]+(\mathbb{E}[y^i]-\mathbb{E}[y^j])^2}{2\mathbb{D}[y^j]}\right]\\
&\geq - \frac{1}{4}\mathbb{E}_{(i,j)\sim \prod[\mathbf{C},\mathbf{C}]}\left[\frac{(\mathbb{E}[y^i]-\mathbb{E}[y^j])^2(\mathbb{D}[y^i]+\mathbb{D}[y^j])}{\mathbb{D}[y^i]\mathbb{D}[y^j]}\right] \\
& \\
&\textcolor{C3}{\#\textrm{Upper bound:}} \\
&\mathbb{E}[-\log(P(x))] - \mathbb{E}[-\log(Q(y))] \\
&= \int_{\Theta}-\log(P(x))P(x)dx + \int_{\Theta}\log(P(y))P(y)dy\\
&=  \int_{\Theta}-\log(P(x))P(x)dx + \int_{\Theta} \log(\mathbb{E}[\frac{1}{\sqrt{2\pi\mathbb{D}[y^j]}}e^{\frac{(y-\mathbb{E}[y^j])^2}{-2\mathbb{D}[y^j]}}])\mathbb{E}[\frac{1}{\sqrt{2\pi\mathbb{D}[y^j]}}e^{\frac{(y-\mathbb{E}[y^j])^2}{-2\mathbb{D}[y^j]}}] dy\\
&\leq \int_{\Theta}-\log(P(x))P(x)dx + \mathbb{E}[\int_{\Theta} \log(\frac{1}{\sqrt{2\pi\mathbb{D}[y^j]}}e^{\frac{(y-\mathbb{E}[y^j])^2}{-2\mathbb{D}[y^j]}})\frac{1}{\sqrt{2\pi\mathbb{D}[y^j]}}e^{\frac{(y-\mathbb{E}[y^j])^2}{-2\mathbb{D}[y^j]}}dy]\\
&= \frac{1}{2}\log(2\pi\mathbb{D}[x]) - \mathbb{E}[\frac{1}{2}\log(2\pi\mathbb{D}[y^j])] \\
& = \frac{1}{2}\left[\log(\mathbb{E}[\mathbb{D}[y^j]]+\mathbb{D}[\mathbb{E}[y^j]])-\mathbb{E}[\log(\mathbb{D}[y^j])]\right] \\
\end{aligned}
\label{eq:apd_b_theorem6_2}
\end{equation}
\end{proof}

\begin{theorem}
\label{the:kl}
Based on Definition~\ref{def:statistical_matching}, if the original distribution is $p_\textrm{mix}$, the Kullback-Leibler divergence $D_\textrm{KL}[p_\textrm{mix}||Q]$ has a upper bound $\mathbb{E}_{i\sim \mathcal{U}[1,\ldots,\mathbf{C}]}\mathbb{E}_{j\sim \mathcal{U}[1,\ldots,\mathbf{C}]}\frac{\mathbb{E}[y^j]^2}{\mathbb{D}[y^i]}$ and $D_\textrm{KL}[p_\textrm{mix}||P]=0$.
\end{theorem}
\begin{proof}
\begin{equation}
\footnotesize
\begin{aligned}
&D_\textrm{KL}[Q||P] \\
&= D_\textrm{KL}\left[\sum_i \frac{N_i}{\sum_{j=1}^\mathbf{C}N_j}\mathcal{N}\left(\frac{\sum_{k=1}^{N_i}x^i_k}{N_i},\frac{\sum_{k=1}^{N_i}\left(x^i_k-\frac{\sum_{k=1}^{N_i}x^i_k}{N_i}\right)^2}{N_i}\right)\bigg\|\mathcal{N}\left(\frac{\sum_{i=1}^Nx_i}{N},\frac{\sum_{i=1}^N\left(x_i-\frac{\sum_{i=1}^Nx_i}{N}\right)^2}{N}\right)\right] \\
&\leq \sum_i \frac{N_i}{\sum_{j=1}^\mathbf{C}N_j} D_\textrm{KL}\left[\mathcal{N}\left(\frac{\sum_{k=1}^{N_i}x^i_k}{N_i},\frac{\sum_{k=1}^{N_i}\left(x^i_k-\frac{\sum_{k=1}^{N_i}x^i_k}{N_i}\right)^2}{N_i}\right)\bigg\|\mathcal{N}\left(\frac{\sum_{i=1}^Nx_i}{N},\frac{\sum_{i=1}^N\left(x_i-\frac{\sum_{i=1}^Nx_i}{N}\right)^2}{N}\right)\right]. \\
\end{aligned}
\label{eq:apd_b_theorem7}
\end{equation}
By applying the notations from Lemma~\ref{lemma:gmm_mean_and_variance} for convenience, we obtain:
\begin{equation}
\footnotesize
\begin{aligned}
&D_\textrm{KL}[Q||P] \\
&\leq \sum_i \omega_i \left[\frac{1}{2}\log\left(\frac{\sum_{j=1}^\mathbf{C}\omega_j[\mu_j^2+\sigma_j^2]-(\sum_{j=1}^\mathbf{C}\omega_j\mu_j)^2}{\sigma_i^2}\right)+\frac{\sum_{j=1}^\mathbf{C}\omega_j[\mu_j^2+\sigma_j^2]-(\sum_{j=1}^\mathbf{C}\omega_j\mu_j)^2}{2\sigma_i^2}\right] - \frac{1}{2} \\
&\leq \frac{1}{2}\log\left(\sum_i \omega_i \frac{\sum_{j=1}^\mathbf{C}\omega_j[\mu_j^2+\sigma_j^2]-(\sum_{j=1}^\mathbf{C}\omega_j\mu_j)^2}{\sigma_i^2}\right)+\frac{1}{2}\sum_i \omega_i \frac{\sum_{j=1}^\mathbf{C}\omega_j[\mu_j^2+\sigma_j^2]-(\sum_{j=1}^\mathbf{C}\omega_j\mu_j)^2}{\sigma_i^2} - \frac{1}{2} \\
&\leq \frac{1}{2}\log\left(1+\sum_i \omega_i \frac{\sum_{j=1}^\mathbf{C}\omega_j\mu_j^2-(\sum_{j=1}^\mathbf{C}\omega_j\mu_j)^2}{\sigma_i^2}\right)+\frac{1}{2}\sum_i \omega_i \frac{\sum_{j=1}^\mathbf{C}\omega_j\mu_j^2-(\sum_{j=1}^\mathbf{C}\omega_j\mu_j)^2}{\sigma_i^2} \\
&\leq \frac{1}{2}\log\left(1+\sum_i\sum_j\omega_i \omega_j\frac{\mu_j^2}{\sigma_i^2}\right)+\frac{1}{2}\sum_i\sum_j\omega_i \omega_j\frac{\mu_j^2}{\sigma_i^2}\\
&\leq \mathbb{E}_{i\sim \mathcal{U}[1,\ldots,\mathbf{C}]}\mathbb{E}_{j\sim \mathcal{U}[1,\ldots,\mathbf{C}]}\frac{\mathbb{E}[y^j]^2}{\mathbb{D}[y^i]}.\\
\end{aligned}
\label{eq:apd_b_theorem7_2}
\end{equation}
\end{proof}
When the sample size is sufficiently large, the original distribution aligns with $Q$. Consequently, we obtain $D_\textrm{KL}[p_\textrm{mix}||P] \leq \mathbb{E}_{i\sim \mathcal{U}[1,\ldots,\mathbf{C}]}\mathbb{E}_{j\sim \mathcal{U}[1,\ldots,\mathbf{C}]}\frac{\mathbb{E}[y^j]^2}{\mathbb{D}[y^i]}$ and establish that $D_\textrm{KL}[p_\textrm{mix}||Q] = 0$.

\section{Decoupled Optimization Objective of Dataset Condensation}
\label{apd:decouple}
In this section, we demonstrate that the training objective, as defined in Eq.~\ref{eq:definition_sm}, can be decoupled into two components—flatness and closeness—using a second-order Taylor expansion, under the assumption that $\mathcal{L}_\textbf{\textrm{syn}}\in \mathbf{C}^2(\mathbf{I},\mathbb{R})$. We define the closest optimization point $\mathbf{o}_i$ for $\mathcal{X}^{\mathcal{S}}$ in relation to the $i$-th matching operator $\mathcal{L}^i_\textbf{\textrm{syn}}(\cdot,\cdot)$. This framework can accommodate all matchings related to $f^i(\cdot)$, including gradient matching\citep{dd_gradient_matching}, trajectory matching~\citep{dd_mtt}, distribution matching~\citep{dd_dist_matching}, and statistical matching~\citep{shao2023generalized}. Consequently, we derive the dual decoupling of flatness and closeness as follows:
\begin{equation}
\footnotesize
\begin{aligned}
 \mathcal{L}_\textbf{\textrm{DD}} & = \mathbb{E}_{\mathcal{L}_\textbf{\textrm{syn}}(\cdot,\cdot)\sim \mathbb{S}_\textrm{match}}[\mathcal{L}_\textbf{\textrm{syn}}(\mathcal{X}^{\mathcal{S}},\mathcal{X}^{\mathcal{T}})] = \frac{1}{|\mathbb{S}_\textrm{match}|}\sum_{i=1}^{|\mathbb{S}_\textrm{match}|}[\mathcal{L}^i_\textbf{\textrm{syn}}(\mathcal{X}^{\mathcal{S}},\mathcal{X}^{\mathcal{T}})] \\
 & = \frac{1}{|\mathbb{S}_\textrm{match}|}\sum_{i=1}^{|\mathbb{S}_\textrm{match}|}[\mathcal{L}^i_\textbf{\textrm{syn}}(\textbf{o}_i,\mathcal{X}^{\mathcal{T}}) + (\mathcal{X}^{\mathcal{S}}-\mathbf{o}_i)\nabla_{\mathcal{X}^{\mathcal{S}}}\mathcal{L}^i_\textbf{\textrm{syn}}(\textbf{o}_i,\mathcal{X}^{\mathcal{T}})+(\mathcal{X}^{\mathcal{S}}-\mathbf{o}_i)^T\mathbf{\mathrm{H}}^i(\mathcal{X}^{\mathcal{S}}-\mathbf{o}_i)] + \mathcal{O}((\mathcal{X}^{\mathcal{S}}-\mathbf{o}_i)^3) \\
 & = \frac{1}{|\mathbb{S}_\textrm{match}|}\sum_{i=1}^{|\mathbb{S}_\textrm{match}|}[\mathcal{L}^i_\textbf{\textrm{syn}}(\textbf{o}_i,\mathcal{X}^{\mathcal{T}}) + (\mathcal{X}^{\mathcal{S}}-\mathbf{o}_i)^T\mathbf{\mathrm{H}}^i(\mathcal{X}^{\mathcal{S}}-\mathbf{o}_i)], \\
\end{aligned}
\label{eq:apd_3}
\end{equation}
where $\mathbf{\mathrm{H}}^i$ refers to the Hessian matrix of $\mathcal{L}^i_\textbf{\textrm{syn}}(\cdot,\mathcal{X}^{\mathcal{T}})$ at the closest optimization point $\mathbf{o}_i$. Note that as the optimization method for deep learning typically involves gradient descent-like approaches (\textit{e.g.,} SGD and AdamW), the first-order derivative $\nabla_{\mathcal{X}^{\mathcal{S}}}\mathcal{L}^i_\textbf{\textrm{syn}}(\textbf{o}_i,\mathcal{X}^{\mathcal{T}})$ can be directly discarded. After that, scanning the two terms in Eq.~\ref{eq:apd_3}, the first one necessarily reaches an optimal solution, while the second one allows us to obtain an upper definitive bound on the Hessian matrix and Jacobi matrix through Theorem 3.1 outlined in~\cite{cwa}. Here, we give a special case under the $\ell_2$-norm to discard the assumption that $\mathbf{\mathrm{H}}^i$ and $(\mathcal{X}^{\mathcal{S}} - \mathbf{o}_i)$ are independent:

\begin{theorem} (improved from Theorem 3.1 in~\citep{cwa}) $\frac{1}{|\mathbb{S}_\textrm{match}|}\sum_{i=1}^{|\mathbb{S}_\textrm{match}|}(\mathcal{X}^{\mathcal{S}}-\mathbf{o}_i)^T\mathbf{\mathrm{H}}^i(\mathcal{X}^{\mathcal{S}}-\mathbf{o}_i)$ $\leq$ $|\mathbb{S}_\textrm{match}|\cdot\mathbb{E}[||\mathbf{\mathrm{H}}^i||_\mathrm{F}]\mathbb{E}[||\mathcal{X}^{\mathcal{S}} - \mathbf{o}_i||_2^2]$, where $\mathbb{E}[||\mathbf{\mathrm{H}}^i||_\mathrm{F}]$ and $\mathbb{E}[||\mathcal{X}^{\mathcal{S}} - \mathbf{o}_i||_2^2]$ denote flatness and closeness, respectively.
\begin{proof}
\begin{equation}
\footnotesize
\begin{aligned}
 & \frac{1}{|\mathbb{S}_\textrm{match}|}\sum_{i=1}^{|\mathbb{S}_\textrm{match}|}(\mathcal{X}^{\mathcal{S}}-\mathbf{o}_i)^T\mathbf{\mathrm{H}}^i(\mathcal{X}^{\mathcal{S}}-\mathbf{o}_i) \leq \frac{1}{|\mathbb{S}_\textrm{match}|}\sum_{i=1}^{|\mathbb{S}_\textrm{match}|}[||(\mathcal{X}^{\mathcal{S}} - \mathbf{o}_i)||_2 ||\mathbf{\mathrm{H}}^i(\mathcal{X}^{\mathcal{S}} - \mathbf{o}_i)||_{2}]\quad\quad \textcolor{C3}{\#\ \textrm{H\"older's inequality}} \\ 
 & = \frac{1}{|\mathbb{S}_\textrm{match}|}\sum_{i=1}^{|\mathbb{S}_\textrm{match}|}[||(\mathcal{X}^{\mathcal{S}} - \mathbf{o}_i)||_2 ||\mathbf{\mathrm{H}}^i||_{2,2}||(\mathcal{X}^{\mathcal{S}} - \mathbf{o}_i)||_{2}]\quad\quad \textcolor{C3}{\#\ \textrm{Definition of matrix norm}}\\
 & \leq |\mathbb{S}_\textrm{match}|\cdot\mathbb{E}[||\mathbf{\mathrm{H}}^i||_{2,2}]\mathbb{E}[||\mathcal{X}^{\mathcal{S}} - \mathbf{o}_i||_2^2] \leq |\mathbb{S}_\textrm{match}|\cdot\mathbb{E}[||\mathbf{\mathrm{H}}^i||_\textrm{F}]\mathbb{E}[||\mathcal{X}^{\mathcal{S}} - \mathbf{o}_i||_2^2] \\
\end{aligned}
\label{eq:apd_4}
\end{equation}
\end{proof}
\end{theorem}

Actually, flatness can be ensured by convergence in a flat region through sharpness-aware minimization (SAM) theory~\citep{iclr2020_sam,sam_llm,nips2022_sam,eccvw_sam}. Specifically, a body of work on SAM has established a connection between the Hessian matrix and the flatness of the loss landscape (\textit{i.e.,} the curvature of the loss trajectory), with a series of empirical studies demonstrating the theory's reliability. Meanwhile, the specific implementation of flatness is elaborated upon in Sec.~\ref{apd:sharpness}. By contrast, the concept of closeness was first introduced in~\cite{cwa}, where it is observed that utilizing more backbones for ensemble can result in a smaller generalization error during the evaluation phase. In fact, closeness has been implicitly implemented since our \textit{baseline} G-VBSM uses a sequence optimization mechanism akin to the official implementation in~\cite{cwa}. Therefore, this paper will not elucidate on closeness and its specific implementation.

\section{Traditional Sharpness-Aware Minimization Optimization Approach}
\label{apd:trad_sam}
For the comprehensive of our paper, let us give a brief yet formal description of sharpness-aware minimization (SAM). The applicable SAM algorithm was first proposed in~\cite{iclr2020_sam}, which aims to solve the following maximum minimization problem:
\begin{equation}
\small
\begin{aligned}
&\min_{\theta}\max_{\epsilon:||\epsilon||\leq \rho}L_{\mathbb{S}}(f_{\theta+\epsilon}),\\
\end{aligned}
\label{eq:begin_sam}
\end{equation}
where $L_\mathbb{S}(f_\theta)$, $\epsilon$, $\rho$, and $\theta$ refer to the loss $\frac{1}{|\mathbb{S}|}\sum_{x_i,y_i \sim \mathbb{S}}\ell(f_{\theta}(x_i),y_i)$, the perturbation, the pre-defined flattened region, and the model parameter, respectively. Let us define the final optimized model parameters as $\theta^*$, then the optimization objective can be rewritten as
\begin{equation}
\small
\begin{aligned}
&{\theta^*} = \operatorname*{arg\,min}_{\theta}R_{\mathbb{S}}(f_{\theta})+L_{\mathbb{S}}(f_\theta),\ \textrm{where}\ R_{\mathbb{S}}(f_{\theta}) = \max_{\epsilon:||\epsilon||\leq \rho} L_{\mathbb{S}}(f_{\theta+\epsilon}) - L_{\mathbb{S}}(f_{\theta}).\\
\end{aligned}
\label{eq:rewritten_sam}
\end{equation}
By expanding $L_{\mathbb{S}}(f_{\theta+\epsilon})$ at $\theta$ and by solving the classical \textit{dual norm} problem, the first maximization objective can be solved as (In the special case of the $\ell_2$-norm)
\begin{equation}
\small
\begin{aligned}
&{\epsilon^*} = \operatorname*{arg\,max}_{\epsilon:||\epsilon||\leq \rho}L_{\mathbb{S}}(f_{\theta+\epsilon})\approx \rho \frac{\nabla_\theta L_{\mathbb{S}}(f_{\theta})}{||\nabla_\theta L_{\mathbb{S}}(f_{\theta})||_2}.\\
\end{aligned}
\label{eq:solve1_sam}
\end{equation}
The specific derivation is as follows:
\begin{proof}
Subjecting $L_{\mathbb{S}}(f_{\theta+\epsilon})$ to a Taylor expansion and retaining only the first-order derivatives:
\begin{equation}
\small
\begin{aligned}
&R_{\mathbb{S}}(f_\theta) = L_{\mathbb{S}}(f_{\theta+\epsilon}) - L_{\mathbb{S}}(f_{\theta}) \approx  L_{\mathbb{S}}(f_{\theta}) + \epsilon^T \nabla_{\theta} L_{\mathbb{S}}(f_{\theta})  -  L_{\mathbb{S}}(f_{\theta}) = \epsilon^T \nabla_{\theta} L_{\mathbb{S}}(f_{\theta}).\\
\end{aligned}
\label{eq:proof_sam_1}
\end{equation}
Then, we can get
\begin{equation}
\small
\begin{aligned}
&{\epsilon^*} = \operatorname*{arg\,max}_{\epsilon:||\epsilon||\leq \rho}L_{\mathbb{S}}(f_{\theta+\epsilon})-L_{\mathbb{S}}(f_{\theta}) = \operatorname*{arg\,max}_{\epsilon:||\epsilon||\leq \rho}\ \left[\epsilon^T \nabla_{\theta} L_{\mathbb{S}}(f_{\theta})\right].\\
\end{aligned}
\label{eq:proof_sam_2}
\end{equation}
Next, we base our solution on the solution of the classical \textit{dual norm} problem, where the above equation can be written as $||\nabla_\theta L_{\mathbb{S}}(f_\theta)||_*$. Firstly, H\"older's inequality gives
\begin{equation}
\small
\begin{aligned}
&\epsilon^T \nabla_{\theta} L_{\mathbb{S}}(f_{\theta}) = \sum_{i=1}^n \epsilon^T_i \nabla_{\theta} L_{\mathbb{S}}(f_{\theta})_i \leq \sum_{i=1}^n |\epsilon^T_i \nabla_{\theta} L_{\mathbb{S}}(f_{\theta})_i| \\
&\leq  ||\epsilon^T \nabla_{\theta} L_{\mathbb{S}}(f_{\theta})||_1 \leq ||\epsilon^T ||_p|| \nabla_{\theta}L_{\mathbb{S}}(f_{\theta})||_q \leq \rho || \nabla_{\theta}L_{\mathbb{S}}(f_{\theta})||_q.\\
\end{aligned}
\label{eq:proof_sam_3}
\end{equation}
So, we just need to find a $\epsilon$ that makes all the above inequality signs equal. Define $m$ as $\textrm{sign}(\nabla_{\theta} L_{\mathbb{S}}(f_{\theta})) |\nabla_{\theta} L_{\mathbb{S}}(f_{\theta})|^{q-1}$, then we can rewritten Eq.~\ref{eq:proof_sam_3} as
\begin{equation}
\small
\begin{aligned}
\epsilon^T \nabla_{\theta} L_{\mathbb{S}}(f_{\theta}) &= \sum_{i=1}^n\textrm{sign}(\nabla_{\theta} L_{\mathbb{S}}(f_{\theta})_i) |\nabla_{\theta} L_{\mathbb{S}}(f_{\theta})_i|^{q-1}\nabla_{\theta} L_{\mathbb{S}}(f_{\theta})_i\\
&= \sum_{i=1}^n|\nabla_{\theta} L_{\mathbb{S}}(f_{\theta})_i| |\nabla_{\theta} L_{\mathbb{S}}(f_{\theta})_i|^{q-1} \\
&= ||\nabla_{\theta} L_{\mathbb{S}}(f_{\theta})||^{q}_q. \\
\end{aligned}
\label{eq:proof_sam_4}
\end{equation}
And we also get
\begin{equation}
\small
\begin{aligned}
&||\epsilon||_p^p = \sum_{i=1}^n |\epsilon|^p =  \sum_{i=1}^n |\textrm{sign}(\nabla_{\theta} L_{\mathbb{S}}(f_{\theta}))|\nabla_{\theta} L_{\mathbb{S}}(f_{\theta})|^{q-1}|^p = ||\nabla_{\theta} L_{\mathbb{S}}(f_{\theta})||_q^q, \\
\end{aligned}
\label{eq:proof_sam_5}
\end{equation}
where $1/p+1/q=1$. We choose a new $\epsilon$, defined as $y=\rho\frac{\epsilon}{||\epsilon||_p}$, which satisfies: $||y||_p=\rho$, and substitute into $\epsilon^T \nabla_{\theta} L_{\mathbb{S}}(f_{\theta})$:
\begin{equation}
\small
\begin{aligned}
&y^T \nabla_{\theta} L_{\mathbb{S}}(f_{\theta})= \sum_{i=1}^n y_i \nabla_{\theta} L_{\mathbb{S}}(f_{\theta})_i = \sum_{i=1}^n \frac{\rho \nabla_{\theta} L_{\mathbb{S}}(f_{\theta})_i}{||\nabla_{\theta} L_{\mathbb{S}}(f_{\theta})||_p}\nabla_{\theta} L_{\mathbb{S}}(f_{\theta})_i =  \frac{\rho}{||\epsilon||_p} \sum_{i=1}^n \epsilon_i\nabla_{\theta} L_{\mathbb{S}}(f_{\theta})_i. \\
\end{aligned}
\label{eq:proof_sam_6}
\end{equation}
Due to $||\epsilon||_p = ||\nabla_{\theta} L_{\mathbb{S}}(f_{\theta})_i||_q^{q/p}$ and $\epsilon^T \nabla_{\theta} L_{\mathbb{S}}(f_{\theta})=||\nabla_{\theta} L_{\mathbb{S}}(f_{\theta})||^{q}_q$, we can further derive and obtain that
\begin{equation}
\small
\begin{aligned}
&\frac{\rho}{||\epsilon||_p} \sum_{i=1}^n \epsilon_i\nabla_{\theta} L_{\mathbb{S}}(f_{\theta})_i = \frac{\rho}{||\nabla_{\theta} L_{\mathbb{S}}(f_{\theta}) ||_q^{q/p}} \sum_{i=1}^n \epsilon_i\nabla_{\theta} L_{\mathbb{S}}(f_{\theta})_i =\rho||\nabla_{\theta}L_{\mathbb{S}}(f_{\theta})||_q. \\
\end{aligned}
\label{eq:proof_sam_7}
\end{equation}
Therefore, $y$ can be rewritten as:
\begin{equation}
\small
\begin{aligned}
&y=\rho\frac{ \textrm{sign}(\nabla_{\theta} L_{\mathbb{S}}(f_{\theta})) |\nabla_{\theta} L_{\mathbb{S}}(f_{\theta})|^{q-1}}{|| \textrm{sign}(\nabla_{\theta} L_{\mathbb{S}}(f_{\theta})) |\nabla_{\theta} L_{\mathbb{S}}(f_{\theta})|^{q-1}||_p} = \rho\frac{ \textrm{sign}(\nabla_{\theta} L_{\mathbb{S}}(f_{\theta})) |\nabla_{\theta} L_{\mathbb{S}}(f_{\theta})|^{q-1}}{||\nabla_{\theta} L_{\mathbb{S}}(f_{\theta})||_{q}^{q-1}}. \\
\end{aligned}
\label{eq:proof_sam_8}
\end{equation}
If $q=2$, $y = \rho\frac{\nabla_{\theta} L_{\mathbb{S}}(f_{\theta})}{||\nabla_{\theta} L_{\mathbb{S}}(f_{\theta})||_{2}}$.
\end{proof}
The above derivation is partly derived from~\cite{iclr2020_sam}, to which we have added another part. To solve the SAM problem in deep learning~\citep{iclr2020_sam}, had to require two iterations to complete a single SAM-based gradient update. Another pivotal aspect to note is that within the context of dataset condensation, $\theta$ transitions from representing the model parameter $f_\theta$ to denoting the synthesized dataset $\mathcal{X}^{\mathcal{S}}$.

\section{Implementation of Flatness Regularization}
\label{apd:sharpness}
As proved in Sec.~\ref{apd:trad_sam}, the optimal solution $\epsilon^*$ is denoted as $\rho \frac{\nabla_\theta L_{\mathbb{S}}(f_{\theta})}{||\nabla_\theta L_{\mathbb{S}}(f_{\theta})||_2}$. Analogously, in the dataset condensation scenario, the joint optimization objective is given by $\sum_{i=1}^{|\mathbb{S}_\textrm{match}|}[\mathcal{L}^i_\textbf{\textrm{syn}}(\mathcal{X}^{\mathcal{S}},\mathcal{X}^{\mathcal{T}})]$. There exists an optimal $\epsilon^*$, which can be written as $\rho\frac{\nabla_{\mathcal{X}^{\mathcal{S}}}\sum_{i=1}^{|\mathbb{S}_\textrm{match}|}[\mathcal{L}^i_\textbf{\textrm{syn}}(\mathcal{X}^{\mathcal{S}},\mathcal{X}^{\mathcal{T}})]}{||\nabla_{\mathcal{X}^{\mathcal{S}}}\sum_{i=1}^{|\mathbb{S}_\textrm{match}|}[\mathcal{L}^i_\textbf{\textrm{syn}}(\mathcal{X}^{\mathcal{S}},\mathcal{X}^{\mathcal{T}})]||_2}$. Thus, a dual-stage approach of flatness regularization is shown below:
\begin{equation}
\small
\begin{aligned}
& \mathcal{X}^{\mathcal{S}}_\textbf{\textrm{new}} \leftarrow \mathcal{X}^{\mathcal{S}} + \frac{\rho}{||\nabla_{\mathcal{X}^{\mathcal{S}}}\sum_{i=1}^{|\mathbb{S}_\textrm{match}|}[\mathcal{L}^i_\textbf{\textrm{syn}}(\mathcal{X}^{\mathcal{S}},\mathcal{X}^{\mathcal{T}})]||_2}\left(\nabla_{\mathcal{X}^{\mathcal{S}}}\sum_{i=1}^{|\mathbb{S}_\textrm{match}|}[\mathcal{L}^i_\textbf{\textrm{syn}}(\mathcal{X}^{\mathcal{S}},\mathcal{X}^{\mathcal{T}})]\right) \\
& \mathcal{X}^{\mathcal{S}}_\textbf{\textrm{next}} \leftarrow \mathcal{X}^{\mathcal{S}}_\textbf{\textrm{new}} - \eta \left(\nabla_{\mathcal{X}^{\mathcal{S}}_\textbf{\textrm{new}}}\sum_{i=1}^{|\mathbb{S}_\textrm{match}|}[\mathcal{L}^i_\textbf{\textrm{syn}}(\mathcal{X}^{\mathcal{S}}_\textbf{\textrm{new}},\mathcal{X}^{\mathcal{T}})]\right),\\
\end{aligned}
\label{eq:proof_sam_9}
\end{equation}
where $\eta$ and $\mathcal{X}^{\mathcal{S}}_\textbf{\textrm{next}}$ denote the learning rate and the synthesized dataset in the next iteration, respectively. However, this optimization approach significantly increases the computational burden, thus reducing its scalability. Enlightened by~\cite{nips2022_sam}, we consider a single-stage optimization strategy implemented via exponential moving average (EMA). Given an EMA-updated synthesized dataset $\mathcal{X}^{\mathcal{S}}_\textbf{\textrm{EMA}} = \beta \mathcal{X}^{\mathcal{S}}_\textbf{\textrm{EMA}} + (1-\beta)\mathcal{X}^{\mathcal{S}}$, where $\beta$ is typically set to 0.99 in our experiments. The trajectories of the synthesized datasets updated via gradient descent (GD) and EMA can be represented as $\{\theta^0_\textbf{\textrm{GD}},\theta^1_\textbf{\textrm{GD}},\cdots,\theta^N_\textbf{\textrm{GD}}\}$ and $\{\theta^0_\textbf{\textrm{EMA}},\theta^1_\textbf{\textrm{EMA}},\cdots,\theta^N_\textbf{\textrm{EMA}}\}$, respectively. Assume that $\ve{g}_j = \nabla_{\mathcal{X}^{\mathcal{S}}}\sum_{i=1}^{|\mathbb{S}_\textrm{match}|}[\mathcal{L}^i_\textbf{\textrm{syn}}(\mathcal{X}^{\mathcal{S}},\mathcal{X}^{\mathcal{T}})]$ at the $j$-th iteration, then $\theta^j_\textbf{\textrm{EMA}}=\theta^j_\textbf{\textrm{GD}} + \sum_{i=1}^{j-1}\beta^{j-i}\ve{g}_i$ with the condition $1\leq j\leq N$\footnote{Neglecting the learning rate for simplicity does not affect the derivation.}, as outlined in~\cite{nips2022_sam}. Consequently, we can provide the EMA-based SAM algorithm and applied to backbone sequential optimization in dataset condensation as follows:
\begin{equation}
\small
\begin{aligned}
& \mathcal{L}_\textbf{\textrm{FR}} = \sum_{i=1}^{|\mathbb{S}_\textrm{match}|}[\mathcal{L}^i_\textbf{\textrm{syn}}(\mathcal{X}^{\mathcal{S}},\mathcal{X}^{\mathcal{S}}_\textbf{\textrm{EMA}})] = \sum_{i=1}^{|\mathbb{S}_\textrm{match}|}[\mathcal{L}^i_\textbf{\textrm{syn}}(\theta^j_\textbf{\textrm{GD}},\theta^j_\textbf{\textrm{EMA}})],\quad\quad \textrm{at the}\ j\textrm{-th iteration}.\\
\end{aligned}
\label{eq:proof_sam_10}
\end{equation}
In the vast majority of dataset distillation algorithms~\citep{yin2024dataset,shao2023generalized,zhou2024self}, the metric function used in matching is set to mean squared error (MSE) loss. Based on this phenomenon, we can rewrite Eq.~\ref{eq:proof_sam_10} to Eq.~\ref{eq:proof_sam_11}, which guarantees flatness.
\begin{equation}
\small
\begin{aligned}
& \nabla_{\theta^j_\textbf{\textrm{GD}}}\sum_{i=1}^{|\mathbb{S}_\textrm{match}|}[\mathcal{L}^i_\textbf{\textrm{syn}}(\theta^j_\textbf{\textrm{GD}},\theta^j_\textbf{\textrm{EMA}})],\quad\quad \textrm{at the}\ j\textrm{-th iteration} \\
& =\nabla_{\theta^j_\textbf{\textrm{GD}}}\sum_{i=1}^{|\mathbb{S}_\textrm{match}|}[\mathcal{L}^i_\textbf{\textrm{syn}}(\theta^j_\textbf{\textrm{GD}},\mathcal{X}^{\mathcal{T}})-\mathcal{L}^i_\textbf{\textrm{syn}}(\theta^j_\textbf{\textrm{EMA}},\mathcal{X}^{\mathcal{T}})]\\
& = \nabla_{\theta^j_\textbf{\textrm{GD}}}\sum_{i=1}^{|\mathbb{S}_\textrm{match}|}[\mathcal{L}^i_\textbf{\textrm{syn}}(\theta^j_\textbf{\textrm{GD}},\mathcal{X}^{\mathcal{T}})-\mathcal{L}^i_\textbf{\textrm{syn}}(\theta^j_\textbf{\textrm{GD}} + \sum_{k=1}^{j-1}\beta^{j-k}\ve{g}_k,\mathcal{X}^{\mathcal{T}})] \\
& = \nabla_{\theta^j_\textbf{\textrm{GD}}}\sum_{i=1}^{|\mathbb{S}_\textrm{match}|}[\mathcal{L}^i_\textbf{\textrm{syn}}(\theta^j_\textbf{\textrm{GD}},\mathcal{X}^{\mathcal{T}})-\mathcal{L}^i_\textbf{\textrm{syn}}(\theta^j_\textbf{\textrm{GD}} + \beta^{j-1}\ve{g}_1,\mathcal{X}^{\mathcal{T}}) + \cdots \\
& \quad+ \mathcal{L}^i_\textbf{\textrm{syn}}(\theta^j_\textbf{\textrm{GD}} + \sum_{k=1}^{j-2}\beta^{j-k}\ve{g}_k,\mathcal{X}^{\mathcal{T}})-\mathcal{L}^i_\textbf{\textrm{syn}}(\theta^j_\textbf{\textrm{GD}} + \sum_{k=1}^{j-1}\beta^{j-k}\ve{g}_k,\mathcal{X}^{\mathcal{T}})] \\
& \approx \nabla_{\theta^j_\textbf{\textrm{GD}}}\sum_{i=1}^{|\mathbb{S}_\textrm{match}|}[(\beta^{j-1}\rho)||\nabla_{\theta^j_\textbf{\textrm{GD}}}\mathcal{L}^i_\textbf{\textrm{syn}}(\theta^j_\textbf{\textrm{GD}},\mathcal{X}^{\mathcal{T}})||_2+\cdots \\
&\quad+(\beta^1\rho)||\nabla_{\theta^j_\textbf{\textrm{GD}} + \sum_{k=1}^{j-2}\beta^{j-k}\ve{g}_k}\mathcal{L}^i_\textbf{\textrm{syn}}(\theta^j_\textbf{\textrm{GD}} + \sum_{k=1}^{j-2}\beta^{j-k}\ve{g}_k,\mathcal{X}^{\mathcal{T}})||_2]\quad\quad\textcolor{C3}{\#\ \textrm{The solution of \textit{dual norm} problem}}\\
& \approx \nabla_{\theta^j_\textbf{\textrm{GD}}}\sum_{i=1}^{|\mathbb{S}_\textrm{match}|}[\sqrt{\mathbb{E}_{(\theta_1,\theta_2)\sim\textbf{\textrm{Unif}}(\theta^j_\textbf{\textrm{GD}},\theta^j_\textbf{\textrm{GD}}+\beta^{j-1}\ve{g}_1,\cdots,\theta^j_\textbf{\textrm{GD}} + \sum_{k=1}^{j-1}\beta^{j-k}\ve{g}_k)}||\nabla_{\theta_1}\mathcal{L}^i_\textbf{\textrm{syn}}(\theta_1,\mathcal{X}^{\mathcal{T}})||_2||\nabla_{\theta_2}\mathcal{L}^i_\textbf{\textrm{syn}}(\theta_2,\mathcal{X}^{\mathcal{T}})||_2}]. \\
\end{aligned}
\label{eq:proof_sam_11}
\end{equation}
Thus, we can further obtain a SAM-like presentation.
\begin{equation}
\small
\begin{aligned}
& \min_{\mathcal{X}^\mathcal{S}} \sum_{i=1}^{|\mathbb{S}_\textrm{match}|}[\mathcal{L}^i_\textbf{\textrm{syn}}(\theta^j_\textbf{\textrm{GD}},\theta^j_\textbf{\textrm{EMA}})],\quad\quad \textrm{at the}\ j\textrm{-th iteration}\\
 &= \min_{\mathcal{X}^\mathcal{S}} \sum_{i=1}^{|\mathbb{S}_\textrm{match}|}[\mathbb{E}_{(\theta_1,\theta_2)\sim\textbf{\textrm{Unif}}(\theta^j_\textbf{\textrm{GD}},\theta^j_\textbf{\textrm{GD}}+\beta^{j-1}\ve{g}_1,\cdots,\theta^j_\textbf{\textrm{GD}} + \sum_{k=1}^{j-1}\beta^{j-k}\ve{g}_k)}||\nabla_{\theta_1}\mathcal{L}^i_\textbf{\textrm{syn}}(\theta_1,\mathcal{X}^{\mathcal{T}})||_2||\nabla_{\theta_2}\mathcal{L}^i_\textbf{\textrm{syn}}(\theta_2,\mathcal{X}^{\mathcal{T}})||_2]\\
& = \min_{\mathcal{X}^\mathcal{S}} \sum_{i=1}^{|\mathbb{S}_\textrm{match}|}[\max_{\epsilon:||\epsilon||\leq \rho}  \mathbb{E}_{(\theta\sim \beta \theta^j_\textbf{\textrm{GD}} +(1-\beta)\theta^j_\textbf{\textrm{EMA}},\beta\sim\mathcal{U}[0,1])}\mathcal{L}^i_\textbf{\textrm{syn}}(\theta+\epsilon,\mathcal{X}^{\mathcal{T}})]. \\
\end{aligned}
\label{eq:proof_sam_12}
\end{equation}
Consequently, optimizing Eq.~\ref{eq:proof_sam_10} effectively addresses the SAM problem during the data synthesis phase, which results in a flat loss landscape. Additionally, Eq.~\ref{eq:proof_sam_12} presents a variant of the SAM algorithm that slightly differs from the traditional form. This variant is specifically designed to ensure sharpness-aware minimization within a $\rho$-ball for each point along a straight path between $\theta^j_\textbf{\textrm{GD}}$ and $\theta^j_\textbf{\textrm{EMA}}$.

\section{Visualization of Prior Dataset Condensation Methods}
\label{apd:visualization}
\begin{figure*}[h]
\centering
\includegraphics[width=1.\textwidth]{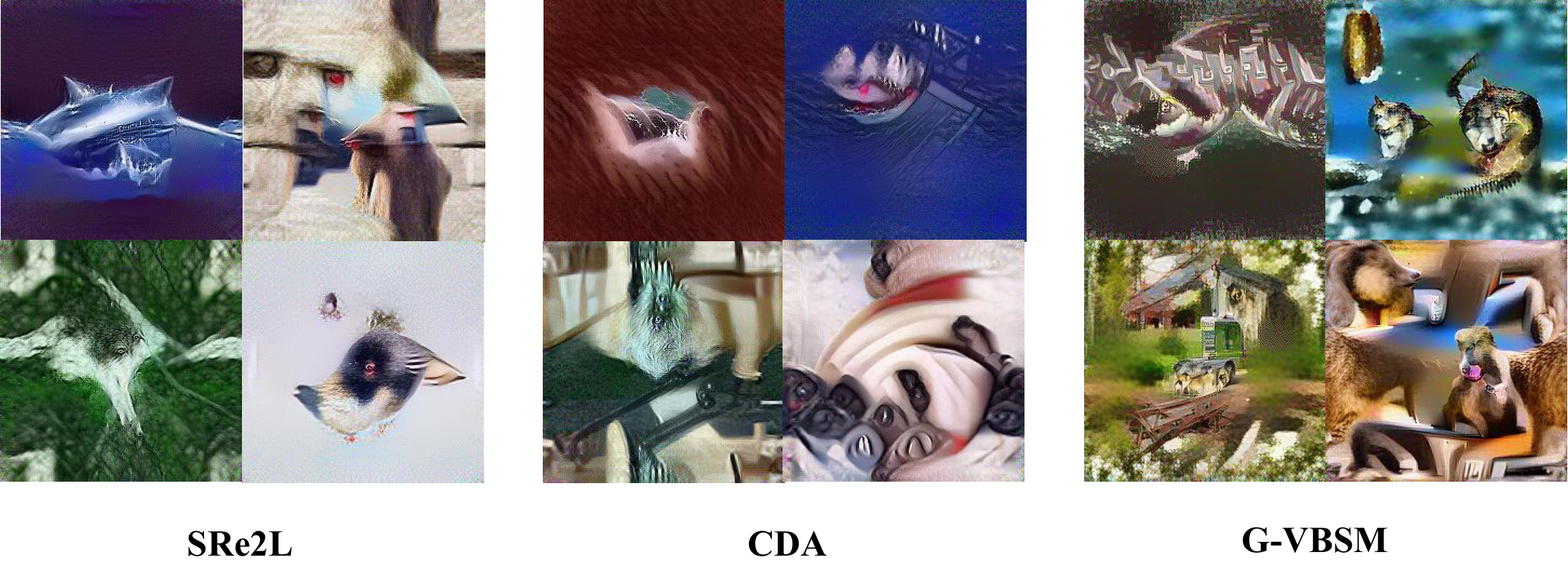}
\caption{Visualization of the synthetic images of prior training-dependent dataset condensation methods.}
\vspace{-1ex}
\label{figure:prior_vis}
\end{figure*}

In Fig.~\ref{figure:prior_vis}, we present the visualization results of previous training-dependent dataset condensation methods. These approaches, which optimize starting from Gaussian noise, tend to produce synthetic images that lack realism and fail to convey clear semantics to the naked eye.

\section{More Ablation Experiments}
\label{apd:add_ab_experiment}

In this section, we present a series of ablation studies to further validate the design choices outlined in the main paper.

\subsection{Backbone Choices of Data Synthesis on ImageNet-1k}
\begin{table}[!h]
\centering
\footnotesize
\renewcommand\arraystretch{0.95}
\vspace{-4pt}
\resizebox{1.0\textwidth}{!}{%
\begin{tabular}{cccccccc|cc}
\toprule
\multicolumn{8}{c|}{Observer Model} & \multicolumn{2}{c}{Verified Model}\\
ResNet-18 & MobileNet-V2 & EfficientNet-B0 & ShuffleNet-V2 & WRN-40-2 & AlexNet & ConvNext-Tiny & DenseNet-121 & ResNet-18 & ResNet-50 \\\midrule
\cmark & \cmark & \cmark & \cmark & & & & & 38.7 & 42.0 \\
\cmark & \cmark & \cmark & \cmark & \cmark & & & & 36.7 & 43.3 \\
\cmark & \cmark & \cmark & \cmark & & \cmark & & & \graycell 39.0 & \graycell 43.8 \\
\cmark & \cmark & \cmark & \cmark & \cmark & \cmark & & & 37.4 & 43.1 \\
\cmark & \cmark & \cmark & \cmark & \cmark & \cmark &  \cmark & \cmark & 34.8 & 40.6 \\
\bottomrule
\end{tabular}}
\vspace{2pt}
\caption{\textbf{Ablation studies on ImageNet-1k with IPC 10.} Verify the influence of backbone choices on data synthesis with \config{C} ($\zeta=1.5$).}
\label{tab:bc_data_synthesis}
\vspace{-4pt}
\end{table}

The results in Table~\ref{tab:bc_data_synthesis} demonstrate the significant impact of backbone architecture selection on the performance of dataset distillation. This study employs the optimal configuration, which includes ResNet-18, MobileNet-V2, EfficientNet-B0, ShuffleNet-V2, and AlexNet.

\subsection{Backbone Choices of Soft Label Generation on ImageNet-1k}
\begin{table}[!h]
\centering
\footnotesize
\renewcommand\arraystretch{0.95}
\vspace{-4pt}
\resizebox{1.0\textwidth}{!}{%
\begin{tabular}{ccccc|c|ccc}
\toprule
\multicolumn{5}{c|}{Observer Model} &\multicolumn{1}{c|}{\multirow{2}{*}{Cost Time (s)}} & \multicolumn{3}{c}{Verified Model}\\
ResNet-18 & MobileNet-V2 & EfficientNet-B0 & ShuffleNet-V2 & AlexNet &  & ResNet-18 & ResNet-50 & ResNet-101 \\\midrule
\cmark & \cmark & \cmark & \cmark & & 598 & 9.1 & 9.5 & 6.2 \\
\cmark & \cmark &  & \cmark & & \graycell 519 & 9.4 & 8.4 & 6.5 \\
\cmark & \cmark &  & \cmark & \cmark & 542 & \graycell 12.8 & \graycell 13.3 & \graycell 8.4 \\
\bottomrule
\end{tabular}}
\vspace{2pt}
\caption{\textbf{Ablation studies on ImageNet-1k with IPC 1.} Verify the influence of backbone choice on soft label generation with \config{G} ($\zeta=2$).}
\label{tab:bc_soft_label_generation}
\vspace{-4pt}
\end{table}

Our strategy better backbone choice, which focuses on utilizing lighter backbone combinations for soft label generation, significantly enhances the generalization capabilities of the condensed dataset. Empirical studies conducted with IPC 1, and the results detailed in Table~\ref{tab:bc_soft_label_generation}, show that optimal performance is achieved by using ResNet-18, MobileNet-V2, EfficientNet-B0, ShuffleNet-V2, and AlexNet for data synthesis. For soft label generation, the combination of ResNet-18, MobileNet-V2, ShuffleNet-V2, and AlexNet demonstrates most effective.
\begin{figure}[t]
\centering
\includegraphics[height=0.5\textwidth,trim={0cm 0cm 0cm 0cm},clip]{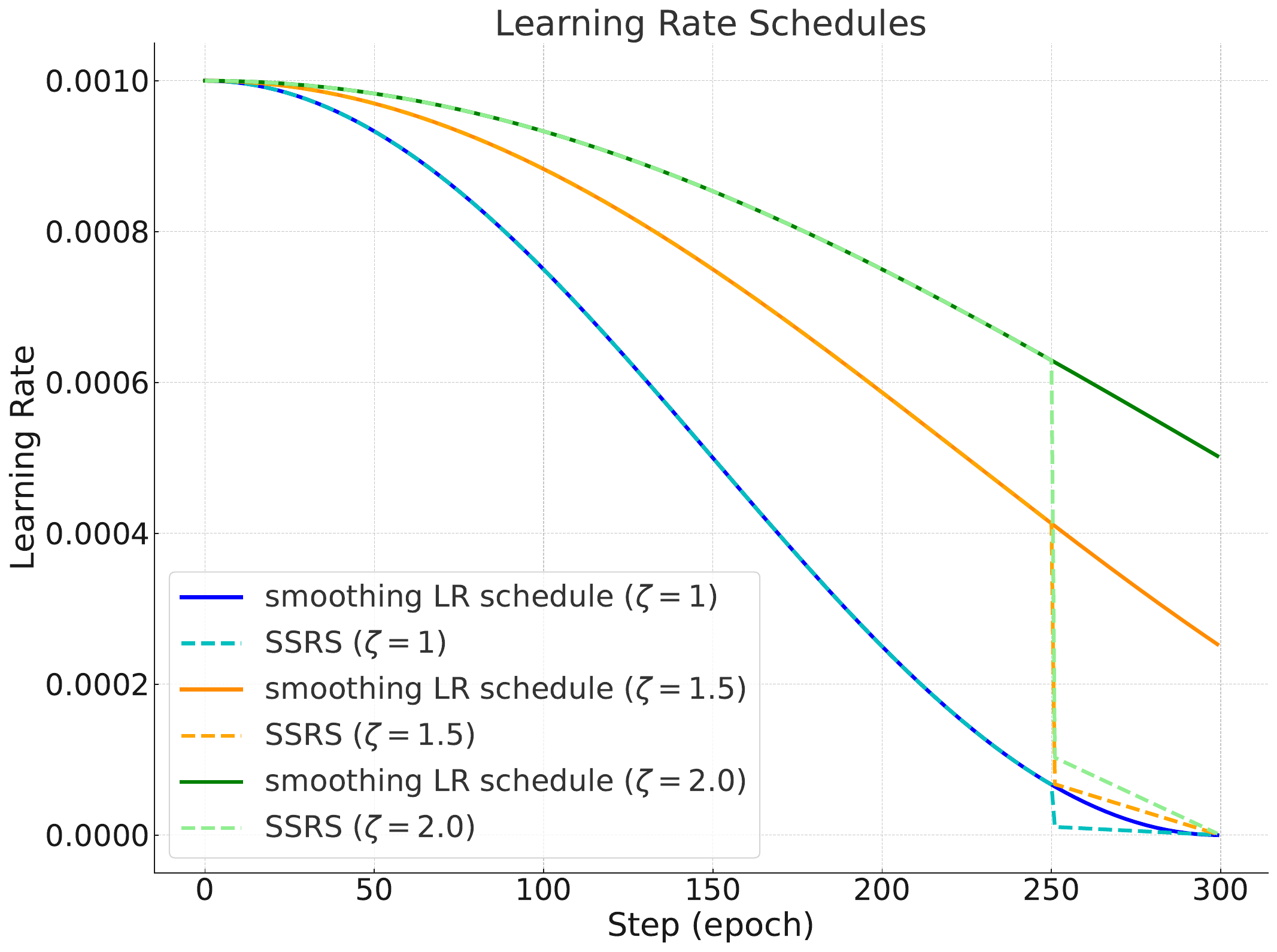}
\vspace{-5pt}
\caption{The visualization of SSRS and smoothing LR schedule.}
\label{fig:SSRS}
\end{figure}
\subsection{Smoothing LR Schedule Analysis}
\begin{table}[!h]
\centering
\footnotesize
\renewcommand\arraystretch{0.95}
\vspace{-4pt}
\resizebox{0.4\textwidth}{!}{%
\begin{tabular}{c|ccccc}
\toprule
\multirow{2}{*}{Config} & \multicolumn{5}{c}{Slowdown Coefficient $\zeta$} \\
& 1.0 & 1.5 & 2.0 & 2.5 & 3.0 \\\midrule
\config{C} & 24.5 & 28.2 & 30.6 & \graycell 32.4 & 31.8 \\
\bottomrule
\end{tabular}}
\vspace{2pt}
\caption{\textbf{Ablation studies on ImageNet-1k with IPC 10.} Additional experimental result of the slowdown coefficient $\zeta$ on the verified model MobileNet-V2.}
\label{tab:add_mobilenet}
\vspace{-4pt}
\end{table}

\begin{table}[!h]
\centering
\footnotesize
\renewcommand\arraystretch{0.95}
\vspace{-4pt}
\resizebox{0.5\textwidth}{!}{%
\begin{tabular}{cc|ccc}
\toprule
\multirow{2}{*}{Config} & \multirow{2}{*}{$\gamma$} & \multicolumn{3}{c}{Verified Model} \\
& & ResNet-18 & ResNet-50 & ResNet-101 \\\midrule
\config{F} & 0.997 & 47.6& 53.5 & \graycell 52.0 \\
\config{F} & 0.9975 & 47.4 & \graycell 54.0 & 50.9\\
\config{F} & 0.99775 & 47.3 & 53.7 & 50.3\\
\config{F} & 0.997875 &\graycell 47.8 & 53.8 & 50.7\\
\bottomrule
\end{tabular}}
\vspace{2pt}
\caption{\textbf{Ablation studies on ImageNet-1k with IPC 10.} Verify the effectiveness of ALRS in post-evaluation.}
\label{tab:alrs}
\vspace{-4pt}
\end{table}

Due to space limitations in the main paper, the experimental results for MobileNet-V2, which are not included in Table~\ref{tab:imagenet_scheduler} Left, are presented in Table~\ref{tab:add_mobilenet}. Additionally, we investigate \textit{Adaptive Learning Rate Scheduler} (ALRS), an algorithm that adjusts the learning rate based on training loss. Although ALRS did not produce effective results, it provides valuable insights for future research. This scheduler was first introduced in~\citep{eccvw_sam} and is described as follows:

$$\mu(i) = \mu(i-1)\gamma^{\mathbbm{1}\left[\frac{|L_i-L_{i-1}|}{|L_i|}\leq h_1\text{ and }|L_i - L_{i-1}|\leq h_2\right]},$$

Here, $\gamma$ represents the decay rate, $L_i$ is the training loss at the $i$-th iteration, and $h_1$ and $h_2$ are the first and second thresholds, respectively, both set by default to 0.02. We list several values of $\gamma$ that demonstrate the best empirical performance in Table~\ref{tab:alrs}. These results allow us to conclude that our proposed smoothing LR schedule outperforms ALRS in the dataset condensation task.

Ultimately, we introduce a learning rate scheduler superior to the traditional smoothing LR schedule in scenarios with high IPC. This enhanced strategy, named \textit{early Smoothing-later Steep Learning Rate Schedule} (SSRS), integrates the smoothing LR schedule with MultiStepLR. It intentionally implements a significant reduction in the learning rate during the final epochs of training to accelerate model convergence. The formal definition of SSRS is as follows:
\begin{equation}
\mu(i) =\begin{cases}
	\frac{1+\textrm{cos}(i\pi/\zeta N)}{2} &,i\leq \frac{5N}{6},\\\vspace{2pt}
\frac{1+\textrm{cos}(5\pi/\zeta 6)}{2}\frac{(6N-6i)}{6N} &,i> \frac{5N}{6}.
	\end{cases}\\
\label{eq:SSRS}
\end{equation}

\begin{table}[!h]
\centering
\footnotesize
\renewcommand\arraystretch{0.95}
\vspace{-4pt}
\resizebox{0.7\textwidth}{!}{%
\begin{tabular}{cc|cccc}
\toprule
\multirow{2}{*}{Config} & \multirow{2}{*}{Scheduler Type} & \multicolumn{4}{c}{Verified Model} \\
& & ResNet-18 & ResNet-50 & ResNet-101 & MobileNet-V2\\\midrule
\config{G} & \makecell{smoothing LR schedule} & 56.4 & 62.2 & 62.3 & 54.7 \\
\config{G} & SSRS & \graycell 57.4 & \graycell 63.0 & \graycell 63.6 & \graycell 56.5 \\
\bottomrule
\end{tabular}}
\vspace{2pt}
\caption{\textbf{Ablation studies on ImageNet-1k with IPC 40.} Verify the effectiveness of SSRS in post-evaluation.}
\label{tab:ssrs}
\vspace{-4pt}
\end{table}

Note that the visualization of SSRS can be found in Fig.~\ref{fig:SSRS}. Meanwhile, the comparative experimental results of SSRS and the smoothing LR schedule are detailed in Table~\ref{tab:ssrs}. Notably, SSRS enhances the verified model's performance without incurring additional overhead.

\subsection{Understanding of EMA-based Evaluation}
\begin{table}[!h]
\centering
\footnotesize
\renewcommand\arraystretch{0.95}
\vspace{-4pt}
\resizebox{0.5\textwidth}{!}{%
\begin{tabular}{c|ccccc}
\toprule
\multirow{2}{*}{\config{F}} & \multirow{1}{*}{EMA Rate} & 0.99 & 0.999 & 0.9999 & 0.999945 \\\cmidrule(lr){2-6}
 & Accuracy& \graycell 48.2 & 48.1 & 22.1 & 0.45 \\
\bottomrule
\end{tabular}}
\vspace{2pt}
\caption{\textbf{Ablation studies on ImageNet-1k with IPC 10.} Verify the effect of EMA Rate in EMA-based Evaluation.}
\label{tab:ema_rate}
\vspace{-4pt}
\end{table}

The EMA Rate, a crucial hyperparameter governing the EMA update rate during post-evaluation, significantly influences the final results. Additional experimental outcomes, presented in Table~\ref{tab:ema_rate}, reveal that the EMA Rate 0.99 we adopt in the main paper yields optimal performance.

\subsection{Ablation Studies on CIFAR-10}
This section details the process of deriving hyperparameter configurations for CIFAR-10 through exploratory studies. The demonstrated superiority of our EDC method over traditional approaches, as detailed in our main paper, suggests that conventional dataset condensation techniques like MTT~\citep{dd_mtt} and KIP~\citep{dd_kip} are not the sole options for achieving superior performance on small-scale datasets.
\begin{table}[!h]
\centering
\footnotesize
\renewcommand\arraystretch{0.95}
\vspace{-4pt}
\resizebox{0.5\textwidth}{!}{%
\begin{tabular}{c|cccccc}
\toprule
Iteration & 25 & 50 & 75 & 100 & 125 & 1000 \\\hline
Accuracy & 42.1 & 42.4 &  \graycell 42.7 & 42.5 & 42.3 & 41.8 \\
\bottomrule
\end{tabular}}
\vspace{2pt}
\caption{\textbf{Ablation studies on CIFAR-10 with IPC 10.} We employ ResNet-18 exclusively for data synthesis and soft label generation, examining the impact of iteration count during post-evaluation and adhering to RDED's consistent hyperparameter settings.}
\label{tab:c10_iteration}
\vspace{-4pt}
\end{table}

\begin{table}[!h]
\centering
\footnotesize
\renewcommand\arraystretch{0.95}
\vspace{-4pt}
\resizebox{0.9\textwidth}{!}{%
\begin{tabular}{cccc|cccc}
\toprule
\multicolumn{2}{c}{Data Synthesis} & \multicolumn{2}{c|}{Soft Label Generation} & \multicolumn{4}{c}{Verified Model} \\
w/ pre-train & w/o pre-train & w/ pre-train & w/o pre-train & ResNet-18 & ResNet-50 & ResNet-101 & MobileNet-V2  \\\midrule
\xmark & \cmark & \xmark & \cmark & \graycell 77.7 & \graycell 73.0 &\graycell  68.2 & 38.2 \\
\xmark & \cmark & \cmark & \xmark & 60.5 & 56.3 & 52.2 & \graycell 39.9 \\
\cmark & \xmark & \cmark & \xmark & 60.0 & 56.1 & 50.7 & 39.0 \\
\cmark & \xmark & \xmark & \cmark & 74.9 & 70.9 & 61.4 & 38.2 \\
\bottomrule
\end{tabular}}
\vspace{2pt}
\caption{\textbf{Ablation studies on CIFAR-10 with IPC 10.} Hyperparameter settings follow those in Table~\ref{tab:setting_cifar_10}, excluding the scheduler and batch size, which are set to smoothing LR schedule ($\zeta=2$) and 50, respectively.}
\label{tab:c10_backbone}
\vspace{-4pt}
\end{table}

\begin{table}[!h]
\centering
\footnotesize
\renewcommand\arraystretch{0.95}
\vspace{-4pt}
\resizebox{0.7\textwidth}{!}{%
\begin{tabular}{lc|cccc}
\toprule
\multirow{2}{*}{EMA Rate} & \multirow{2}{*}{Batch Size} & \multicolumn{4}{c}{Verified Model} \\
& & ResNet-18 & ResNet-50 & ResNet-101 & MobileNet-V2  \\\hline
0.99 & 50 & 77.7 & 73.0 & 68.2 & 38.2 \\
0.999 & 50 & 13.1 & 11.8 & 11.6 & 11.2 \\
0.9999 & 50 & 10.0 & 10.0 & 10.0 & 10.0 \\
0.99 & 25 & \graycell 78.1 & \graycell 76.0  & \graycell 71.8 & \graycell 42.1 \\
0.99 & 10 & 76.0 & 70.0  & 57.7 & 42.9 \\
\bottomrule
\end{tabular}}
\vspace{2pt}
\caption{\textbf{Ablation studies on CIFAR-10 with IPC 10.} Explore the influence of EMA Rate and batch size in post-evaluation. Hyperparameter settings follow those in Table~\ref{tab:setting_cifar_10}, excluding the scheduler, which are set to smoothing LR schedule ($\zeta=2$).}
\label{tab:c10_ema_rate_batch_size}
\vspace{-4pt}
\end{table}

\begin{table}[!h]
\centering
\footnotesize
\renewcommand\arraystretch{0.95}
\vspace{-4pt}
\resizebox{1.0\textwidth}{!}{%
\begin{tabular}{lc|cccc}
\toprule
\multirow{2}{*}{Scheduler} & \multirow{2}{*}{Option} & \multicolumn{4}{c}{Verified Model} \\
& & ResNet-18 & ResNet-50 & ResNet-101 & MobileNet-V2  \\\midrule
Smoothing LR Schedule & $\zeta=2$ & 78.1 & \graycell 76.0 & \graycell 71.8 & \graycell 42.4 \\
Smoothing LR Schedule & $\zeta=3$ & 77.3 & 75.0 & 68.5 & 41.1 \\
MultiStepLR & \makecell{$\gamma=0.5$, milestones=[800,900,950]} & \graycell 79.1 & \graycell 76.0 & 67.1 & 42.0 \\
MultiStepLR & \makecell{$\gamma=0.25$, milestones=[800,900,950]} & 77.7 & 75.8 & 67.0 & 40.3 \\
\bottomrule
\end{tabular}}
\vspace{2pt}
\caption{\textbf{Ablation studies on CIFAR-10 with IPC 10.} Explore the influence of various scheduler in post-evaluation. Hyperparameter settings follow those in Table~\ref{tab:setting_cifar_10}.}
\label{tab:c10_schedule}
\vspace{-4pt}
\end{table}

Our quantitative experiments, detailed in Table~\ref{tab:c10_iteration}, pinpoint 75 iterations as the empirically optimal count. This finding highlights that, for smaller datasets with limited samples and fewer categories, fewer iterations are required to achieve superior results.

Subsequently, we evaluate the effectiveness of using a pre-trained model on ImageNet-1k for dataset condensation on CIFAR-10. Our study differentiates two training pipelines: the first involves 100 epochs of pre-training followed by 10 epochs of fine-tuning (denoted as `w/ pre-train'), and the second comprises training from scratch for 10 epochs (denoted as `w/o pre-train'). The results, presented in Table~\ref{tab:c10_backbone}, indicate that pre-training on ImageNet-1k does not significantly enhance dataset distillation performance.

We further explore how batch size and EMA Rate affect the generalization abilities of the condensed dataset. Results in Table~\ref{tab:c10_ema_rate_batch_size} show that a reduced batch size of 25 enhances performance on CIFAR-10.

In our final set of experiments, we compare MultiStepLR and smoothing LR schedules. As detailed in Table~\ref{tab:c10_schedule}, MultiStepLR is superior for ResNet-18 and ResNet-50, whereas the smoothing LR schedule is more effective for ResNet-101 and MobileNet-V2.

\section{Synthesized Image Visualization}
\label{apd:syn_image_vis}

\begin{figure}[!t]
\centering
\includegraphics[width=\textwidth]{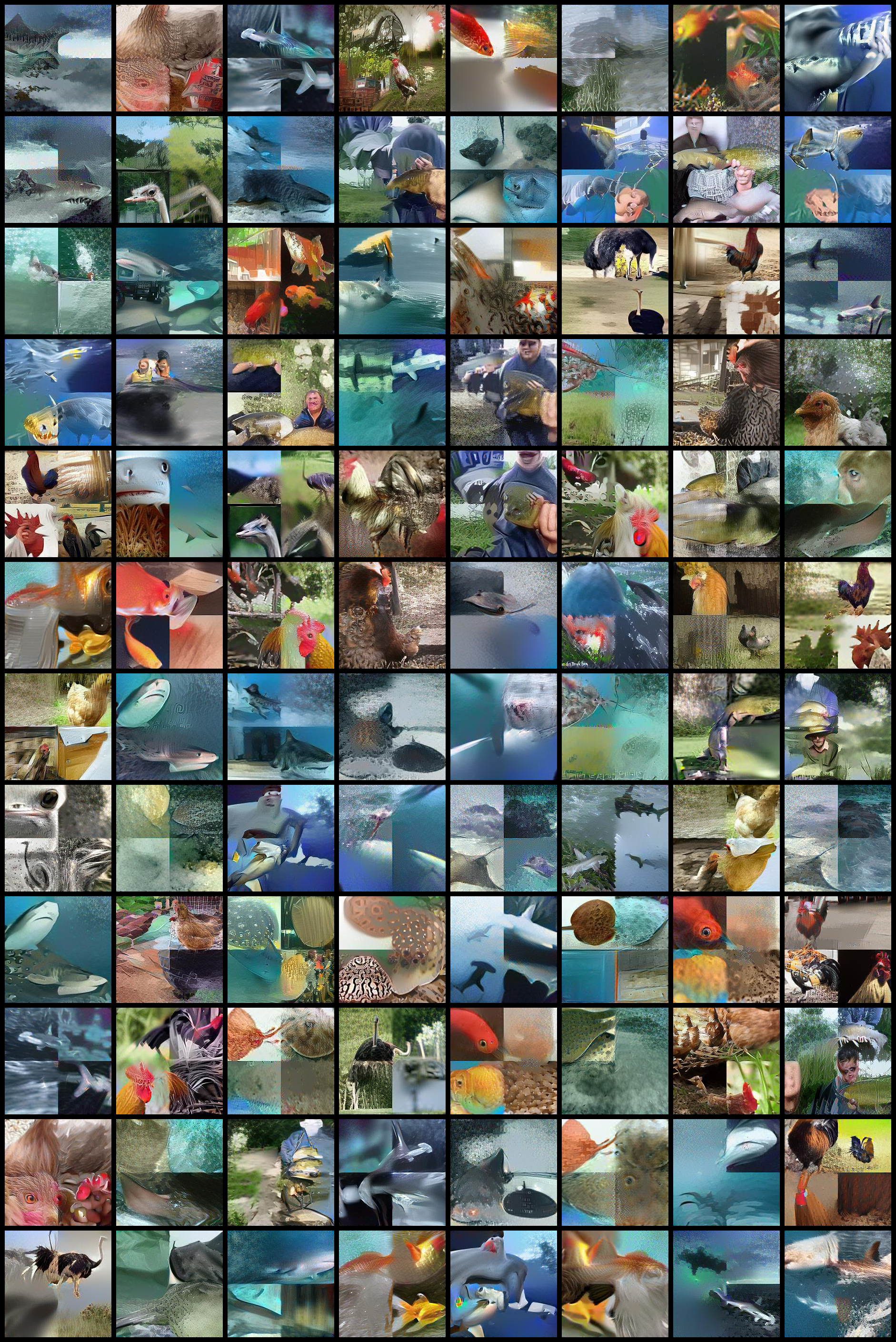}
        \caption{Synthetic data visualization on ImageNet-1k randomly selected from EDC.}
        \label{fig:vis_imagenet_1k}
\end{figure}

\begin{figure}[!t]
\centering
\includegraphics[width=\textwidth]{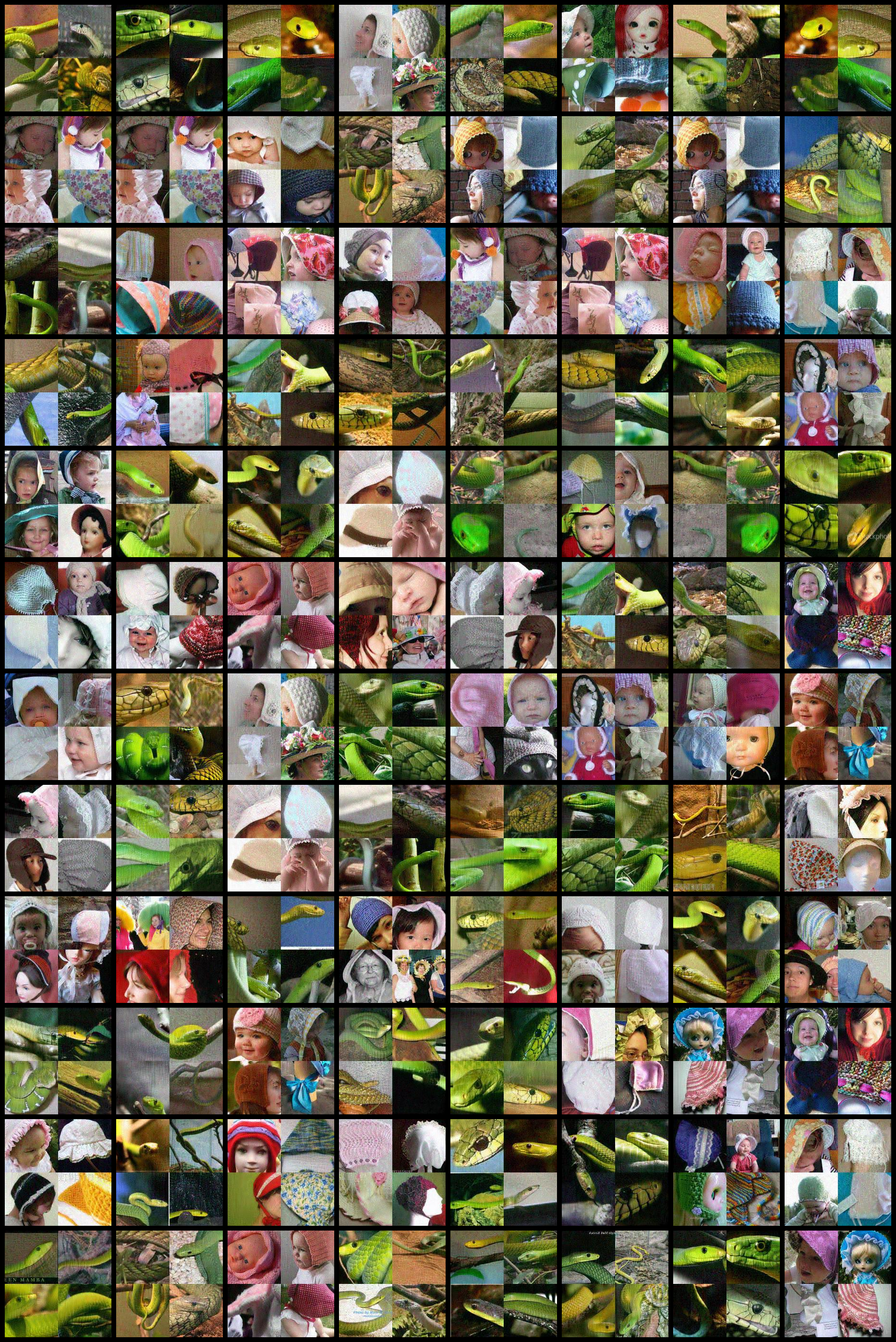}
        \caption{Synthetic data visualization on ImageNet-10 randomly selected from EDC.}
        \label{fig:vis_imagenet_10}
\end{figure}

\begin{figure}[!t]
\centering
\includegraphics[width=0.92\textwidth]{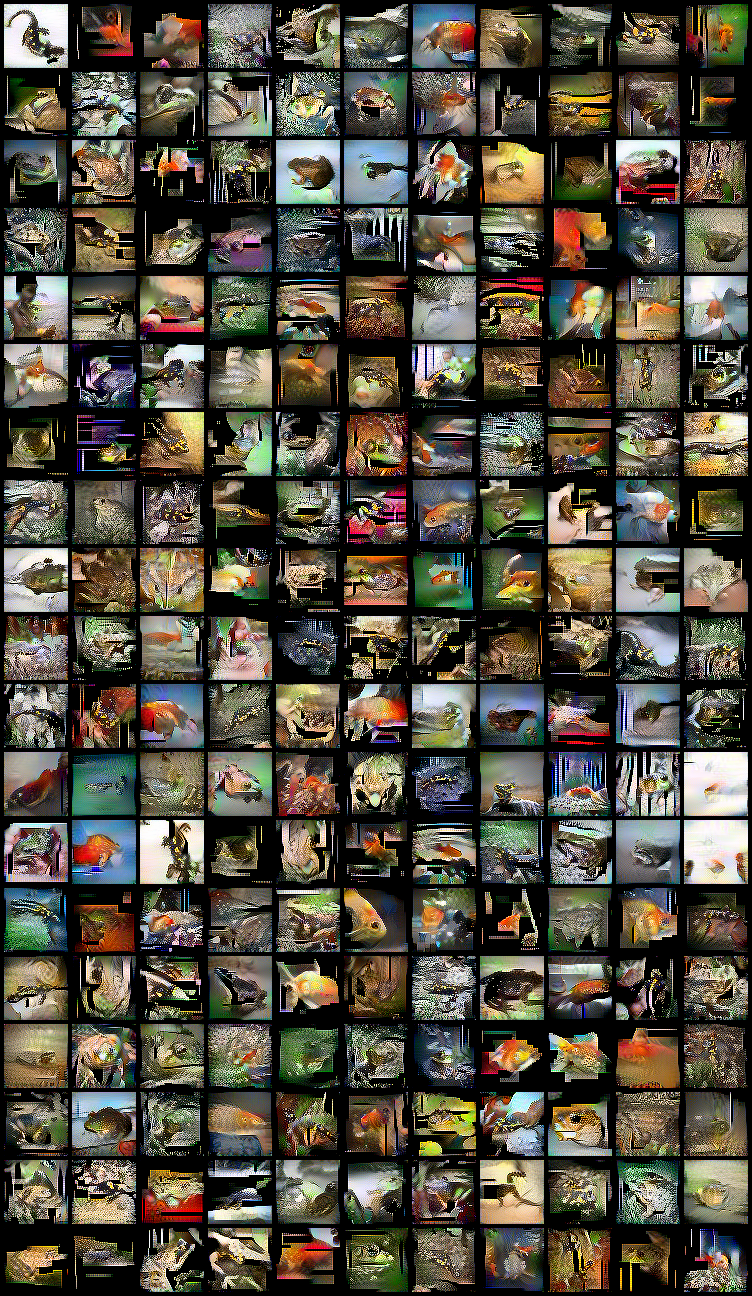}
        \caption{Synthetic data visualization on Tiny-ImageNet randomly selected from EDC.}
        \label{fig:vis_tiny_imagenet}
\end{figure}

\begin{figure}[!t]
\centering
\includegraphics[width=\textwidth]{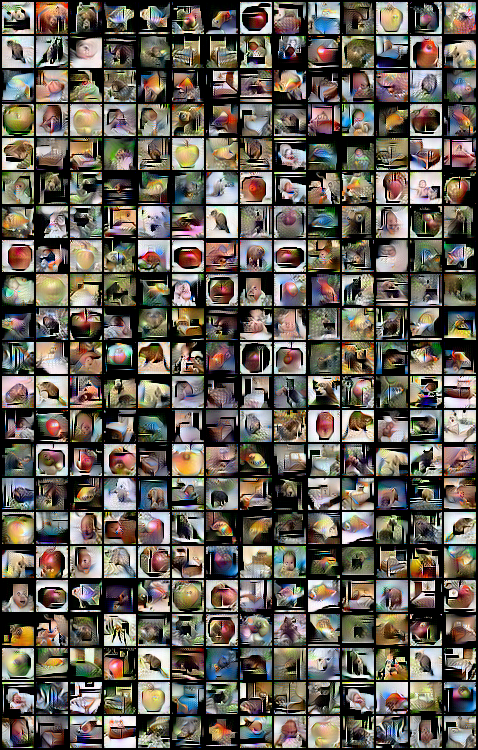}
        \caption{Synthetic data visualization on CIFAR-100 randomly selected from EDC.}
        \label{fig:vis_cifar_100}
\end{figure}

\begin{figure}[!t]
\centering
\includegraphics[width=\textwidth]{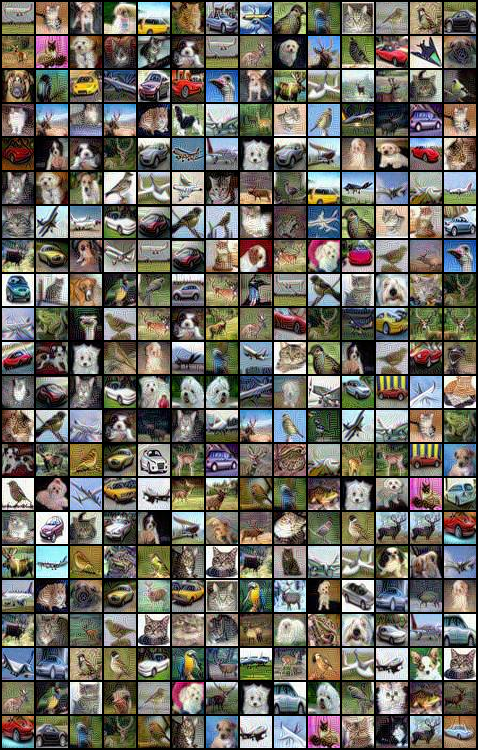}
        \caption{Synthetic data visualization on CIFAR-10 randomly selected from EDC.}
        \label{fig:vis_cifar_10}
\end{figure}

The visualization of the condensed dataset is showcased across Figs.~\ref{fig:vis_imagenet_1k} to \ref{fig:vis_cifar_10}. Specifically, Figs.~\ref{fig:vis_imagenet_1k}, \ref{fig:vis_tiny_imagenet}, \ref{fig:vis_cifar_100}, and \ref{fig:vis_cifar_10} present the datasets synthesized from ImageNet-1k, Tiny-ImageNet, CIFAR-100, and CIFAR-10, respectively.

\section{Ethics Statement} \label{ethics_statement}

Our research utilizes synthetic data to avoid the use of actual personal information, thereby addressing privacy and consent issues inherent in datasets with identifiable data. We generate synthetic data using a methodology that distills from real-world data but maintains no direct connection to individual identities. This method aligns with data protection laws and minimizes ethical risks related to confidentiality and data misuse. However, it is important to note that models trained on synthetic data may not achieve the same accuracy levels as those trained on the full original dataset.

\section{Limitations} \label{appendix_limitations}

The paper offers an extensive examination of the design space for dataset condensation, but it might still miss some potentially valuable strategies due to the broad scope. Additionally, as the IPC count grows, the performance of the described approach converges with that of the \textit{baseline} RDED.

\clearpage

\clearpage

\section{Additional Experiments, Theories and Descriptions (Rebuttal Stage Supplement)}

Here we add some experiments, theories and explanations that we think it is necessary to add.

\subsection{Scalability on ImageNet-21k}

\begin{table}[!h]
\centering
\footnotesize
\renewcommand\arraystretch{0.95}
\vspace{-4pt}
\begin{tabular}{ccccc}
\toprule
SRe$^2$L & CDA & RDED & EDC & Original Dataset \\ \midrule
18.5 & 22.6 & 25.6 & \graycell 26.8 & 38.5 \\
\bottomrule
\end{tabular}
\vspace{2pt}
\caption{\textbf{Comparison of Different Methods on ImageNet-21k.}}
\label{tab:comparison_methods_imagenet_21k}
\vspace{-4pt}
\end{table}

We conduct experiments on a larger scale dataset ImageNet-21k-P with IPC 10. The results in Table~\ref{tab:comparison_methods_imagenet_21k} indicate that our method outperforms the state-of-the-art method CDA~\citep{yin2024dataset} on this dataset, demonstrating that EDC can scale to larger datasets.

\subsection{Complexity of Implementation}

\begin{table}[!h]
\centering
\footnotesize
\renewcommand\arraystretch{0.95}
\vspace{-4pt}
\begin{tabular}{cccc}
\toprule
Configuration & GPU Memory (G/per GPU) & Time Spent (hours) & Top-1 Accuracy (\%) \\ \midrule
CONFIG A & 4.616 & 9.77 & 31.4 \\
CONFIG B & 4.616 & \graycell 4.89 & 34.4 \\
CONFIG C & 4.616 & \graycell 4.89 & 38.7 \\
CONFIG D & 4.616 & 4.91 & 39.5 \\
CONFIG E & 4.697 & 4.91 & 46.2 \\
CONFIG F & 4.923 & 5.11 & 48.0 \\
CONFIG G & 4.923 & 5.11 & \graycell 48.6 \\
\bottomrule
\end{tabular}
\vspace{2pt}
\caption{\textbf{Comparison of computational resources on 4 RTX 4090.}}
\label{tab:resource_comparison}
\vspace{-4pt}
\end{table}

Here we present Table~\ref{tab:resource_comparison} to complement the computational overhead in Fig.~\ref{figure:illustration} in the main paper. EDC is an efficient algorithm as it reduces the number of iterations by half, compared to the \textit{baseline} G-VBSM. As illustrated in the table above, although transitioning from CONFIG A to CONFIG G adds small GPU memory overhead, it is minor compared to the reduction in time spent. Additionally, introducing EDC to other tasks often requires significant effort for tuning hyper-parameters or even redesigning statistical matching, which is a challenge EDC should address.

\subsection{Robustness Evaluation}

\begin{table}[!h]
\centering
\footnotesize
\renewcommand\arraystretch{0.95}
\vspace{-4pt}
\begin{tabular}{lccc}
\toprule
Attack Methods & MTT & SRe2L & EDC (Ours) \\ \midrule
Clean Accuracy & 26.16 & 43.24 & \graycell 57.21 \\
FGSM & 1.82 & 5.73 & \graycell 12.39 \\
PGD & 0.41 & 2.70 & \graycell 10.71 \\
CW & 0.36 & 2.94 & \graycell 5.27 \\
VMI & 0.42 & 2.60 & \graycell 10.73 \\
Jitter & 0.40 & 2.72 & \graycell 10.64 \\
AutoAttack & 0.26 & 1.73 & \graycell 7.94 \\
\bottomrule
\end{tabular}
\vspace{2pt}
\caption{\textbf{Comparison on DD-RobustBench.}}
\label{tab:attack_methods}
\vspace{-4pt}
\end{table}

We follow the pipeline in~\cite{wu2024dd} to evaluate the robustness of models trained on condensed datasets, utilizing the well-known adversarial attack library available at~\cite{kim2020torchattacks}. As illustrared in Table~\ref{tab:attack_methods}. Our experiments are conducted on Tiny-ImageNet with IPC 50, with the test accuracy presented in the table above. Evidently, EDC demonstrates significantly higher robustness compared to other methods. We attribute this to improvements in post-evaluation techniques, such as EMA-based evaluation and smoothing LR schedule, which help reduce the sharpness of the loss landscape.

\subsection{Theoretical Explanation of Irrational Hyperparameter Setting (Sketch!!)}

The smoothing LR schedule is designed to address suboptimal solutions that arise due to the scarcity of sample sizes in condensed datasets. Additionally, the use of small batch size is implemented because the gradient of the condensed dataset more closely resembles the global gradient of the original dataset, as illustrated at the bottom of Fig.~\ref{fig:empirical_analysis}. Against the latter, we can propose a complete chain of theoretical derivation:

\begin{equation}
    \begin{split}
\mathcal{L}_{syn} & = \mathbb{E}_{c_i\sim C} \Vert p_\theta(\mu | X^S, c_i) - p(\mu | X^T, c_i) ||_2 \\
&+ \Vert p_\theta(\sigma^2 | X^S, c_i) - p(\theta^2 | X^T, c_i) \Vert_2 \text{\quad\# (Our statistical matching)} \\
\partial L_{syn} / \partial \theta & = \int_{c_i} ( \partial L_{syn} / \partial p_\theta(\cdot| X^S, c_i) ) ( \partial p_\theta(\cdot| X^S, c_i) / \partial \theta ) d_{c_i} \\
& \approx \int_{c_i} ( [ p_\theta(\mu | X^S, c_i) - p(\mu | X^T, c_i) ] + [ p_\theta(\sigma^2 | X^S, c_i) - p(\sigma^2 | X^T, c_i) ] ) ( \partial p_\theta(\cdot | X^S, c_i) / \partial \theta ) d_{c_i} \\
\end{split}
\end{equation}
where $p_\theta(· | X^S, c_i)$ and $p(· | X^T, c_i)$ refer to a Gaussian component in the Gaussian Mixture Model. Consider post-evaluation, We can derive the gradient of the MSE loss as:

\begin{equation}
    \begin{split}
        & \partial \mathbb{E}_{x_i\sim X^S} \Vert f_\theta(x_i) - y_i \Vert_2^2 / \partial\theta = 2\mathbb{E}_{x_i\sim X^S} [ ( f_\theta(x_i) - y_i ) ( \partial f_\theta(x_i) / \partial\theta ) ] \\
        & = 2\mathbb{E}_{x_i\sim X^S} [ ( f_\theta(x_i) - y_i ) \int_{c_i} ( \partial f_\theta(x_i) / \partial p_\theta(\cdot| X^S, c_i) ) ( \partial p_\theta(\cdot| X^S, c_i) / \partial\theta ) d_{c_i} ] \\
        & \approx 2\mathbb{E}_{(x_j, x_i) \sim  (X^S, X^T)} [ ( f_\theta(x_j) - y_j ) \int_{c_i} ( \partial f_\theta(x_i) / \partial p_\theta(\cdot| X^T, c_i) ) ( \partial p_\theta(\cdot| X^T, c_i) / \partial\theta ) d_{c_i} ] \\
        & \approx \partial \mathbb{E}_{x_i\sim X^T} || f_\theta(x_i) - y_i ||_2^2 / \partial\theta, \\
    \end{split}
\end{equation}
where $\theta$ stands for the model parameter. The right part of the penultimate row results from the loss $\mathcal{L}_\textrm{syn}$, which ensures the consistency of $p(\cdot| X^T, c_i)$ and $p(\cdot| X^S, c_i)$. If the model initialization during training is the same, the left part of the penultimate row is a scalar and has little influence on the direction of the gradient. Since $X^T$ is the complete original dataset with a global gradient, the gradient of $X^S$ approximates the global gradient of $X^T$, thus enabling the use of small batch size.

\subsection{Additional Related Work}

We additionally discuss the differences between published related papers~\citep{dd_datadam,add_related_work2,add_related_work3} and our work.

\paragraph{DataDAM~\citep{dd_datadam} vs. EDC.} Both DataDAM and EDC do not require model parameter updates during training. However, DataDAM struggles to generalize effectively to ImageNet-1k because it relies on randomly initialized models for distribution matching. As noted in SRe$^2$L, models trained for fewer than 50 epochs can experience significant performance degradation. DataDAM does not explore the soft label generation and post-evaluation phases as EDC does, limiting its competitiveness.

\paragraph{DANCE~\citep{add_related_work1} vs. EDC.} DANCE is a DM-based algorithm that, unlike traditional distribution matching, does not require model updates during data synthesis. Instead, it interpolates between pre-trained and randomly initialized models, using this interpolated model for distribution matching. Similarly, EDC also does not need to update the model parameters, but it uses a pre-trained model with a different architecture and does not incorporate random interpolation. The ``random interpolation'' technique was not adopted because it did not yield performance gains on ImageNet-1k. Although DANCE considers both intra-class and inter-class perspectives, it limits inter-class analysis to the logit level and intra-class analysis to the feature map level. In contrast, EDC performs both intra-class and inter-class matching at the feature map level, where inter-class matching is crucial. To support this, last year, SRe$^2$L focused solely on inter-class matching at the feature map level and still achieved state-of-the-art performance on ImageNet-1k. EDC is the first dataset distillation algorithm to simultaneously improve data synthesis, soft label generation, and post-evaluation stages. In contrast, DANCE only addresses the data synthesis stage. While DANCE can be effectively applied to ImageNet-1k, the introduction of soft label generation and post-evaluation improvements is essential for DANCE to achieve more competitive results.

\paragraph{M3D~\citep{add_related_work2} vs. EDC.} M3D is a DM-based algorithm, but its data synthesis paradigm aligns with DataDAM by relying solely on randomly initialized models, which limits its generalization to ImageNet-1k. M3D, similar to SRe$^2$L, G-VBSM, and EDC, takes into account second-order information (variance), but this is not a unique contribution of EDC. The key contributions of EDC in data synthesis are real image initialization, flatness regularization, and the consideration of both intra-class and inter-class matching.

\paragraph{Deng et al.~\citep{add_related_work3} vs. EDC.} Deng et al.~\citep{add_related_work3} is a DM-based algorithm, but its data synthesis paradigm is consistent with M3D and DataDAM, as it considers only randomly initialized models, which cannot be generalized to ImageNet-1k.
Deng et al.~\citep{add_related_work3} considers both interclass and intraclass information, similar to EDC. However, while EDC obtains interclass information by traversing the entire training set, Deng et al.~\citep{add_related_work3} derives interclass information from only one batch, making its information richness inferior to that of EDC. Deng et al.~\citep{add_related_work3} only explores data synthesis and does not explore soft label generation or post-evaluation. Additionally, Deng et al.~\citep{add_related_work3} only shares some similarity with Soft Category-Aware Matching among the 10 design choices in EDC.

\subsection{Implementation of Cropping}

The implementation of this crop operation refers to $\texttt{torchvision.transforms.RandomResizedCrop}$, where the minimum area threshold is controlled by the parameter scale[0]. The default value is 0.08, meaning that the cropped image can be as small as 8\% of the original image. Since 0.08 is too small for the model to extract complete semantic information during data synthesis, increasing the value to 0.5 resulted in a significant performance gain.

\subsection{Comprehensive Comparison Experiment}

\begin{table}[ht]
    \centering
        \resizebox{1.0\textwidth}{!}{
    \begin{tabular}{c|c|c|c|c|c|c|c|c|c}
\toprule
        Dataset & IPC & MTT & TESLA & SRe$^2$L & G-VBSM & CDA & WMDD & RDED & EDC (Ours) \\
        \hline
        \multirow{3}{*}{CIFAR-10} & 1 & - & - & - & - & - & - & 22.9 ± 0.4 & \graycell 32.6 ± 0.1 \\
        & 10 & 46.1 ± 1.4 & 48.9 ± 2.2 & 27.2 ± 0.4 & 53.5 ± 0.6 & - & - & 37.1 ± 0.3 &  \graycell 79.1 ± 0.3 \\
        & 50 & - & - & 47.5 ± 0.5 & 59.2 ± 0.4 & - & - & 62.1 ± 0.1 &  \graycell 87.0 ± 0.1 \\
        \hline
        \multirow{3}{*}{CIFAR-100} & 1 & - & - & 2.0 ± 0.2 & 25.9 ± 0.5 & - & - & 11.0 ± 0.3 &  \graycell 39.7 ± 0.1 \\
        & 10 & 26.8 ± 0.6 & 27.1 ± 0.7 & 31.6 ± 0.5 & 59.5 ± 0.4 & - & - & 42.6 ± 0.2 &  \graycell 63.7 ± 0.3 \\
        & 50 & - & - & 49.5 ± 0.3 & 65.0 ± 0.5 & - & - & 62.6 ± 0.1 &  \graycell 68.6 ± 0.2 \\
        \hline
        \multirow{3}{*}{Tiny-ImageNet} & 1 & - & - & - & - & - & 7.6 ± 0.2 & 9.7 ± 0.4 &  \graycell 39.2 ± 0.4 \\
        & 10 & - & - & - & - & - & 41.8 ± 0.1 & 41.9 ± 0.2 &  \graycell 51.2 ± 0.5 \\
        & 50 & 28.0 ± 0.3 & - & 41.1 ± 0.4 & 47.6 ± 0.3 & 48.7 & 59.4 ± 0.5 &  \graycell 58.2 ± 0.1 & 57.2 ± 0.2 \\
        \hline
        \multirow{3}{*}{ImageNet-10} & 1 & - & - & - & - & - & - & 24.9 ± 0.5 &  \graycell 45.2 ± 0.2 \\
        & 10 & - & - & - & - & - & - & 53.3 ± 0.1 &  \graycell 63.4 ± 0.2  \\
        & 50 & - & - & - & - & - & - & 75.5 ± 0.5 &  \graycell 82.2 ± 0.1 \\
        \hline
        \multirow{3}{*}{ImageNet-1k} & 1 & - & - & - & - & - & 3.2 ± 0.3 & 6.6 ± 0.2 &  \graycell 12.8 ± 0.1 \\
        & 10 & - & 17.8 ± 1.3 & 21.3 ± 0.6 & 31.4 ± 0.5 & - & 38.2 ± 0.2 & 42.0 ± 0.1 &  \graycell 48.6 ± 0.3 \\
        & 50 & - & 27.9 ± 1.2 & 46.8 ± 0.2 & 51.8 ± 0.4 & 53.5 & 57.6 ± 0.5 & 56.5 ± 0.1 &  \graycell 58.0 ± 0.2 \\
\bottomrule
    \end{tabular}}
    \caption{\small \textbf{Comparison with the SOTA baseline dataset condensation methods.} MTT, TESLA, SRe$^2$L, CDA, WMDD and RDED utilize ResNet-18 for data synthesis, whereas G-VBSM and EDC leverage various backbones for this purpose.}
    \label{tab:comprehensive_results}
\end{table}

Due to space constraints in the main paper and for aesthetic reasons, we have not fully presented the experimental results of other methods. However, since the benchmark for dataset distillation is uniform and well-recognized, the performance of other algorithms can be found in their respective papers. We present the related experimental results of the popular convolutional architecture ResNet-18 in Table~\ref{tab:comprehensive_results}.

\clearpage
\section*{NeurIPS Paper Checklist}

\begin{enumerate}

\item {\bf Claims}
    \item[] Question: Do the main claims made in the abstract and introduction accurately reflect the paper's contributions and scope?
    \item[] Answer: \answerYes{} 
    \item[] Justification: In the introduction and abstract, we state a comprehensive design framework for dataset condensation, incorporating specific and effective strategies supported by empirical evidence and theoretical foundations.

\item {\bf Limitations}
    \item[] Question: Does the paper discuss the limitations of the work performed by the authors?
    \item[] Answer: \answerYes{}
    \item[] Justification: Please see Sec.~\ref{appendix_limitations}.

\item {\bf Theory Assumptions and Proofs}
    \item[] Question: For each theoretical result, does the paper provide the full set of assumptions and a complete (and correct) proof?
    \item[] Answer: \answerYes{}
    \item[] Justification: Please see Sec.~\ref{theory} in Appendix.

    \item {\bf Experimental Result Reproducibility}
    \item[] Question: Does the paper fully disclose all the information needed to reproduce the main experimental results of the paper to the extent that it affects the main claims and/or conclusions of the paper (regardless of whether the code and data are provided or not)?
    \item[] Answer: \answerYes{}
    \item[] Justification: Our supplemental materials contain the reproducible code.

\item {\bf Open access to data and code}
    \item[] Question: Does the paper provide open access to the data and code, with sufficient instructions to faithfully reproduce the main experimental results, as described in supplemental material?
    \item[] Answer: \answerYes{}
    \item[] Justification: Code has been provided in supplemental materials.

\item {\bf Experimental Setting/Details}
    \item[] Question: Does the paper specify all the training and test details (e.g., data splits, hyperparameters, how they were chosen, type of optimizer, etc.) necessary to understand the results?
    \item[] Answer: \answerYes{} 
    \item[] Justification: The details have been presented in Appendix~\ref{tab:hyperparameter_settings}.

\item {\bf Experiment Statistical Significance}
    \item[] Question: Does the paper report error bars suitably and correctly defined or other appropriate information about the statistical significance of the experiments?
    \item[] Answer: \answerYes{}
    \item[] Justification: Please see Table~\ref{tab:main}.

\item {\bf Experiments Compute Resources}
    \item[] Question: For each experiment, does the paper provide sufficient information on the computer resources (type of compute workers, memory, time of execution) needed to reproduce the experiments?
    \item[] Answer: \answerYes{} 
    \item[] Justification: All experiments are conducted using 4$\times$ RTX 4090 GPUs, as detailed in the experiment section.
    
\item {\bf Code Of Ethics}
    \item[] Question: Does the research conducted in the paper conform, in every respect, with the NeurIPS Code of Ethics \url{https://neurips.cc/public/EthicsGuidelines}?
    \item[] Answer: \answerYes{}
    \item[] Justification: Please see Sec.~\ref{ethics_statement}.

\item {\bf Broader Impacts}
    \item[] Question: Does the paper discuss both potential positive societal impacts and negative societal impacts of the work performed?
    \item[] Answer: \answerYes{}
    \item[] Justification: Please see Sec.~\ref{ethics_statement}.
    
\item {\bf Safeguards}
    \item[] Question: Does the paper describe safeguards that have been put in place for responsible release of data or models that have a high risk for misuse (e.g., pretrained language models, image generators, or scraped datasets)?
    \item[] Answer: \answerNA{} 
    \item[] Justification: There are no risk factors present here.

\item {\bf Licenses for existing assets}
    \item[] Question: Are the creators or original owners of assets (e.g., code, data, models), used in the paper, properly credited and are the license and terms of use explicitly mentioned and properly respected?
    \item[] Answer: \answerYes{} 
    \item[] Justification: In our paper and accompanying code, we have carefully cited and credited the works of G-VBSM and RDED, which form the foundation of our implementation.

\item {\bf New Assets}
    \item[] Question: Are new assets introduced in the paper well documented and is the documentation provided alongside the assets?
    \item[] Answer: \answerYes{}
    \item[] Justification: We have attached our code and user instructions in the supplementary materials.

\item {\bf Crowdsourcing and Research with Human Subjects}
    \item[] Question: For crowdsourcing experiments and research with human subjects, does the paper include the full text of instructions given to participants and screenshots, if applicable, as well as details about compensation (if any)? 
    \item[] Answer: \answerNA{} 
    \item[] Justification: This paper does not have any experiments or research relevant to human subjects.

\item {\bf Institutional Review Board (IRB) Approvals or Equivalent for Research with Human Subjects}
    \item[] Question: Does the paper describe potential risks incurred by study participants, whether such risks were disclosed to the subjects, and whether Institutional Review Board (IRB) approvals (or an equivalent approval/review based on the requirements of your country or institution) were obtained?
    \item[] Answer: \answerNA{} 
    \item[] Justification: Not applicable.

\end{enumerate}

\end{document}